\documentclass[10pt,twocolumn,letterpaper]{article}

\usepackage[pagenumbers]{cvpr}

\usepackage[dvipsnames]{xcolor}

\newcommand\method{{UrbanIR~}}

\usepackage{times}
\usepackage{epsfig}
\usepackage{amsmath}
\usepackage{amssymb}
\usepackage{amsfonts}
\usepackage{bbm}
\usepackage{booktabs}
\usepackage{xcolor}
\usepackage{makecell}
\usepackage{url}            
\usepackage{booktabs}       
\usepackage{amsfonts}       
\usepackage{nicefrac}       
\usepackage{microtype}      
\usepackage{pifont}
\usepackage{color}
\usepackage{multirow}
\usepackage{adjustbox}
\usepackage{bm}
\usepackage{algorithm} 
\usepackage{algpseudocode}
\usepackage{wrapfig}
\usepackage{fancybox}
\usepackage{tikz}
\usepackage[accsupp]{axessibility}

\definecolor{cvprblue}{rgb}{0.21,0.49,0.74}
\usepackage[pagebackref,breaklinks,colorlinks,citecolor=cvprblue]{hyperref}

\def\paperID{102} %
\def\confName{3DV\xspace}
\def\confYear{2025\xspace}

\title{UrbanIR: Large-Scale Urban Scene Inverse Rendering from a Single Video}

\vspace{-10mm}
\author{
Chih-Hao Lin$^{1}$ \quad Bohan Liu$^{1}$ \quad Yi-Ting Chen$^{2}$ \quad Kuan-Sheng Chen$^{1}$ \\ David Forsyth$^{1}$ \quad Jia-Bin Huang$^{2}$ \quad Anand Bhattad$^{1}$ \quad Shenlong Wang$^{1}$\\
$^{1}$University of Illinois Urbana-Champaign \quad $^{2}$University of Maryland, College Park \\
\large{\href{https://urbaninverserendering.github.io/}{https://urbaninverserendering.github.io/}}
}

\begin{document}

\newcommand{\todocite}[1]{\textcolor{blue}{Citation needed []}}
\newcommand{\shenlongsay}[1]{\textcolor{blue}{[{\it Shenlong: #1}]}}

\newcommand{\argmin}[0]{{\texttt{argmin}}}
\newcommand{\mfigure}[2]{\includegraphics[width=#1\linewidth]{#2}}
\newcommand{\mpage}[2]
{
\begin{minipage}{#1\linewidth}\centering
#2
\end{minipage}
}

\newcommand{\xpar}[1]{\noindent\textbf{#1}\ \ }
\newcommand{\vpar}[1]{\vspace{3mm}\noindent\textbf{#1}\ \ }

\newcommand{\sect}[1]{Section~\ref{#1}}
\newcommand{\sects}[1]{Sections~\ref{#1}}
\newcommand{\eqn}[1]{Equation~\ref{#1}}
\newcommand{\eqns}[1]{Equations~\ref{#1}}
\newcommand{\fig}[1]{Figure~\ref{#1}}
\newcommand{\figs}[1]{Figures~\ref{#1}}
\newcommand{\tab}[1]{Table~\ref{#1}}
\newcommand{\tabs}[1]{Tables~\ref{#1}}

\newcommand{\ignorethis}[1]{}
\newcommand{\norm}[1]{\lVert#1\rVert}
\newcommand{\fcseven}{$\mbox{fc}_7$}

\renewcommand*{\thefootnote}{\fnsymbol{footnote}}

\def\naive{na\"{\i}ve\xspace}
\def\Naive{Na\"{\i}ve\xspace}

\makeatletter
\DeclareRobustCommand\onedot{\futurelet\@let@token\@onedot}
\def\@onedot{\ifx\@let@token.\else.\null\fi\xspace}

\def\iid{\emph{i.i.d}\onedot}
\def\eg{\emph{e.g}\onedot} \def\Eg{\emph{E.g}\onedot}
\def\ie{\emph{i.e}\onedot} \def\Ie{\emph{I.e}\onedot}
\def\cf{\emph{c.f}\onedot} \def\Cf{\emph{C.f}\onedot}
\def\etc{\emph{etc}\onedot} \def\vs{\emph{vs}\onedot}
\def\wrt{w.r.t\onedot} \def\dof{d.o.f\onedot}
\def\etal{\emph{et al}\onedot}
\makeatother

\definecolor{citecolor}{RGB}{34,139,34}
\definecolor{mydarkblue}{rgb}{0,0.08,1}
\definecolor{mydarkgreen}{rgb}{0.02,0.6,0.02}
\definecolor{mydarkred}{rgb}{0.8,0.02,0.02}
\definecolor{mydarkorange}{rgb}{0.40,0.2,0.02}
\definecolor{mypurple}{RGB}{111,0,255}
\definecolor{myred}{rgb}{1.0,0.0,0.0}
\definecolor{mygold}{rgb}{0.75,0.6,0.12}
\definecolor{myblue}{rgb}{0,0.2,0.8}
\definecolor{mydarkgray}{rgb}{0.66,0.66,0.66}

\newcommand{\myparagraph}[1]{\vspace{-1.2em}\paragraph{#1}\hspace{-1em}}

\newcommand{\dafparagraph}[1]{{\it #1}}

\newcommand{\bbR}{{\mathbb{R}}}
\newcommand{\bK}{\mathbf{K}}
\newcommand{\bX}{\mathbf{X}}
\newcommand{\bY}{\mathbf{Y}}
\newcommand{\bk}{\mathbf{k}}
\newcommand{\bN}{\mathbf{N}}

\newcommand{\bx}{\mathbf{x}}
\newcommand{\by}{\mathbf{y}}
\newcommand{\bhy}{\hat{\mathbf{y}}}
\newcommand{\bty}{\tilde{\mathbf{y}}}
\newcommand{\bG}{\mathbf{G}}
\newcommand{\bI}{\mathbf{I}}
\newcommand{\bg}{\mathbf{g}}
\newcommand{\bS}{\mathbf{S}}
\newcommand{\bs}{\mathbf{s}}
\newcommand{\bM}{\mathbf{M}}
\newcommand{\bw}{\mathbf{w}}
\newcommand{\eye}{\mathbf{I}}
\newcommand{\bU}{\mathbf{U}}
\newcommand{\bV}{\mathbf{V}}
\newcommand{\bW}{\mathbf{W}}
\newcommand{\bn}{\mathbf{n}}
\newcommand{\bv}{\mathbf{v}}
\newcommand{\bq}{\mathbf{q}}
\newcommand{\bR}{\mathbf{R}}
\newcommand{\bi}{\mathbf{i}}
\newcommand{\bj}{\mathbf{j}}
\newcommand{\bp}{\mathbf{p}}
\newcommand{\bt}{\mathbf{t}}
\newcommand{\bJ}{\mathbf{J}}
\newcommand{\bu}{\mathbf{u}}
\newcommand{\bB}{\mathbf{B}}
\newcommand{\bD}{\mathbf{D}}
\newcommand{\bz}{\mathbf{z}}
\newcommand{\bP}{\mathbf{P}}
\newcommand{\bC}{\mathbf{C}}
\newcommand{\bA}{\mathbf{A}}
\newcommand{\bZ}{\mathbf{Z}}
\newcommand{\bff}{\mathbf{f}}
\newcommand{\bF}{\mathbf{F}}
\newcommand{\bo}{\mathbf{o}}
\newcommand{\bc}{\mathbf{c}}
\newcommand{\bT}{\mathbf{T}}
\newcommand{\bQ}{\mathbf{Q}}
\newcommand{\bL}{\mathbf{L}}
\newcommand{\bl}{\mathbf{l}}
\newcommand{\ba}{\mathbf{a}}
\newcommand{\bE}{\mathbf{E}}
\newcommand{\bH}{\mathbf{H}}
\newcommand{\bd}{\mathbf{d}}
\newcommand{\br}{\mathbf{r}}
\newcommand{\bb}{\mathbf{b}}
\newcommand{\bh}{\mathbf{h}}

\newcommand{\btheta}{\bm{\theta}}
\newcommand{\bhh}{\hat{\mathbf{h}}}
\newcommand{\ci}{{\cal I}}
\newcommand{\ct}{{\cal T}}
\newcommand{\co}{{\cal O}}
\newcommand{\ck}{{\cal K}}
\newcommand{\cu}{{\cal U}}
\newcommand{\cS}{{\cal S}}
\newcommand{\cQ}{{\cal Q}}
\newcommand{\cT}{{\cal S}}
\newcommand{\cC}{{\cal C}}
\newcommand{\cE}{{\cal E}}
\newcommand{\cF}{{\cal F}}
\newcommand{\cL}{{\cal L}}
\newcommand{\X}{{\cal{X}}}
\newcommand{\Y}{{\cal Y}}
\newcommand{\cH}{{\cal H}}
\newcommand{\cP}{{\cal P}}
\newcommand{\cN}{{\cal N}}
\newcommand{\cU}{{\cal U}}
\newcommand{\cV}{{\cal V}}
\newcommand{\cX}{{\cal X}}
\newcommand{\cY}{{\cal Y}}
\newcommand{\graph}{{\cal H}}
\newcommand{\bayes}{{\cal B}}
\newcommand{\cx}{{\cal X}}
\newcommand{\cg}{{\cal G}}
\newcommand{\cm}{{\cal M}}
\newcommand{\cM}{{\cal M}}
\newcommand{\cG}{{\cal G}}
\newcommand{\cR}{\cal{R}}
\newcommand{\R}{\cal{R}}
\newcommand{\eig}{\mathrm{eig}}

\newcommand{\bbS}{\mathbb{S}}

\newcommand{\D}{{\cal D}}
\newcommand{\bfp}{{\bf p}}
\newcommand{\bfd}{{\bf d}}

\newcommand{\cv}{{\cal V}}
\newcommand{\ce}{{\cal E}}
\newcommand{\cy}{{\cal Y}}
\newcommand{\cz}{{\cal Z}}
\newcommand{\cb}{{\cal B}}
\newcommand{\cq}{{\cal Q}}
\newcommand{\cd}{{\cal D}}
\newcommand{\bcf}{{\cal F}}
\newcommand{\cI}{\mathcal{I}}

\newcommand{\ut}{^{(t)}}
\newcommand{\up}{^{(t-1)}}

\newcommand{\bpi}{\boldsymbol{\pi}}
\newcommand{\bphi}{\boldsymbol{\phi}}
\newcommand{\bPhi}{\boldsymbol{\Phi}}
\newcommand{\bmu}{\boldsymbol{\mu}}
\newcommand{\bSigma}{\boldsymbol{\Sigma}}
\newcommand{\bGamma}{\boldsymbol{\Gamma}}
\newcommand{\bbeta}{\boldsymbol{\beta}}
\newcommand{\bomega}{\boldsymbol{\omega}}
\newcommand{\blambda}{\boldsymbol{\lambda}}
\newcommand{\bkappa}{\boldsymbol{\kappa}}
\newcommand{\btau}{\boldsymbol{\tau}}
\newcommand{\balpha}{\boldsymbol{\alpha}}
\def\bgamma{\boldsymbol\gamma}

\newcommand{\prox}{{\mathrm{prox}}}

\newcommand{\pardev}[2]{\frac{\partial #1}{\partial #2}}
\newcommand{\dev}[2]{\frac{d #1}{d #2}}
\newcommand{\dw}{\delta\bw}
\newcommand{\lab}{\mathcal{L}}
\newcommand{\unlab}{\mathcal{U}}
\newcommand{\ind}{1{\hskip -2.5 pt}\hbox{I}}
\newcommand{\ff}[2]{   \cf_{\prec (#1 \rightarrow #2)}}
\newcommand{\dd}[2]{   \delta_{#1 \rightarrow #2}}
\newcommand{\ld}[2]{   \lambda_{#1 \rightarrow #2}}
\newcommand{\en}[2]{  \bD(#1|| #2)}
\newcommand{\ex}[3]{  \bE_{#1 \sim #2}\left[ #3\right]} 
\newcommand{\exd}[2]{  \bE_{#1 }\left[ #2\right]}

\newcommand{\se}[1]{\mathfrak{se}(#1)}
\newcommand{\SE}[1]{\mathbb{SE}(#1)}
\newcommand{\so}[1]{\mathfrak{so}(#1)}
\newcommand{\SO}[1]{\mathbb{SO}(#1)}

\newcommand{\poselow}{\xi}
\newcommand{\pose}{\bm{\poselow}}
\newcommand{\linpose}{\pose^\ell}
\newcommand{\cbpose}{\pose^c}
\newcommand{\rateparam}{v_i}
\newcommand{\bapose}{\bm{\poselow}_i}
\newcommand{\trackingpose}{\bm{\poselow}}
\newcommand{\rotlow}{\omega}
\newcommand{\rot}{\bm{\rotlow}}
\newcommand{\translow}{v}
\newcommand{\trans}{\bm{\translow}}
\newcommand{\hnorm}[1]{\left\lVert#1\right\rVert_{\gamma}}
\newcommand{\lnorm}[1]{\left\lVert#1\right\rVert}
\newcommand{\barate}{v_i}
\newcommand{\trackingrate}{v}
\newcommand{\imgpt}{\mathbf{u}_{i,k,j}}
\newcommand{\mappt}{\mathbf{X}_{j}}
\newcommand{\timet}[1]{\bar{t}_{#1}}
\newcommand{\mf}[1]{\text{MF}_{#1}}
\newcommand{\kmf}[1]{\text{KMF}_{#1}}
\newcommand{\Exp}{\text{Exp}}
\newcommand{\Log}{\text{Log}}

\newcommand{\shiftleft}[2]{\makebox[0pt][r]{\makebox[#1][l]{#2}}}
\newcommand{\shiftright}[2]{\makebox[#1][r]{\makebox[0pt][l]{#2}}}


\begin{figure}
\twocolumn[{
    \renewcommand\twocolumn[1][]{#1}
    \maketitle
    \input{figures/teaser-daf} 
    }]
\end{figure}

\begin{abstract}
    We present {\it UrbanIR}~({\bf Urban} Scene {\bf I}nverse {\bf R}endering), a new inverse graphics model that enables realistic, free-viewpoint renderings of scenes under various lighting conditions with a single video. It accurately infers shape, albedo, visibility, and sun and sky illumination from wide-baseline videos, such as those from car-mounted cameras, differing from NeRF's dense view settings. In this context, standard methods often yield subpar geometry and material estimates, such as inaccurate roof representations and numerous `floaters'. UrbanIR addresses these issues with novel losses that reduce errors in inverse graphics inference and rendering artifacts. Its techniques allow for precise shadow volume estimation in the original scene. The model's outputs support controllable editing, enabling photorealistic free-viewpoint renderings of night simulations, relit scenes, and inserted objects, marking a significant improvement over existing state-of-the-art methods. Our code and data will be made publicly available upon acceptance.
\end{abstract}

\section{Introduction}

We show how to build a model that allows realistic, free-viewpoint renderings of a scene under novel lighting conditions from a video.  
So, for example, a sunny afternoon video of a large urban scene can be shown at different times of day or night (as in Fig.~\ref{fig:teaser}), viewed from novel viewpoints, and shown with inserted objects.  
Our method --- {\it UrbanIR}~({\bf Urban} Scene {\bf I}nverse {\bf R}endering) --- computes an inverse graphics representation from the video.  
UrbanIR jointly infers shape, albedo, visibility, and sun and sky illumination \emph{from a single video of unbounded outdoor scenes} with \emph{unknown lighting}. %
The resulting representations enable controllable editing, delivering photorealistic free-viewpoint renderings of relit scenes and inserted objects, as demonstrated in Fig.~\ref{fig:teaser}.

UrbanIR obtains its intrinsic scene representations 
from a video under a \emph{single illumination condition}, but producing realistic novel views requires accurate inferences of physical parameters.
UrbanIR uses a novel visibility rendering scheme and loss to precisely estimate shadow volumes in the original scene and control albedo errors.  
UrbanIR combines monocular intrinsic decomposition and inverse rendering with other key contributions to control errors in renderings. To our knowledge, UrbanIR is the first in its class capable of performing inverse rendering and relighting applications from a single monocular video in large-scale scenes, without requiring multiple illumination, depth sensing, or both. 

UrbanIR representations are constructed from cameras mounted on cars with a narrow range of views of each scene point.  
Typical NeRF-style systems yield poor geometry estimates (for example, roofs) and ``floaters'' under these conditions; they
are usually trained with a wide range of views.  
Our experiments showcase that UrbanIR outperforms these baselines with significantly reduced artifacts in our sparse view setting.  
Finally, we show how to use UrbanIR to simulate night scenes from a single daytime-captured video, producing a controllable, realistic, physically plausible, and consistent simulation. In summary, our contributions are:
\begin{itemize}
\item We present UrbanIR for recovering a \emph{relightable} neural radiance field in a constrained setting of an \emph{unbounded scene}, using a \emph{single monocular video} captured under a \emph{single illumination condition}.
\item We describe a novel inverse rendering framework that \emph{builds precise shadow volumes} in large outdoor scenes with heavy shadows, resulting in significant improvements in inverse graphics estimates and relighting. 
\item We demonstrate a new physics-informed night simulation framework. To our knowledge, UrbanIR is the first simulation to offer realistic, \emph{free-viewpoint night simulation} from a single daytime video capture. 

  \end{itemize}

\section{Related Work}

\paragraph{Inverse Graphics} involves inferring illumination and intrinsic properties of a scene. 
The problem is underconstrained, and there is much reliance on priors~\cite{land1971lightness,horn1974determining,horn1975obtaining,Barrow1978, Zhang-sfs-99, barron2014shape, sato1997object,marschner1998inverse,yu1999inverse} or on managed lighting conditions~\cite{hauagge2013photometric,barron2014shape,grosse2009ground,hauagge2013photometric,barron2014shape,zhang2022iron}, known
geometry~\cite{sato2003illumination,lensch2003image,dong2014appearance,laffont2012rich}, or material
simplifications~\cite{zhou2015learning,ma2018single, Zhang-sfs-99}.
Recent methods use deep learning techniques to reason about material properties~\cite{lombardi2015reflectance, lombardi2016radiometric, lombardi2019neural, yu2019inverserendernet,zhang2020image,munkberg2022extracting}. 
Models trained on synthetic data~\cite{lichy2021shape} or pair-wise annotated data~\cite{Bell2014} have shown promising results. 
Learned predictors of albedo or shading are described and reviewed in~\cite{sengupta2019neural, forsyth2021intrinsic, bhattad2023stylegan}. 
Neural representations of material or illumination appear in~\cite{lombardi2019neural, li2018learning, li2018materials,  li2020inverse, li2022physically, bhattad2023stylitgan}. 
Like these methods, we exploit monocular cues, such as shadows and surface normals.  
In contrast, we combine learning-based monocular cues and model-based relightable NeRF optimization to infer the scene's intrinsic properties and illumination.

\paragraph{Shadow modeling} using images is challenging. 
Methods trained to cast shadows from images~\cite{wang2021repopulating, zhang2019shadowgan} 
are tailored for particular objects
(pedestrians, cars, etc).
Learned methods can detect and remove shadows from 2D images~\cite{guo2023shadowformer, guo2022shadowdiffusion,wan2022style}. 
But inverse graphics require modeling the full 3D geometry, intrinsic scene properties, and ensuring temporal consistency. 
Model-based optimization methods can infer shadows but rely on accurate scene geometry~\cite{story2015hybrid, laine2005soft, wu2007natural}.
Using visibility fields to model shadows results in difficulty providing consistent shadows in relation to the underlying geometry~\cite{srinivasan2021nerv, yang2022s, rudnev2022nerfosr, zhang2022invrender}. In contrast, our method combines the strengths of learning-based monocular shadow prediction and removal and model-based inverse graphics.

\paragraph{Relightable Neural Fields:}
 Relightable neural radiance field methods~\cite{zhang2021physg, bossNeuralPILNeuralPreIntegrated2021, zhang2021nerfactor, boss2021nerd, munkbergExtractingTriangular3D2021, hasselgren2022nvdiffrecmc, wang2023neural, yu2019inverserendernet} aim to factor the neural field into multiple intrinsic components and leverage neural shading equations for illumination and material modeling. 
These methods admit realistic and controllable rendering of scenes with varying lighting conditions and materials. 
However, most relightable NeRF methods focus on objects with surrounding views or small bounded indoor environments.
Important exceptions are: NeRF-OSR~\cite{rudnev2022nerfosr}, which assumes access to multiple lighting sources 
for decomposition, and FEGR~\cite{wang2023fegr}, which either uses multiple lighting or exploits depth sensing, such as LiDAR. 

We compare the problem setting, input requirement with recent methods in Tab.~\ref{tab:methods}. \method~addresses inverse rendering for large-scale urban scenes that object-centric methods~\cite{zhang2021nerfactor, jin2023tensoir, zhang2022invrender}  fails to reconstruct. Furthermore, our method takes videos under single illuminations as input, which is more applicable to a broader range of scenes. To estimate the geometry of large-scale driving scenes, FEGR~\cite{wang2023fegr} and LightSim~\cite{pun2023neural} rely on captures from five to six cameras and LiDAR sensors. On the other hand, \method~only needs videos from single or stereo cameras without any guidance from LiDAR. Our method also performs nighttime simulation by inserting local light sources (e.g. streetlight, vehicle light), which is not demonstrated in previous works.

\begin{table}[t]
\centering
\resizebox{1.0\linewidth}{!}{%
\begin{tabular}{lccccc}
\toprule
Method              & Scene  & \makecell{Illumination \\ Conditions} &  \makecell{RGB \\ Only} & \makecell{Explicit \\ shadow} & \makecell{Night \\ Sim.}\\
\midrule
NeRFFactor~\cite{zhang2021nerfactor}   & Object  & Multi   & Yes &  & \\

TensoIR~\cite{jin2023tensoir}          & Object  & Single  & Yes & \checkmark & \\

InvRender~\cite{zhang2022invrender}    & Object  & Single  & Yes & & \\

NeRF-OSR~\cite{rudnev2022nerfosr}      & Front-Facing & Multi  &  Yes &  & \\

FEGR~\cite{wang2023fegr}               & Large Scene  & Single/Multi  &  LiDAR & \checkmark &  \\

LightSim~\cite{pun2023neural}          & Large Scene  & Single/Multi  &  LiDAR & \checkmark &  \\

UrbanIR (Ours)                         & Large Scene  & Single  &  Yes & \checkmark  &  \checkmark\\
\bottomrule
\end{tabular}
}
\caption{Comparison of various recent relightable NeRF methods. UrbanIR is among the first to offer single-illumination and RGB-only relightable NeRF capabilities suitable for large-scale scenes.}
\vspace{-5mm}
\label{tab:methods}
\end{table}

\begin{figure*}[t]
    \centering
    \includegraphics[width=\textwidth]{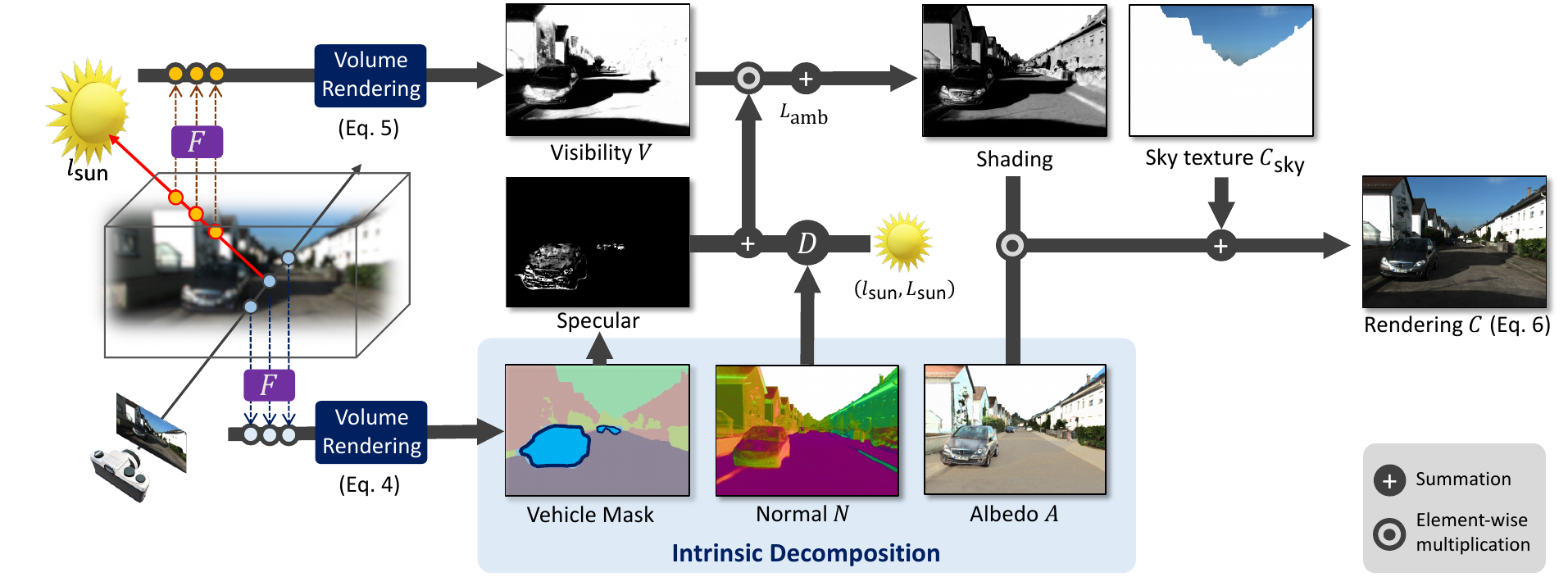}
    \vspace{-20pt}
    \caption{\textbf{Rendering Pipeline.} 
    \method retrieves scene intrinsics (normal $N$, semantics $S$, albedo $A$) from camera rays, and estimate visibility $V$ from tracing rays to the light source. The shading model computes diffuse and specular reflection and adds ambient sky light $\bL_{\text{sky}}$ for the final shading map. We multiply shading \& albedo, and render the sky appearance for final rendering. (Eq.~\ref{eq:shading} for more details.)} 
    \label{fig:rendering}
        \vspace{-5pt}
\end{figure*}

\section{Method}
\method~takes a multi-frame video of a scene under single illumination; as the camera moves, its motion is known.  
Write  $\{I_i, E_i, K_i\}$, where $I_i \in \mathbb{R}^{H \times W \times 3}$ is the RGB image;
$E_i \in \text{SE}(3)$ is the camera pose; and $K_i$ is camera intrinsic matrix.
We produce a neural field model that can be viewed from \emph{novel camera viewpoints} under \emph{novel lighting conditions}. 
We do so by constructing a neural scene model that encodes albedo, normal, semantics, and visibility in a unified manner (Sec.~\ref{sect:scene}). 
This model is rendered from a given camera pose with given illumination using an end-to-end differentiable volume renderer (Sec.~\ref{sect:rendering}). 
Our inference is by joint optimization of all properties (Sec.~\ref{sect:optimization}). 
Applications include changing the sun angle (\cref{fig:teaser}; top right), day-to-night transitions (\cref{fig:teaser}; bottom right), and object insertion (\cref{fig:teaser}; middle right). More details about applications are in Sec.~\ref{sect:simulation}. Fig.~\ref{fig:rendering} provides an overview of our proposed inverse graphics and simulation framework.

\subsection{Relightable Neural Scene Model}
\label{sect:scene}
\paragraph{The scene representation} is built on Instant-NGP~\cite{muller2022instant,
  queianchen_ngp}, a spatial hash-based voxel NeRF representation. 
Instant-NGP offers numerous advantages, including low memory consumption; high efficiency in training and rendering; and compatibility with expansive outdoor scenes. 
Write $\bx \in \bbR^3$ for position in 3D, $\bd$ for query ray direction, $\theta$ for learnable
scene parameters; NeRF models, including Instant-NGP, learn a radiance field
$F_\theta(\mathbf{x}, \bd) = (\bc, \sigma)$, where $\bc \in \bbR^3$ and $\sigma \in \bbR$ represent observed color and opacity respectively. 
Standard NeRFs have view- and lighting-dependent effects, such as shading, shadow, and specularity, baked into their observed color, making them non-relightable.

In contrast, \method learns a model of the intrinsic scene attributes field independent of viewing angles and lighting conditions.
Write diffuse albedo $\ba$, surface normal $\bn$, semantic vector $\bs$, and density $\sigma$; then \method
learns:
\begin{equation}
\label{eq:scene}
    F_\theta(\bx) = (\ba, \bn, \bs, \sigma)
\end{equation}
where $\theta$ is learnable parameters. 
The diffuse albedo represents the intrinsic color and texture of the material; 
the normal represents the intrinsic surface geometry; 
density encodes the spatial opacity, and semantics is used as a key to query surface reflectance. 
Following Instant-NGP~\cite{muller2022instant}, we learn a dense feature hash table to represent the scene, and an individual MLP header is used to decode each attribute given a queried feature at point $\bx$. 
We provide the details of the architecture in the supplementary. 
The geometry of the scene is implicitly encoded in $\sigma$. 
In contrast to existing relightable outdoor scene models that demand coupled explicit geometry~\cite{rudnev2022nerfosr, wang2023fegr}, our scene model is implicit, providing compactness and consistency to appearance modeling.

\myparagraph{The lighting model}
\label{sect:light_model}
is a parametric sun-sky model~\cite{lalonde2014lighting, zhang2019all}. This encodes outdoor illumination as: 
\begin{equation}
\label{eq:lighting}
    \bL = \{ (\bL_\textrm{sun}, \psi_\textrm{sun}, \phi_\textrm{sun}), \bL_\textrm{amb}, \bL_\textrm{sky} \}.
\end{equation} 
Our sun model is a 5-DoF representation, encoding sun color $\bL_\textrm{sun}$ along with the azimuth and zenith $\psi_\textrm{sun}, \phi_\textrm{sun}$. 
The $\bL_\textrm{amb}$ model is represented as a 3-DoF ambient light. 
The sky dome model infers the sky texture from the viewing direction: $\bC_{\text{sky}} = \bL_\textrm{sky}(\br)$. 
We chose this minimalist sun-sky model as it is more compact than other alternatives (e.g., HDR dome or Spherical Gaussians) yet has proven highly effective in modeling various outdoor illumination effects~\cite{lalonde2014lighting, zhang2019all}.

\subsection{Rendering}
\label{sect:rendering}

Given the scene model $F_\theta$ and a lighting model $\bL$, rendering involves two steps: 
1) volume rendering of the scene's intrinsic properties and visibility map onto the image plane, and 2) a shading process to produce the final result with view-dependent and lighting-dependent effects:
\begin{equation}
\label{eq:rendering}
\bC = \texttt{Shade}(\texttt{Intrinsic}(F_\theta, \br), \texttt{Shadow}(F_\theta, \br, \bL), \bL)
\end{equation}
where $\bL$ is the lighting model, $\bC$ is the final RGB color.

\emph{Intrinsics images}  
are obtained by volume rendering.  
We accumulate predictions from $F(\cdot; \theta)$ along the query ray.
Multiple points are sampled along the ray, and intrinsics at the query pixel along the ray~\cite{kajiya1984ray, mildenhall2020nerf}. 
In particular, the albedo $\bA$, normal $\mathbf{N}$, and semantics $\mathbf{S}$ are predicted as:
\begin{equation}\label{eq:volume_rendering}
    \bA(\br) \hspace{-.2em}=\hspace{-.2em} \sum_{i=1}^{N}w_i\ba_i, \mathbf{N}(\br) \hspace{-.2em}= \hspace{-.2em}\sum_{i=1}^{N}w_i\bn_i,\mathbf{S}(\br)\hspace{-.2em} = \hspace{-.2em}\sum_{i=1}^{N}w_i\bs_i,   \\
\end{equation}
where $w_i = \text{exp} (-\sum_{j=1}^{i-1}\sigma_j\delta_j ) \left(1 - \text{exp}(-\sigma_i\delta_i)\right)$ is alpha-composition weight, $\delta_i = t_i - t_{i-1}$. 
We perform rendering for each camera ray and get the final semantic map, albedo map, and the normal map.

\emph{Shadow} modeling and rendering are essential for obtaining realistic-looking outdoor images.  
Modeling the visibility of the sun with a per-scene optimized MLP head (as in~\cite{zhang2022invrender, zhang2021nerfactor}) is impractical because we need to change the sun's position in relighting but can learn from only one position.
An alternative is to construct an explicit geometry model to cast shadows~\cite{wang2023fegr}, but this model might not be consistent with the other neural fields, and imposing consistency is difficult. 
Instead, we first compute an estimate $\bx(\br)$ of the 3D point being shaded, then estimate the visibility $V(\bx, \text{sun})$. 
Our key insight is that shadows in outdoor scenes are primarily due to the visibility of a single directional sunlight. 

We obtain $\bx(\br)$ for each ray by volume rendering depth (so substitute $\hat{t} = \sum w_it_i$ into the equation for the ray being rendered). 
Now, to check whether $\bx$ is visible to the light source, we compute the transmittance along the ray segment between $\bx$ and the light source using volume rendering:
\begin{equation}
\label{eq:visibility}
V(\bx, \text{sun}) \hspace{-.2em}= \hspace{-.2em} \text{exp}\left(\hspace{-.2em}-\sum_i\sigma_i(\bx_i) \delta_i\hspace{-.2em}\right) \hspace{-.2em}\text{\ where\ } \bx_i = \bx + t_i \bl_\text{sun}
\end{equation}
Lower transmittance along a ray from a surface point to a light source suggests fewer obstacles between the point and the light source. 
Eq.~\ref{eq:visibility} establishes a strong link between transmittance, lighting, and visibility fields used in training.
In particular, a point in a training image known as shadowed (resp. out of shadow) should have
large (resp. small) accumulated transmittance. 
We use this constraint to adjust distant geometry during training.
Compared to other alternatives \cite{zhang2022invrender, wang2023fegr}, our proposed visibility test is simple to compute, flexible for relighting, and aligns with intrinsic properties with a few mild assumptions for outdoor scenes. \\

\emph{Shading}
\label{sect:shading_model}
is performed by a Blinn-Phong model~\cite{blinn1977models}
that incorporates sun and sky terms for the foreground scene and an MLP query for the background sky.
For $\bS(\br) \in \text{sky}$, we use
\label{eq:shading}
    $\bC(\br) =  \bL_\textrm{sky}(\br)$
    and otherwise, we use
    \begin{equation}
    \bC(\br)=\bA(\br) \left (\bL_{\text{sun}} \bD \bV + \bL_{\text{amb}} \right) \end{equation}
where $\bD = \text{max}(\mathbf{N}(\br)\cdot {\bf l}_{\text{sun}}, 0)$ is the diffuse lighting at the surface, ${\bf l}_{\text{sun}}$ is the sunlight direction (derived from
$\psi_{\textrm{sun}}, \phi_{\textrm{sun}}$).
The visibility $V(\bx, \text{sun})$ is $1$ if $\bx(\br)$ can see the sun and $0$ otherwise.
This shading model is capable of producing a realistic appearance with shadows following varying lighting
conditions. 
The model can readily be extended with additional lighting sources at the relighting stage, as later shown in the night simulation.

\begin{figure*}[t]
    \centering
    \setlength\tabcolsep{0.1em}
    \resizebox{1.0\textwidth}{!}{%
    \begin{tabular}{@{}clcccc@{}}
    
     Input & & Reconstruction & Albedo  & Surface Normal & Shadow \\[0.2em]
    
    \frame{\includegraphics[width=0.2\textwidth]{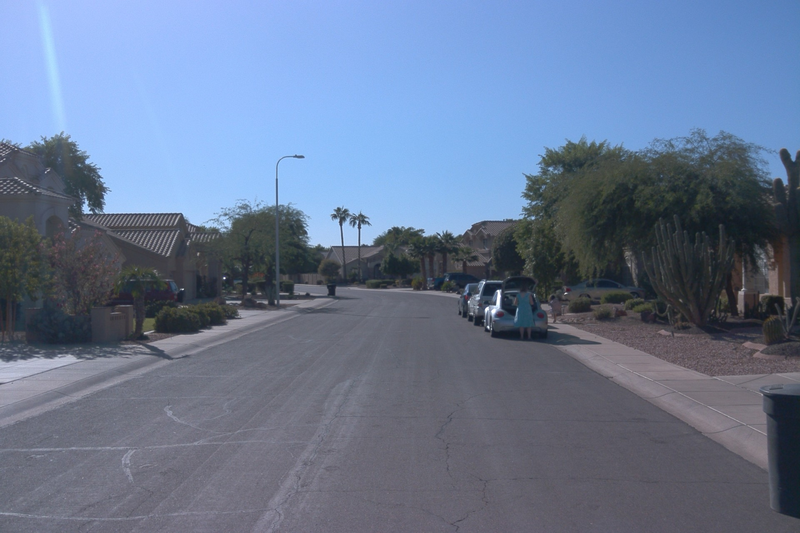}} & \raisebox{2.5\normalbaselineskip}[0pt][0pt]{\rotatebox[origin=c]{90}{\footnotesize FEGR~\cite{wang2023fegr}}} &\frame{\includegraphics[width=0.2\textwidth]{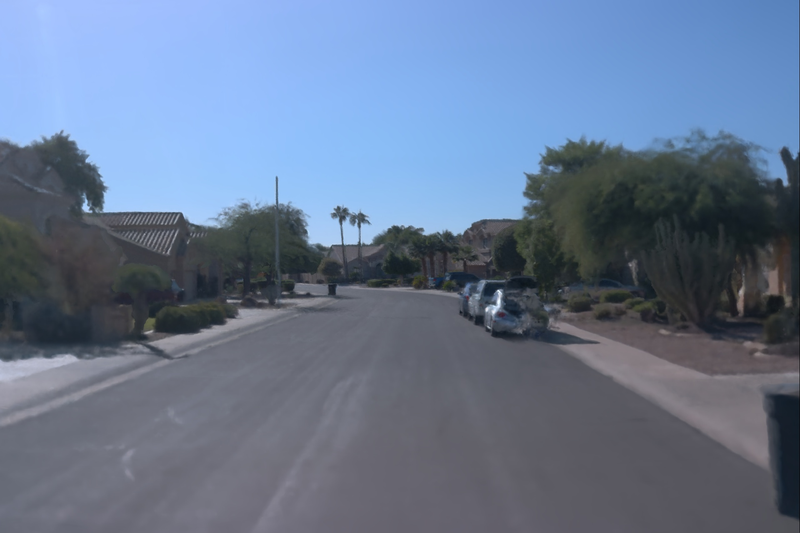}} &\frame{\includegraphics[width=0.2\textwidth]{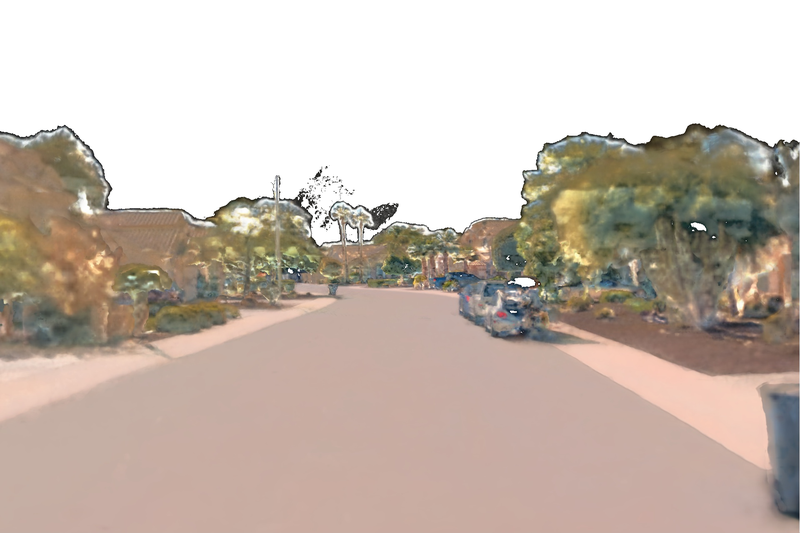}} &\frame{\includegraphics[width=0.2\textwidth]{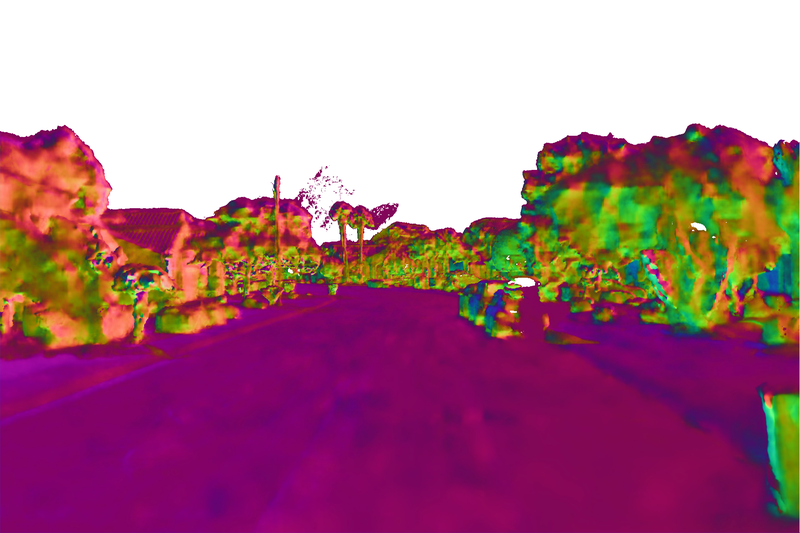}} &  \frame{\includegraphics[width=0.2\textwidth]{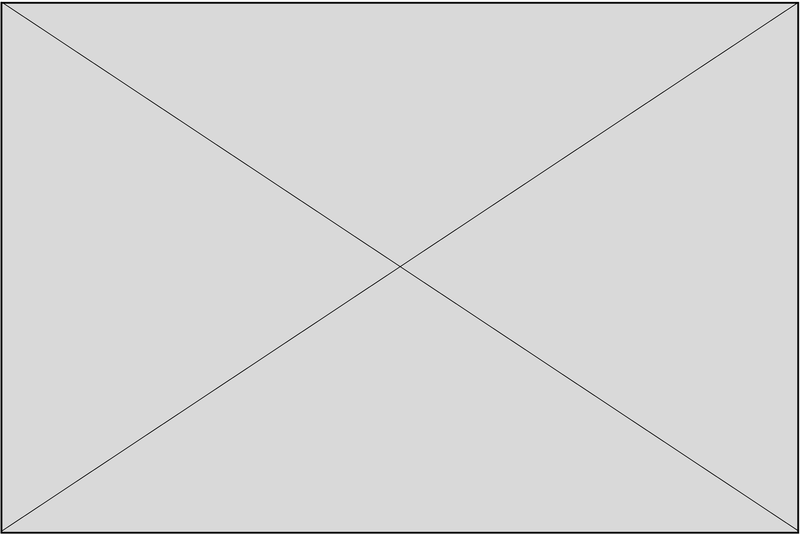}}\\
     & \raisebox{2.5\normalbaselineskip}[0pt][0pt]{\rotatebox[origin=c]{90}{\footnotesize Ours}} &\frame{\includegraphics[width=0.2\textwidth]{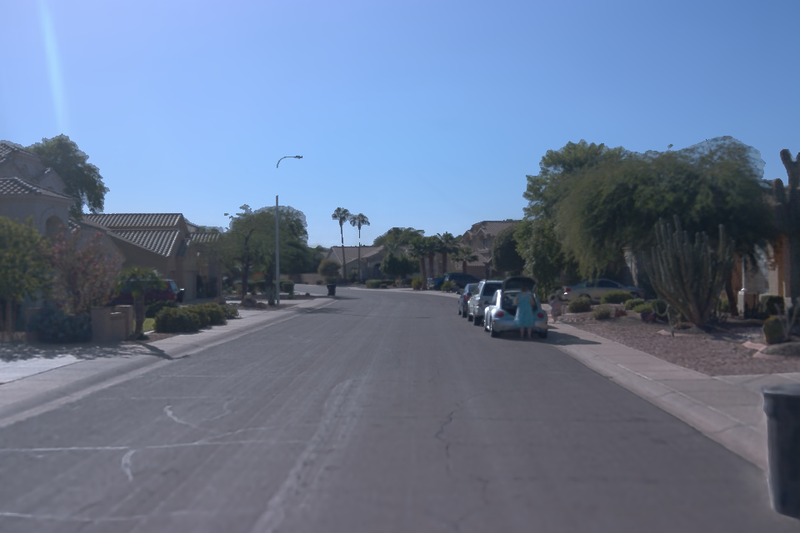}} &\frame{\includegraphics[width=0.2\textwidth]{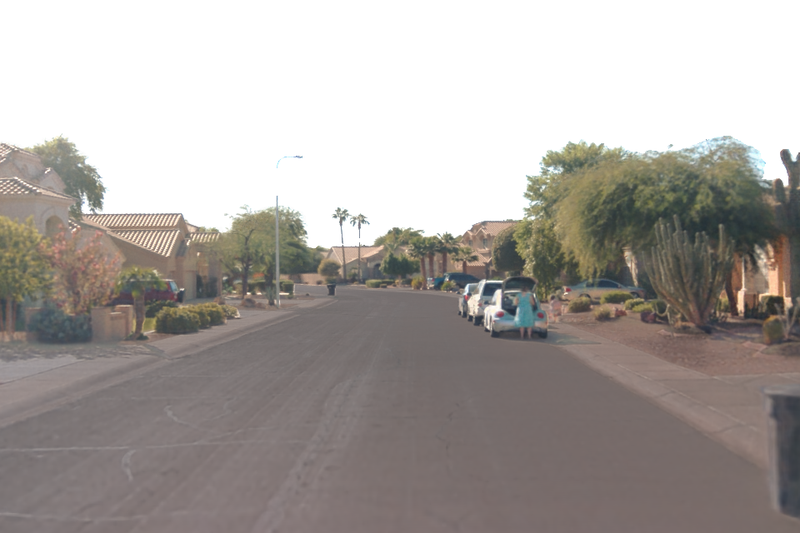}} &\frame{\includegraphics[width=0.2\textwidth]{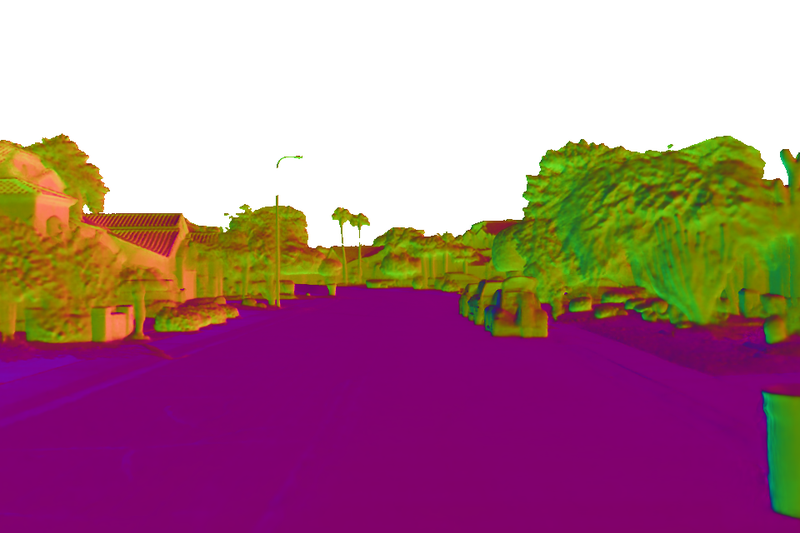}} & \frame{\includegraphics[width=0.2\textwidth]{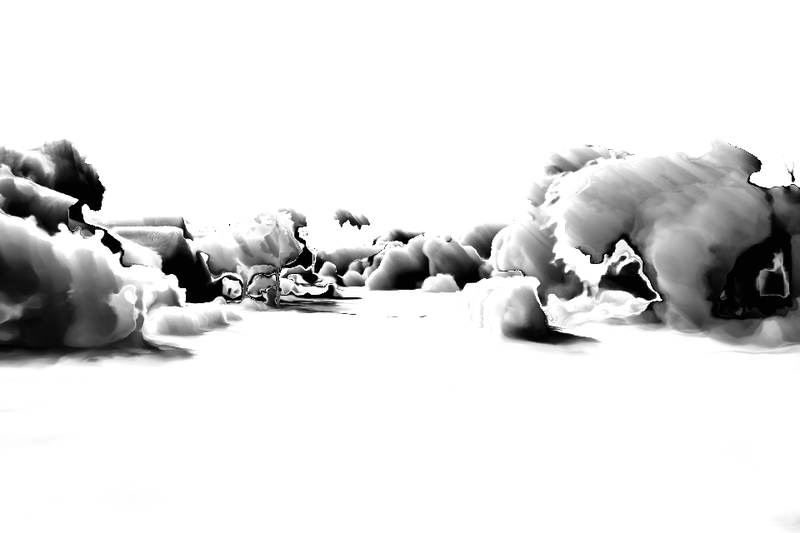}} \\

    \end{tabular}%
    }
    \vspace{-3mm}
    \caption{\textbf{Intrinsic Decomposition of Waymo Open Dataset~\cite{Sun_2020_CVPR}.} We thank the FEGR authors for sharing the results of their Waymo testing sequence with us for comparison. \method~not only decomposes albedo and shadow better but also produces smoother and more detailed albedo and normal. We recommend readers zoom in to view the difference in the intrinsic images.
    }
    \label{fig:waymo_intrinsic}
\end{figure*}

\begin{table*}[t]
    \centering
    \setlength\tabcolsep{0.1em} %
    \resizebox{\textwidth}{!}{
    \begin{tabular}{lcccc}

        \raisebox{5mm}[0pt][0pt]{\rotatebox[origin=c]{90}{\tiny{Input}}} & \frame{\includegraphics[width=0.24\textwidth]{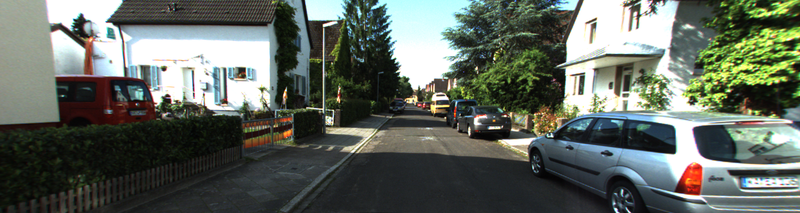}} &  &  &  \\
    
        \raisebox{5mm}[0pt][0pt]{\rotatebox[origin=c]{90}{\tiny{NeRF-OSR~\cite{rudnev2021neural}}}} & \frame{\includegraphics[width=0.24\textwidth]{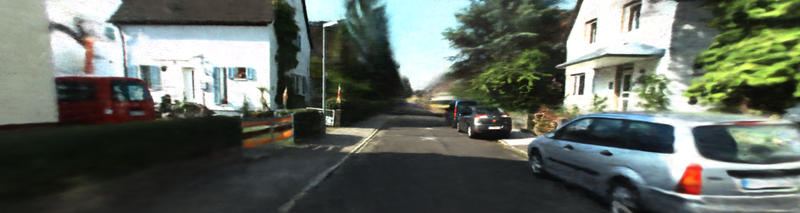}} & \frame{\includegraphics[width=0.24\textwidth]{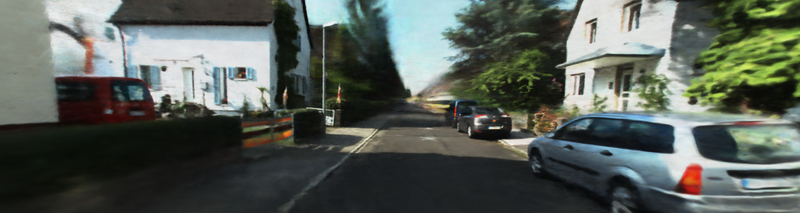}} & \frame{\includegraphics[width=0.24\textwidth]{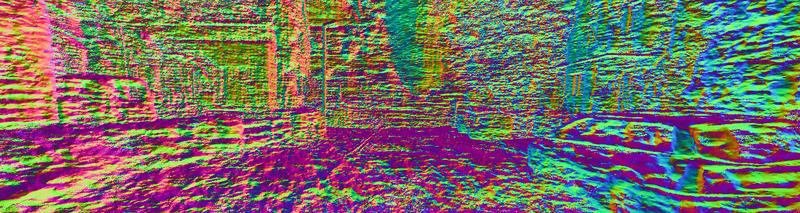}} & \frame{\includegraphics[width=0.24\textwidth]{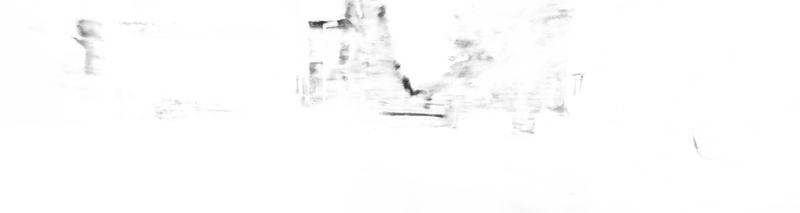}} \\
        \raisebox{5mm}[0pt][0pt]{\rotatebox[origin=c]{90}{\tiny{RelightNet~\cite{yu20relightNet}}}} & \frame{\includegraphics[width=0.24\textwidth]{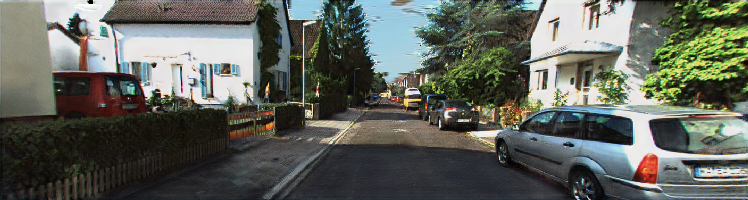}} & \frame{\includegraphics[width=0.24\textwidth]{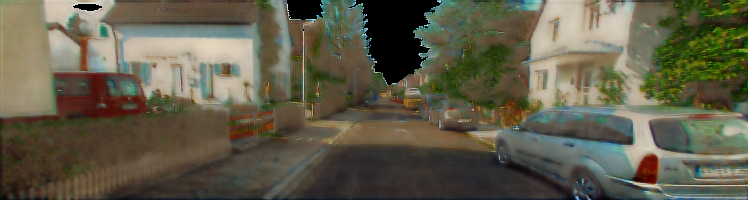}} & \frame{\includegraphics[width=0.24\textwidth]{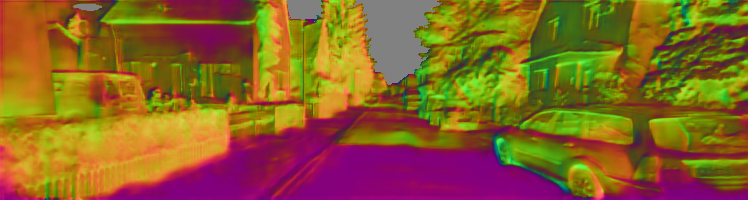}} & \frame{\includegraphics[width=0.24\textwidth]{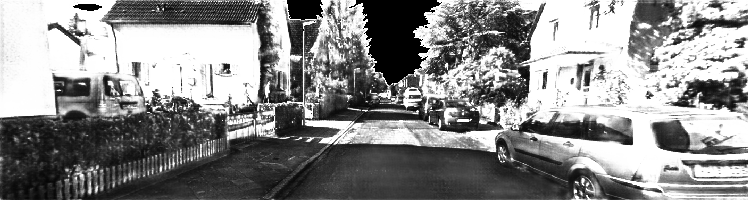}}\\
        \raisebox{5mm}[0pt][0pt]{\rotatebox[origin=c]{90}{\tiny{Ours}}} & \frame{\includegraphics[width=0.24\textwidth]{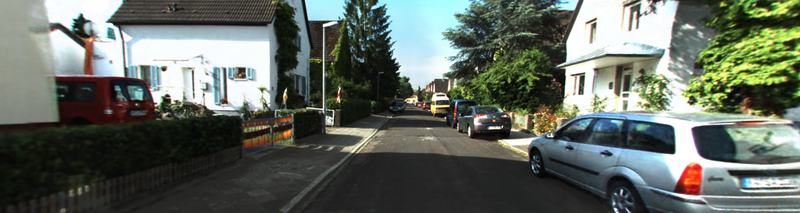}} & \frame{\includegraphics[width=0.24\textwidth]{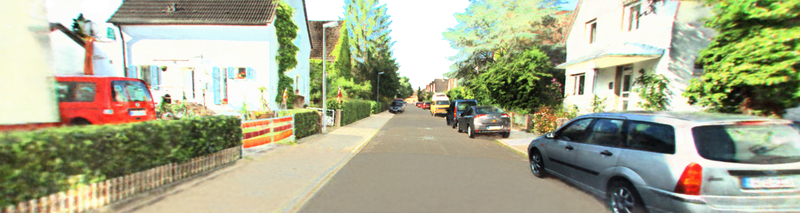}}& \frame{\includegraphics[width=0.24\textwidth]{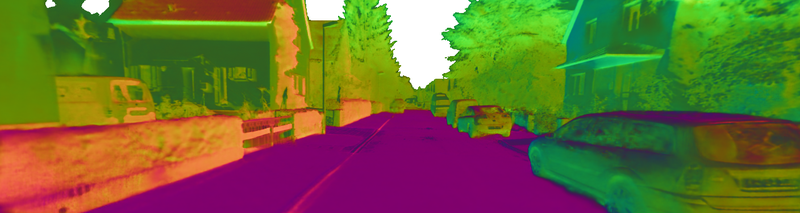}} & \frame{\includegraphics[width=0.24\textwidth]{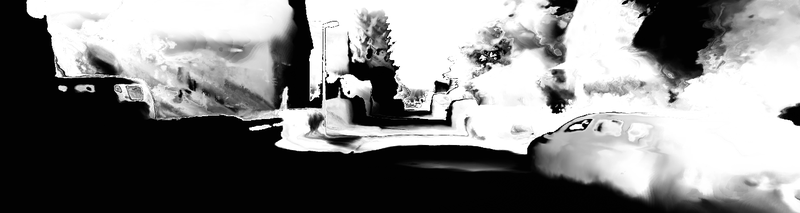}} \\
        & Reconstruction & Albedo & Normal & Shadow/Visibility
    \end{tabular}
    }
    \captionof{figure}{\textbf{Intrinsic Decomposition Comparison.} Please note that NeRF-OSR~\cite{rudnev2021neural} fails to decompose intrinsic, and RelightNet~\cite{rudnev2021neural} tends to bake shadow in the albedo.}
    \label{tab:decompose_compare}
    \vspace{-10pt}
\end{table*}

\subsection{Inverse graphics}
\label{sect:optimization}
We train scene $F(\cdot)$ (Eq.~\ref{eq:scene}) and lighting $\bL$ (Eq.~\ref{eq:lighting}) models jointly using a loss:

\begin{equation}
 \min_{\theta, \bL} \mathcal{L}_{\text{render}} + \lambda_1\mathcal{L}_{\text{visibility}} + \lambda_2\mathcal{L}_{\text{normal}} + \lambda_3\mathcal{L}_{\text{semantics}} + \lambda_4\mathcal{L}_{\text{reg}}, \label{eq:loss}
\end{equation}
where individual loss terms are described below. 

\emph{Rendering loss} measures the agreement between observed images and images rendered from the model using the training view and lighting, yielding
$\mathcal{L}_{\text{render}} = \sum_{\br} \|\bC_{\text{gt}}(\br) - \bC(\br)\|_2^2$, where $\bC$ is rendered color per ray, as defined in Eq.~\ref{eq:rendering}, and $\bC_\text{gt}$ is the observed ``ground-truth'' color. 
Minimizing the rendering loss ensures our scene model can reproduce the observed images.

\emph{Visibility loss} 
recovers unseen geometry with shadow guidance, improving shadow synthesis for relighting.
Specifically, a pixel that is known to be in shadow must be at a point that \emph{cannot see the sun}, so constraining geometry along a ray from that pixel to the sun.  
This loss could be computed by simply comparing visibility $V(\bx, \text{sun})$ with the shadow mask detection~\cite{chen20MTMT}.
However, the 2D shadow detection is not consistent across different frames, making optimization unstable if visibility is supervised with the masks directly. Therefore, we construct an intermediate ``guidance'' visibility estimate $\hat{V}(\br)$ which is an MLP head trained to reproduce the shadow masks, and compute
\begin{equation*}\label{eq:loss_vis}
    \mathcal{L}_{\text{visibility}} = \displaystyle\sum_{\br \in \mathcal{R}}\text{CE}\left(M(\br), \hat{V}(\br)\right) +
    \text{CE}\left(V(\br), \hat{V}(\br)\right),
\end{equation*} 
where $M(\br)$ is the shadow mask at pixel $\br$, , and $\text{CE}(.,.)$ is a cross-entropy loss.  
Here, the first term forces the $\hat{V}$ to generate consistent shadow masks, and the second forces $V$ to agree with $\hat{V}$, recovering scene geometry that is not captured in the images but still cast shadows (e.g. top of the buildings).

\emph{Normal loss} is computed by comparing results $N_{\text{gt}}$ from an off-the-shelf normal estimator~\cite{eftekhar2021omnidata, kar20223d} to the output of the normal MLP.
An alternate estimate of the normal follows from the density field: $\hat{N}(\br) = - \frac{\nabla \sigma(\bx)}{\|\nabla\sigma(\bx)\|}$. We found that enforcing the consistency between the normal estimation improves the geometry, thus enhances relighting quality significantly. Then our normal loss is given by:
\begin{equation*}\label{eq:normal_loss}
    \mathcal{L}_{\text{normal}} = \displaystyle\sum_{\br \in \mathcal{R}}\left(\|N_{\text{gt}}(\br) - N(\br)\|^2 + \|N(\br) - \hat{N}(\br)\|^2\right).
\end{equation*}
We also adopt normal regularization from Ref-NeRF~\cite{verbin2022refnerf} to produce smoother geometry. 

\emph{Semantic loss} is computed by comparing
predicted semantics $\bs$ with labels in the dataset~\cite{Liao2021ARXIV} or detected with~\cite{mmseg2020}.
We use an additional loss to encourage high-depth values in the sky region, reducing floaters in the sky:
\begin{equation*}\label{eq:sem_loss}
    \mathcal{L}_{\text{semantics}} = \displaystyle\sum_{\br \in \mathcal{R}}\text{CE}\left(S_{\text{gt}}(\br), S(\br)\right) - \displaystyle\sum_{\br \in \text{sky}}D(\br).
\end{equation*}

\emph{A regularization term} is used to regularize the albedo of the scene and ambient light intensity. This is necessary due to the ill-posed nature of our optimization process. However, removing the hard shadow from the sunlight in the albedo field $\bA$ remains a challenge, particularly in urban driving sequences. To address this challenge, we introduce a prior that ensures the ground albedo is homogeneous. This is important because the ground region typically shares a similar albedo value.
More specifically, we first compute the average ground albedo $\bar{\bA_{\text{g}}}$ from albedo $\bA$ and semantic $S_{\text{gt}}$ and regularize the albedo using $\mathcal{L}_{\text{albedo}} = \displaystyle\sum_{\br \in \text{ground}} \| \bA(\br) - \bar{\bA_{\text{g}}} \|_2 $.

We also calculate an \emph{ambient regularization} term as $\|\bL_\textrm{amb}\|_2$. We regularize the intensity of ambient light to avoid unnatural color shifts in the recovered albedo caused by a large intensity of ambient light.
Our regularization term is thus $\mathcal{L}_{\text{reg}} = \mathcal{L}_{\text{albedo}} + \|\bL_\textrm{amb}\|_2$.

\subsection{Applications}
\label{sect:simulation}

As the geometry, lighting, albedo, and semantics are recovered, \method~enalbes numerous scene-editing applications, including (1) change sunlight direction and cast the corresponding shadow; (2) turn off sunlight and introduce new light sources (e.g. streetlights) for nighttime simulation; and (3) insert virtual objects and synthesize realistic shading. We encourage the readers to read the supplementary material for implementation details.

\begin{table*}[t]
    \centering
    \setlength\tabcolsep{0.1em} %
    \resizebox{\textwidth}{!}{
    \begin{tabular}{ccc}
        \frame{\includegraphics[width=0.32\textwidth]{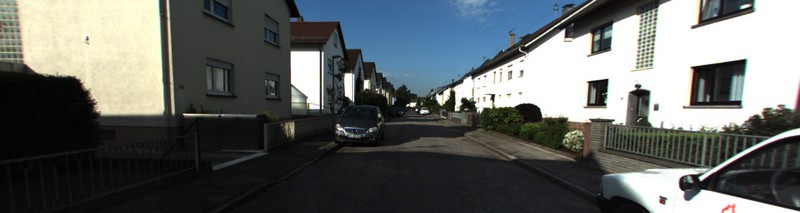}} & \frame{\includegraphics[width=0.32\textwidth]{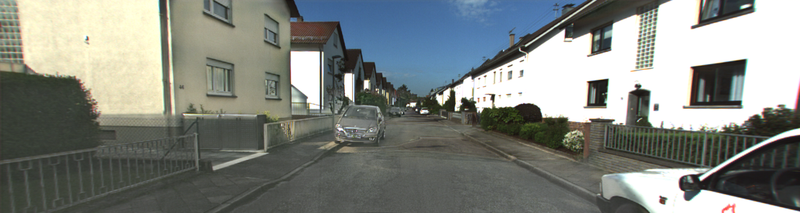}} & \frame{\includegraphics[width=0.32\textwidth]{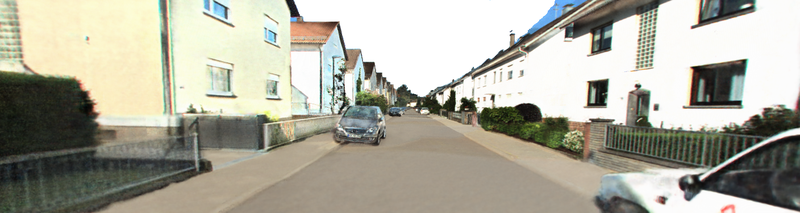}} \\

    Input Image & ShadowFormer~\cite{guo2023shadowformer} & Our Albedo
    \end{tabular}
    }
    \vspace{-3mm}
    \captionof{figure}{\textbf{Shadow Removal in Albedo.} Our method correctly recovers albedo under a shadow while ShadowFormer~\cite{guo2023shadowformer} fails to.
    }
    \label{tab:deshadow}
    \vspace{-5pt}
\end{table*}

\begin{table*}[t]
    \centering
    \setlength\tabcolsep{0.1em} 
    \renewcommand{\arraystretch}{0.4}%
    \resizebox{\textwidth}{!}{
    \begin{tabular}{lccc}
        \raisebox{7mm}[0pt][0pt]{\rotatebox[origin=c]{90}{\footnotesize{Input}}} & \frame{\includegraphics[width=0.32\textwidth]{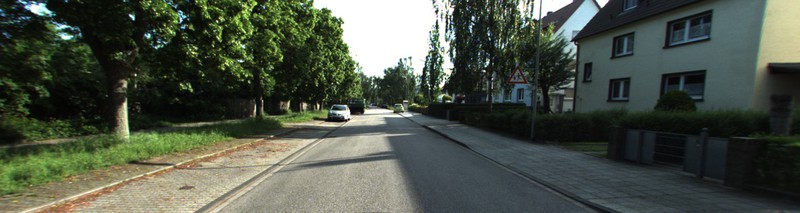}} & \frame{\includegraphics[width=0.32\textwidth]{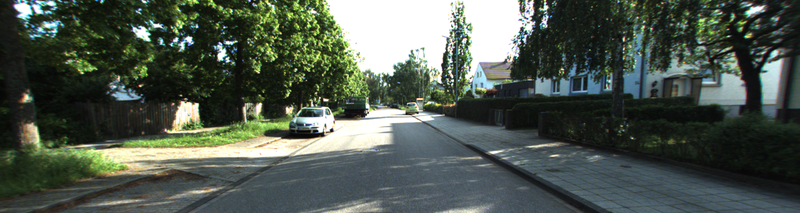}} & \frame{\includegraphics[width=0.32\textwidth]{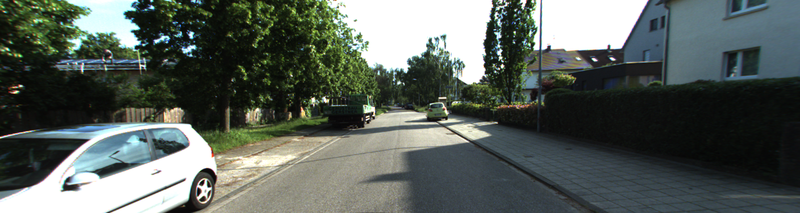}}  \\

        \raisebox{7mm}[0pt][0pt]{\rotatebox[origin=c]{90}{\footnotesize{I-N2N~\cite{instructnerf2023}}}} & \frame{\includegraphics[width=0.32\textwidth]{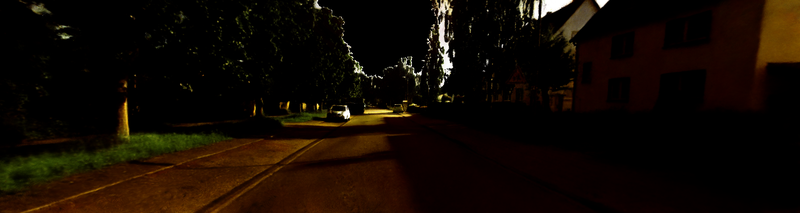}} & \frame{\includegraphics[width=0.32\textwidth]{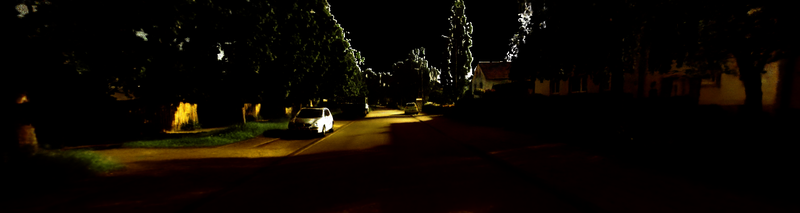}} & \frame{\includegraphics[width=0.32\textwidth]{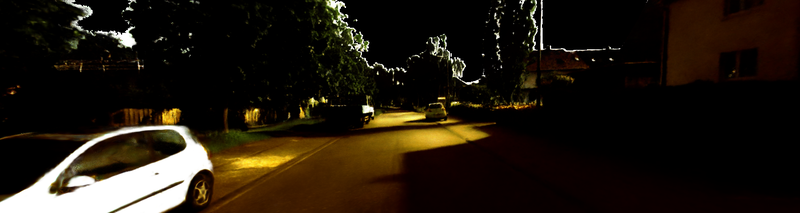}} \\

        \raisebox{7mm}[0pt][0pt]{\rotatebox[origin=c]{90}{\footnotesize{Ours}}} & \frame{\includegraphics[width=0.32\textwidth]{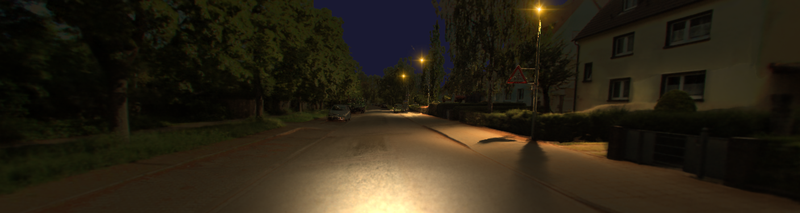}} & \frame{\includegraphics[width=0.32\textwidth]{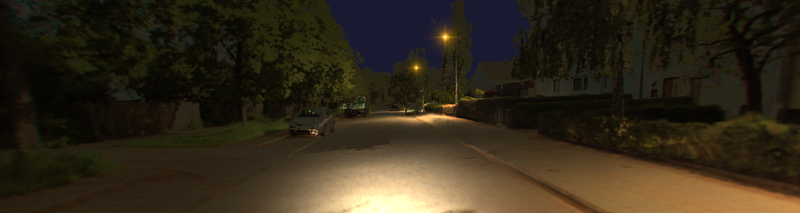}} & \frame{\includegraphics[width=0.32\textwidth]{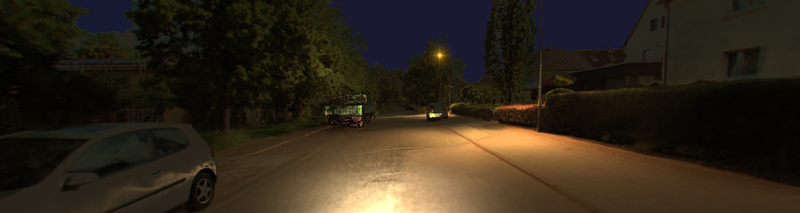}} \\ \\
        \raisebox{7mm}[0pt][0pt]{\rotatebox[origin=c]{90}{\footnotesize{Input}}} & \frame{\includegraphics[width=0.32\textwidth]{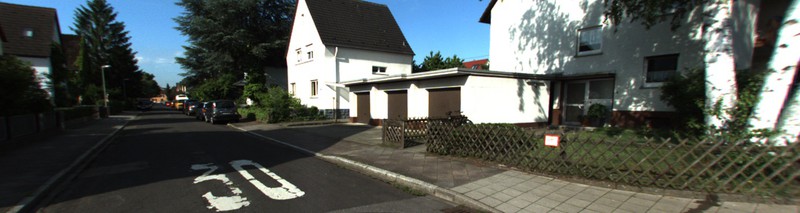}} & \frame{\includegraphics[width=0.32\textwidth]{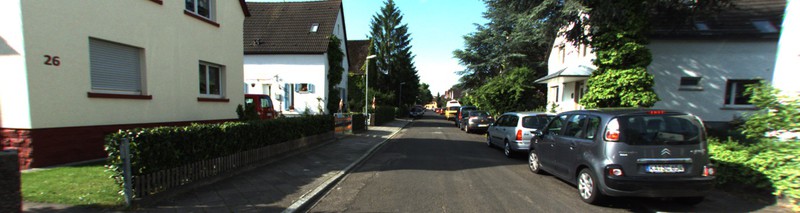}} & \frame{\includegraphics[width=0.32\textwidth]{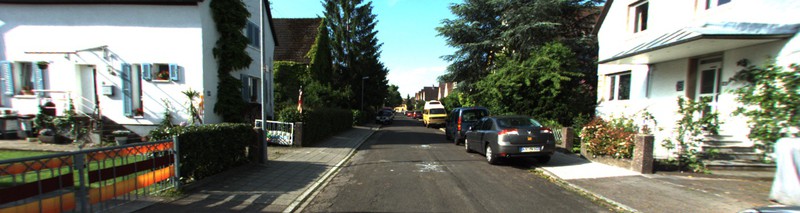}} \\

        \raisebox{7mm}[0pt][0pt]{\rotatebox[origin=c]{90}{\footnotesize{I-N2N~\cite{instructnerf2023}}}} & \frame{\includegraphics[width=0.32\textwidth]{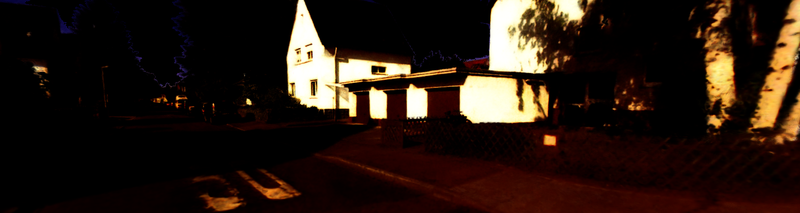}} & \frame{\includegraphics[width=0.32\textwidth]{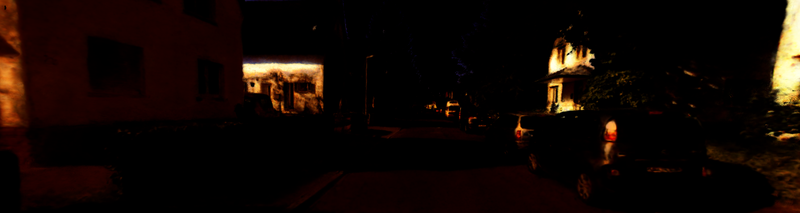}} & \frame{\includegraphics[width=0.32\textwidth]{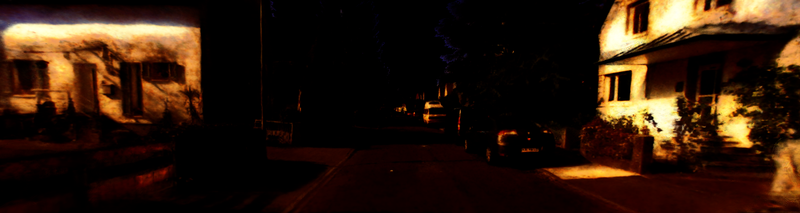}} \\ 
        
        \raisebox{7mm}[0pt][0pt]{\rotatebox[origin=c]{90}{\footnotesize{Ours}}} & \frame{\includegraphics[width=0.32\textwidth]{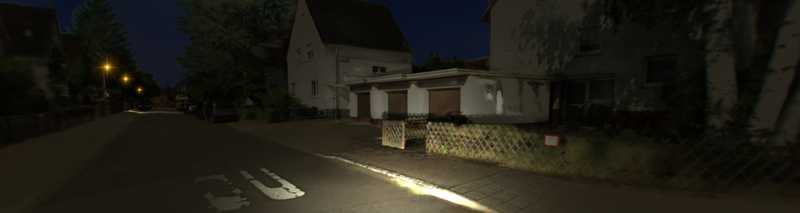}} & \frame{\includegraphics[width=0.32\textwidth]{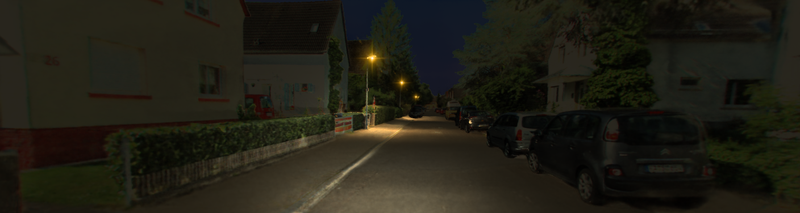}} & \frame{\includegraphics[width=0.32\textwidth]{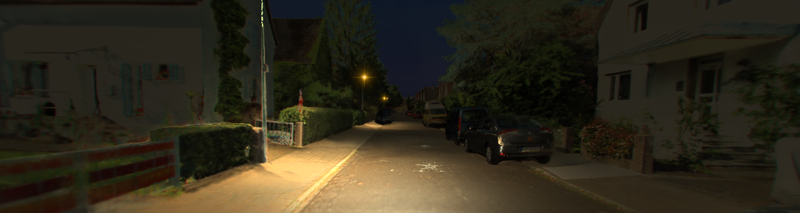}} \\

        & Camera pose 0 (t = 0s) & Camera pose 1 (t = 1s) & Camera pose 2 (t = 2s) \\

    \end{tabular}
    }
    \captionof{figure}{{\bf Nighttime rendering.} The scene changes from daytime to nighttime by introducing new light sources, such as headlights on a car and streetlights. The top three and bottom three rows are from the same driving video but at different times. \method~successfully removes dark shadows with sharp boundaries, resulting in a more realistic rendering of new light sources (such as streetlights and headlights) during night-time simulations. Our method is superior to Instruct-NeRF2NeRF~\cite{instructnerf2023}, which relies on generative prior.}
    \label{tab:relight_night}
    \vspace{-5pt}
\end{table*}

\section{Experiment Results}
\subsection{Datasets}
We evaluate \method on two datasets: the KITTI-360 dataset~\cite{Liao2021ARXIV} and the Waymo Open Dataset~\cite{Sun_2020_CVPR}. The KITTI-360 dataset~\cite{Liao2021ARXIV} consists of 9 stereo video sequences showcasing urban scenes. For our analysis, we selected 7 non-overlapping clipped sequences, each containing around 100 images. These sequences cover various light directions, vehicle trajectories, and layouts of buildings and vegetation. The dataset includes RGB images from stereo cameras, semantic labels, camera poses, and RTK-GPS poses. On the other hand, the Waymo Open Dataset (WOD)~\cite{Sun_2020_CVPR} captures driving sequences from five cameras and one 64-beam LiDAR sensor at 10 Hz. However, we only used the single camera from the front view and did not use any LiDAR information for our evaluation. 

Quantitative evaluation of relighting sequences is difficult as most datasets only capture the same location under a single illumination, and no ground truth for relighting is available. Therefore, we recorded a scene at different times of the day, covering different illuminations. The images were captured by a stereo camera, and the poses were estimated using RTK-GPS information.

\begin{table*}[t]
    \centering
    \setlength\tabcolsep{0.1em}
    \renewcommand{\arraystretch}{0.2}
    \resizebox{\textwidth}{!}{
    \begin{tabular}{lccc}

    \raisebox{7mm}[0pt][0pt]{\rotatebox[origin=c]{90}{\footnotesize{Ours}}} & \frame{\includegraphics[width=0.32\textwidth]{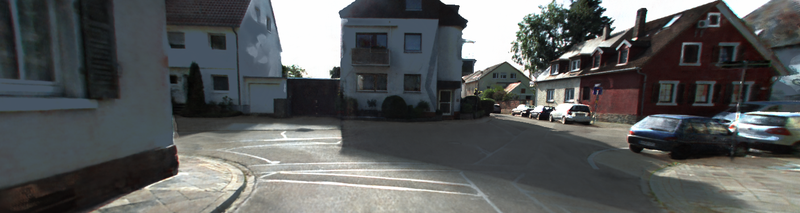}} & \frame{\includegraphics[width=0.32\textwidth]{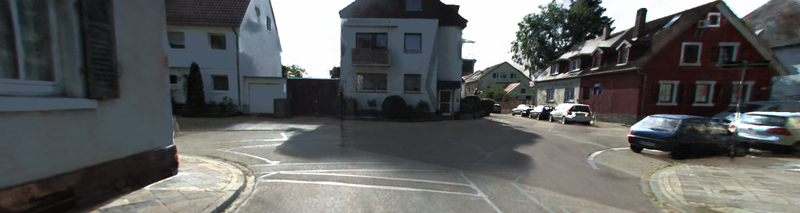}} & \frame{\includegraphics[width=0.32\textwidth]{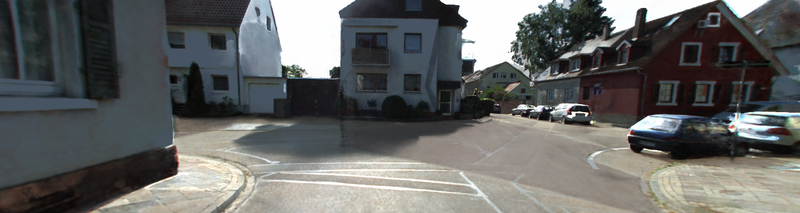}} \\
    \\
       
    & Reconstruction(Original lighting) &  Novel sunlight direction 1 &  Novel sunlight direction 2 \\
    
    \end{tabular}
    }
    \vspace{-5pt}
    \captionof{figure}{{\bf Rendering and relighting comparison}. UrbanIR leverages optimization to enable realistic and controllable relighting effects, demonstrating effectiveness in simulating different sunlight directions from a single video input. 
}
    \label{tab:relight_compare}
    \vspace{-5pt}
\end{table*}

\begin{figure*}[t]
    \centering
    \setlength\tabcolsep{0.1em}
    \resizebox{1.0\textwidth}{!}{%
    \begin{tabular}{@{}ccccc@{}}
    
     Input &  Sun pose 1 & Sun pose 2  & Sun pose 3 & Sun pose 4 \\[0.2em]

    \frame{\includegraphics[width=0.2\textwidth]{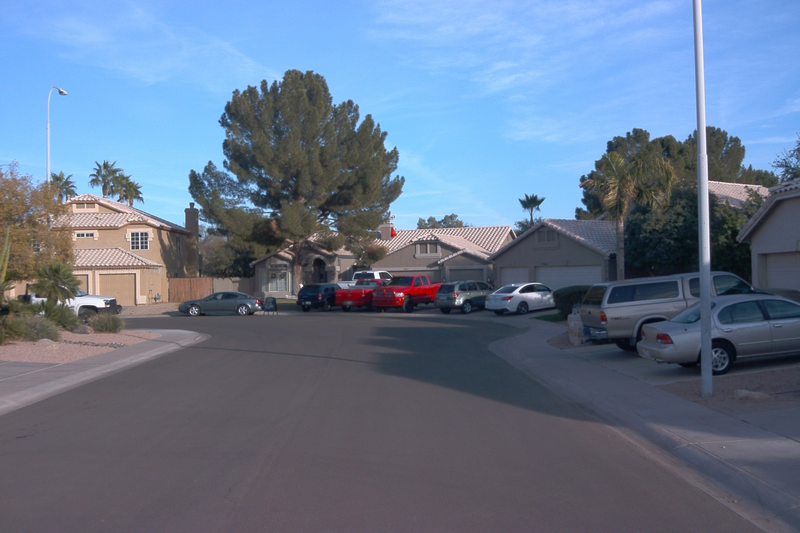}} & \frame{\includegraphics[width=0.2\textwidth]{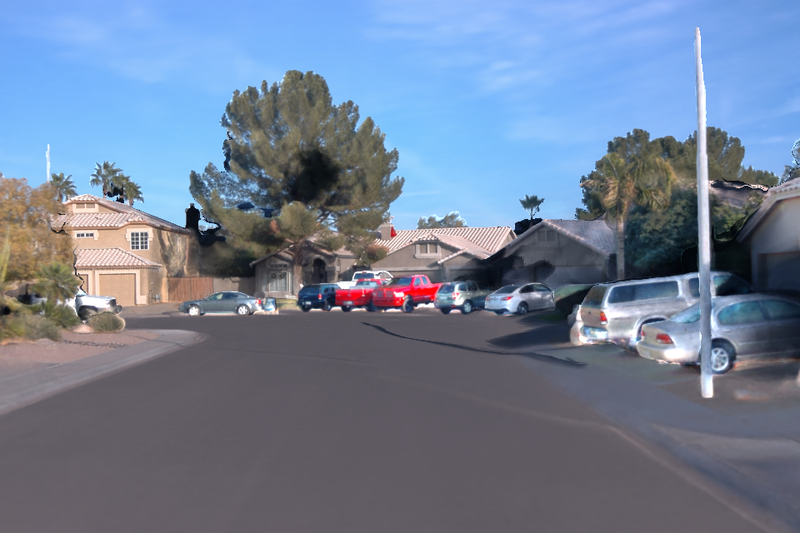}} & \frame{\includegraphics[width=0.2\textwidth]{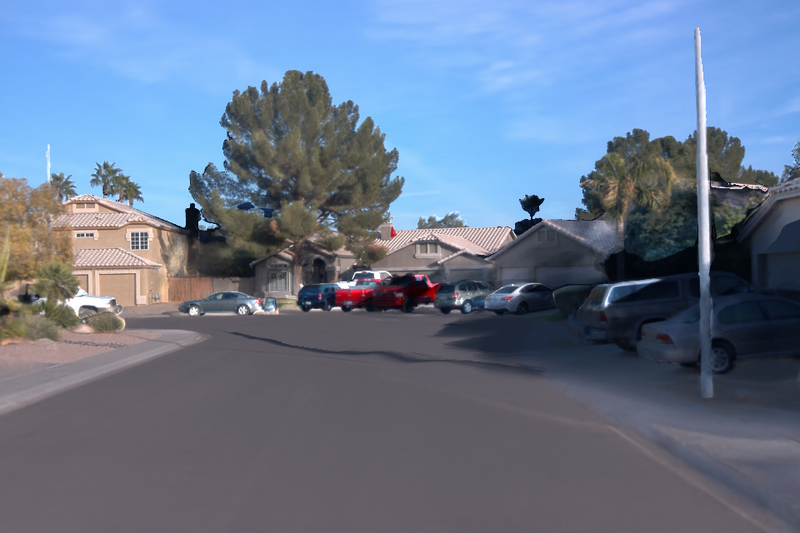}}& \frame{\includegraphics[width=0.2\textwidth]{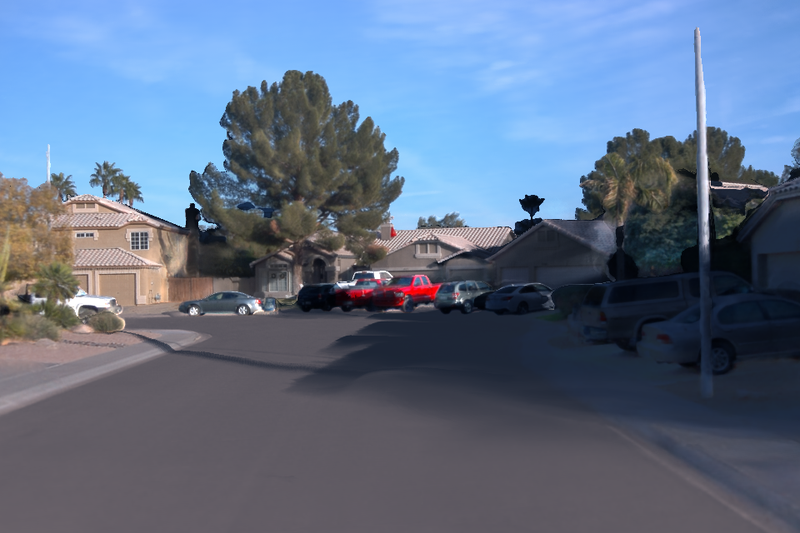}} &\frame{\includegraphics[width=0.2\textwidth]{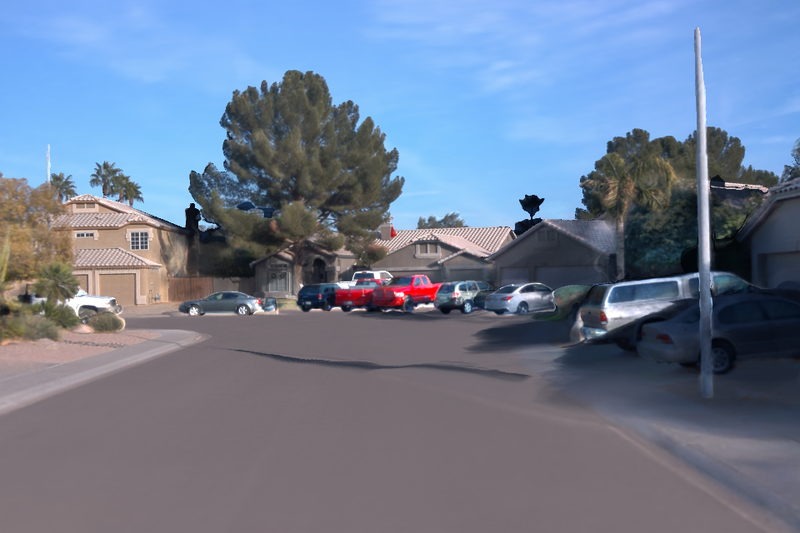}} \\

    Input &  Evening & Streetlights + Headlight  & Headlight & Single streetlight \\[0.2em]
    \frame{\includegraphics[width=0.2\textwidth]{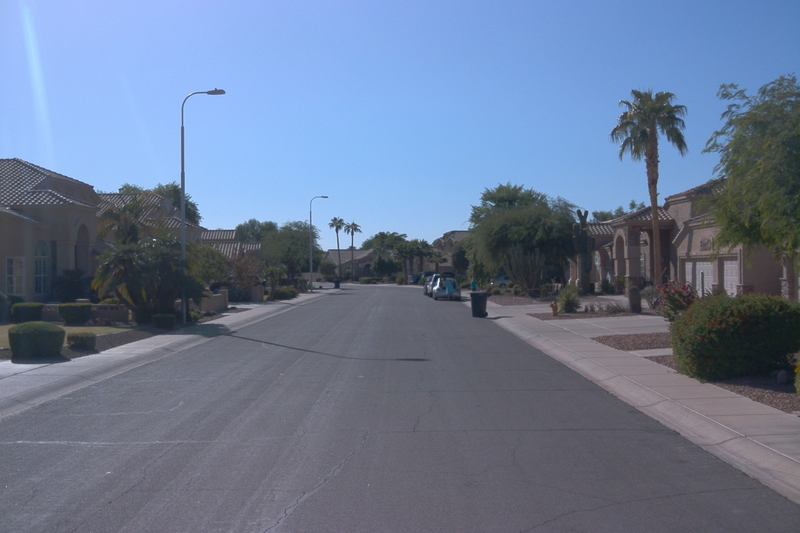}} & \frame{\includegraphics[width=0.2\textwidth]{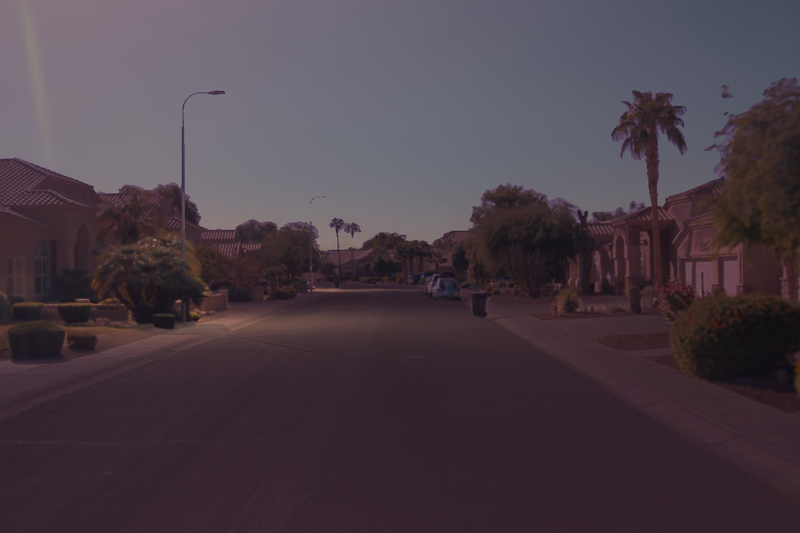}} & \frame{\includegraphics[width=0.2\textwidth]{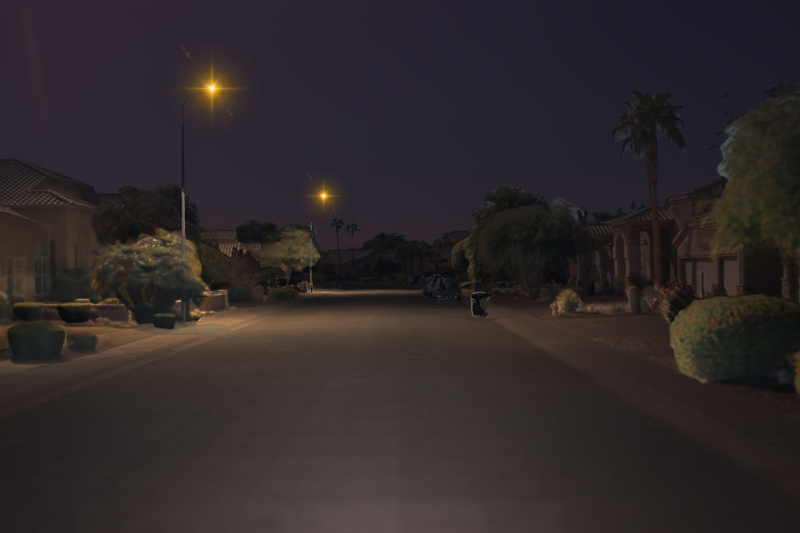}}& \frame{\includegraphics[width=0.2\textwidth]{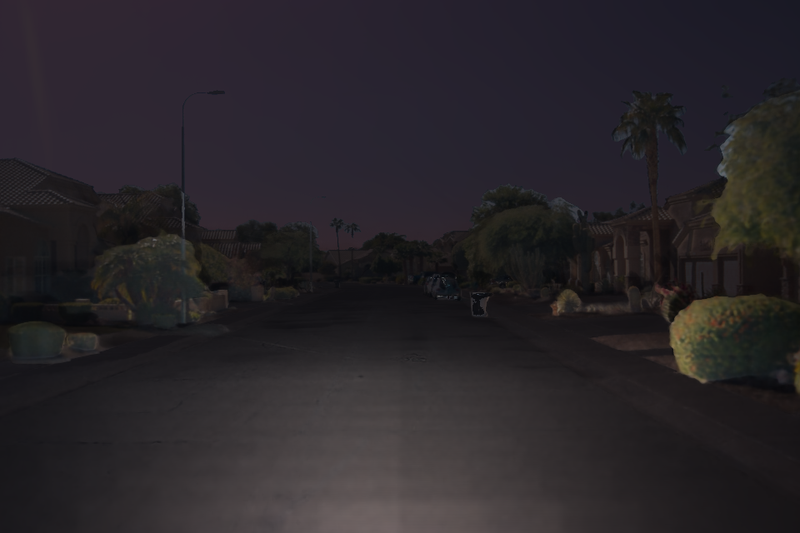}} &\frame{\includegraphics[width=0.2\textwidth]{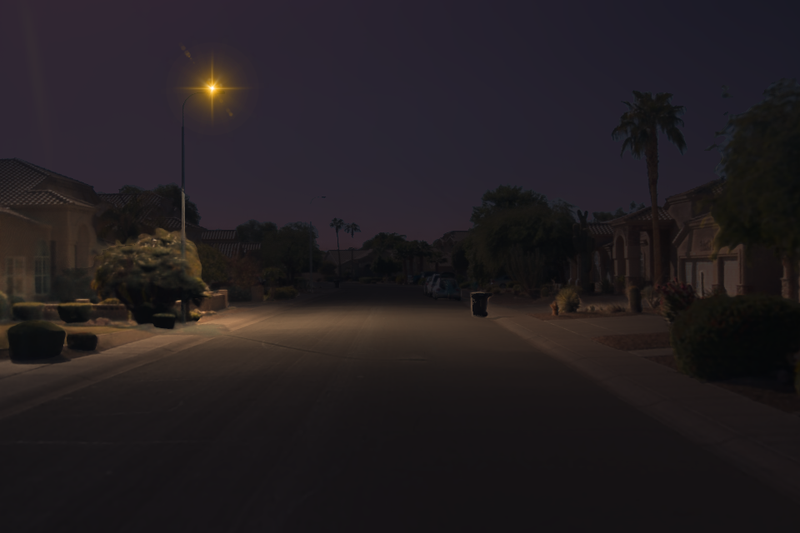}}

    \end{tabular}%
    }
    \vspace{-3mm}
    \caption{\textbf{Controllable Relighting of Waymo Open Dataset~\cite{Sun_2020_CVPR}.} The first row shows different lighting during the day, and the second row changes the input image into night-time with different lighting configurations.
    }
    \vspace{-3mm}
    \label{fig:waymo_relight}
\end{figure*}

\subsection{Baselines}
We compare \method with scene relighting and editing methods: FEGR~\cite{wang2023fegr}, 
Instruct NeRF2NeRF~\cite{instructnerf2023}, NeRF-OSR~\cite{rudnev2022nerfosr}, RelightNet~\cite{yu20relightNet}. 
 Implementation details are in the supplementary material.

\subsection{Decomposition Quality}
We evaluate intrinsic decomposition on the Waymo Open Dataset~\cite{Sun_2020_CVPR} and present the comparison in Fig~\ref{fig:waymo_intrinsic}. NeRF-OSR~\cite{rudnev2021neural} requires multi-illumination as input and fails to decompose albedo and shadow, leaving severe artifacts due to noisy normal estimation. FEGR~\cite{wang2023fegr} uses five cameras and LiDAR for reconstruction but still bakes shadow patterns into the albedo and normal. However, \method only requires a single camera as input without any LiDAR information. Integrating monocular prior in optimization successfully decomposes clean albedo, normal, and shadow maps under single illumination.

We also compare with NeRF-OSR~\cite{rudnev2022nerfosr} and RelightNet~\cite{yu20relightNet} on KITTI-360~\cite{Liao2021ARXIV} in Fig.~\ref{tab:decompose_compare}.
NeRF-OSR reconstructs a noisy normal map and cannot capture the scene shadows from a single lighting condition, leaving dark shadow patterns in the albedo. RelightNet predicts better normals but still bakes shadows into the albedo. 
\method generates clean and sharp albedo and normal fields and also produces a geometry-aware shadow from the input video sequence.  In Fig.~\ref{tab:deshadow}, we compare the learned albedo with the output of shadow removal network~\cite{guo2023shadowformer}. 
ShadowFormer~\cite{guo2023shadowformer} recovers albedo well on the ground but cannot estimate the correct albedo for the building and
vehicles. Our optimization process uses albedo regularization ($\mathcal{L}_{\text{reg}}$). This helps \method recover a cleaner albedo field on most surfaces.

\begin{figure}[t]
    \centering
    \setlength\tabcolsep{0.1em}
    \resizebox{1.0\linewidth}{!}{%
    \begin{tabular}{cccc}
    
    GT & Albedo & Normal  & Relighting \\[0.2em]
    
    \frame{\includegraphics[width=0.6\linewidth]{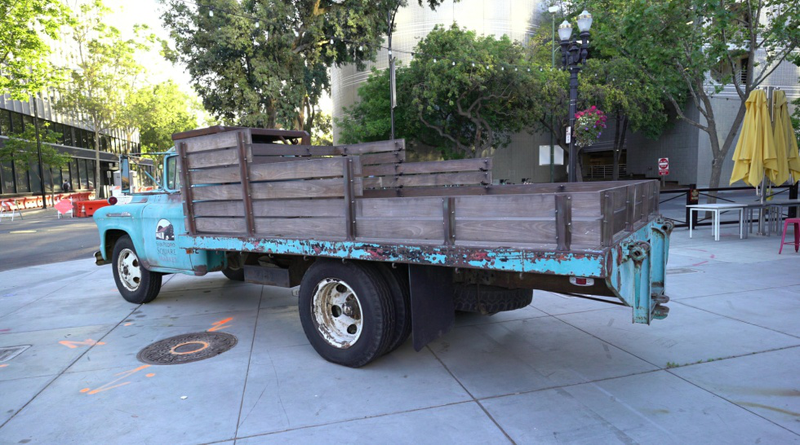}} & \frame{\includegraphics[width=0.6\linewidth]{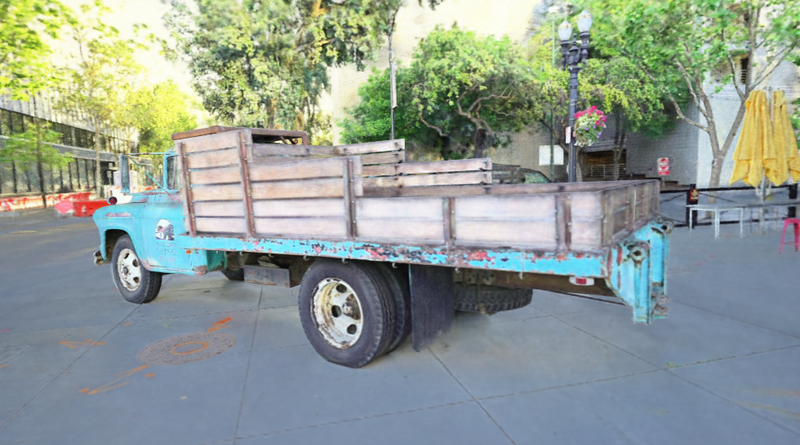}} &\frame{\includegraphics[width=0.6\linewidth]{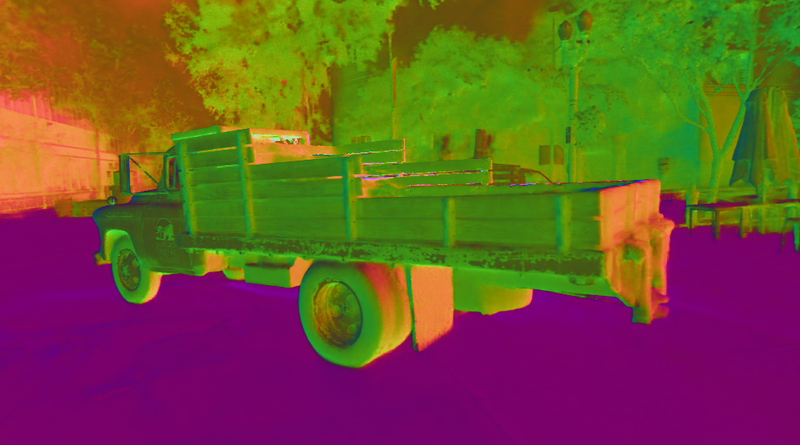}}  &\frame{\includegraphics[width=0.6\linewidth]{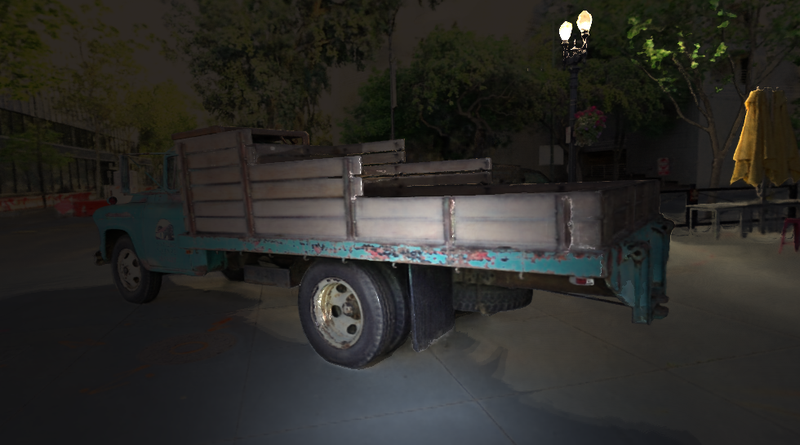}} \\

    \frame{\includegraphics[width=0.6\linewidth]{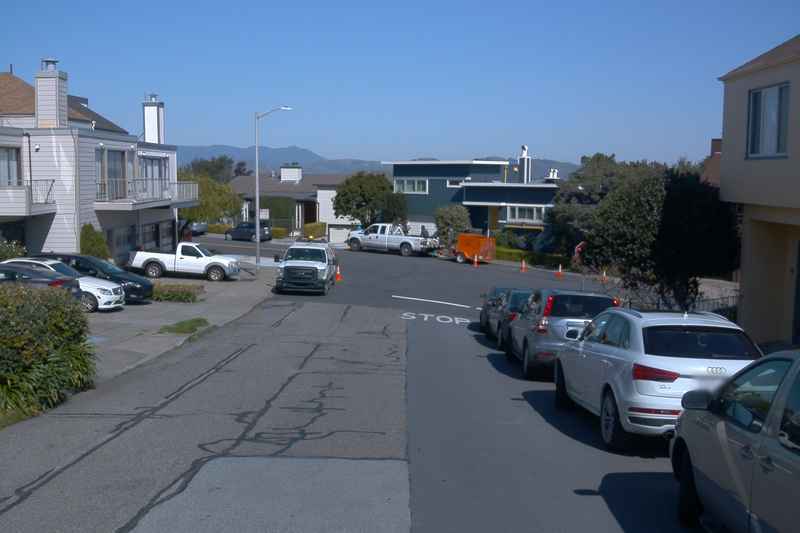}} & \frame{\includegraphics[width=0.6\linewidth]{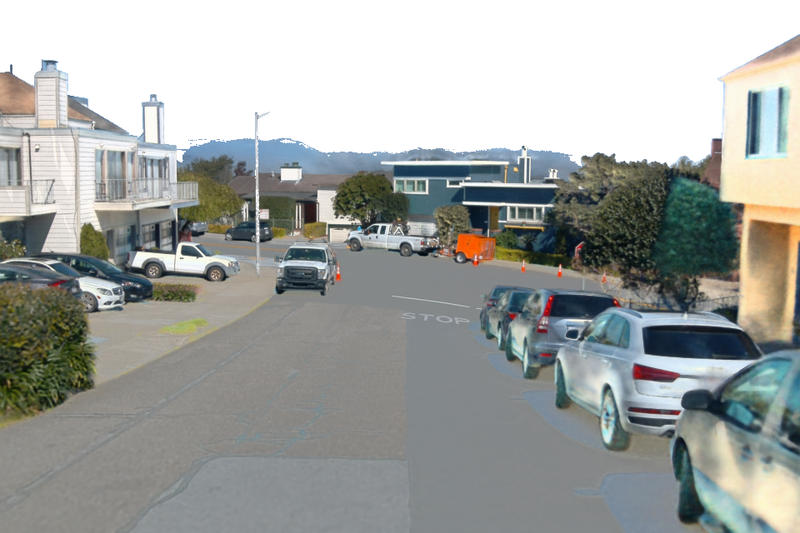}} &\frame{\includegraphics[width=0.6\linewidth]{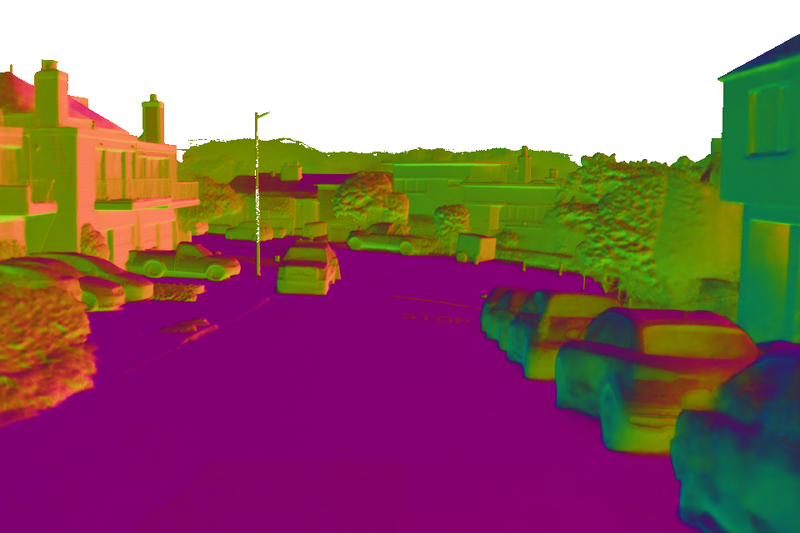}} &\frame{\includegraphics[width=0.6\linewidth]{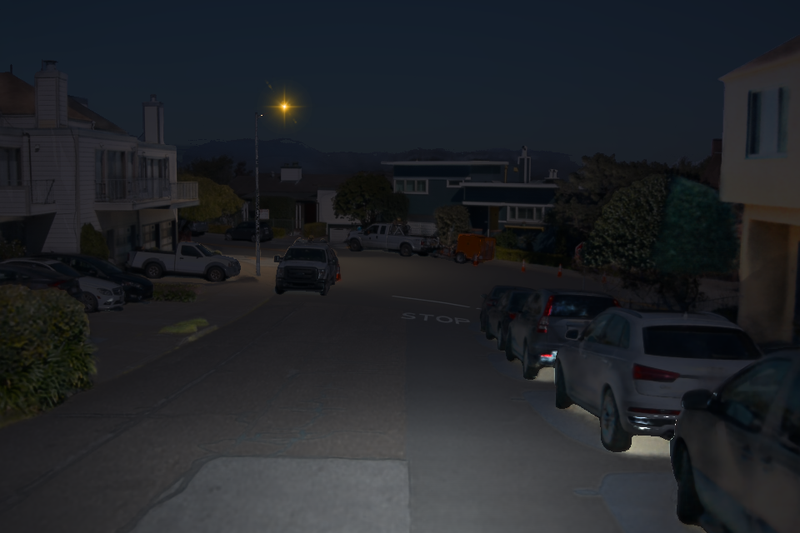}} \\
    \frame{\includegraphics[width=0.6\linewidth]{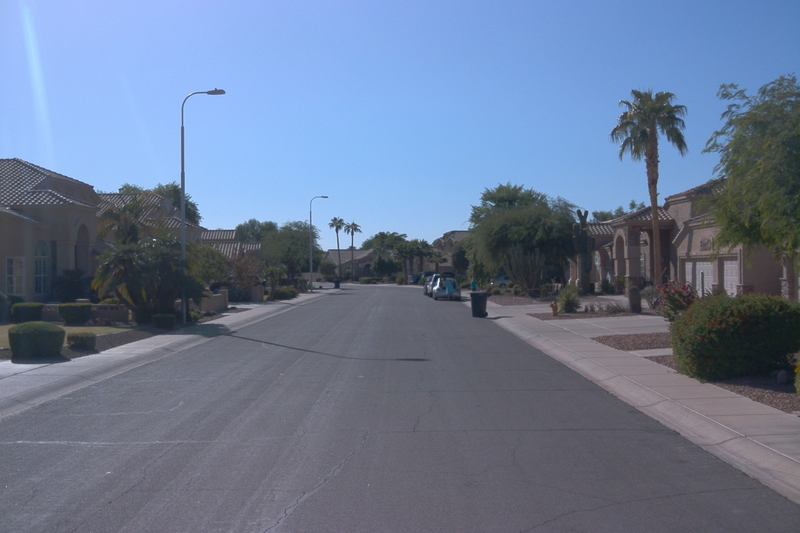}} & \frame{\includegraphics[width=0.6\linewidth]{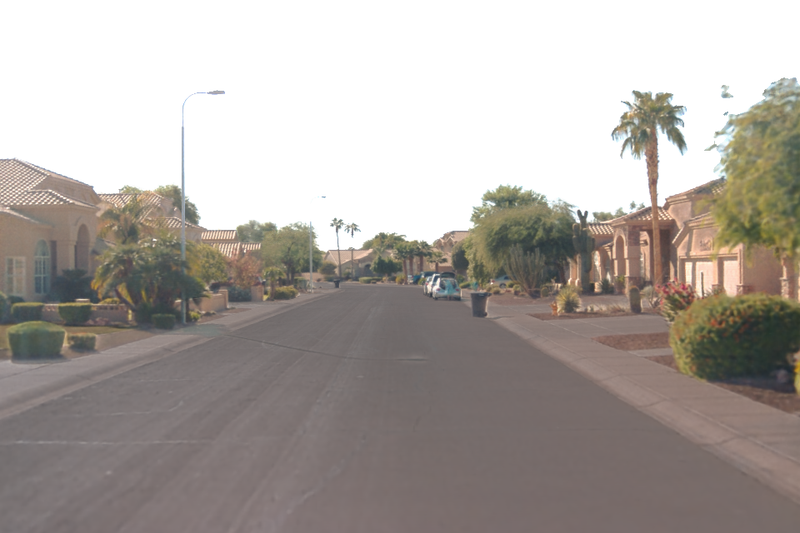}} &\frame{\includegraphics[width=0.6\linewidth]{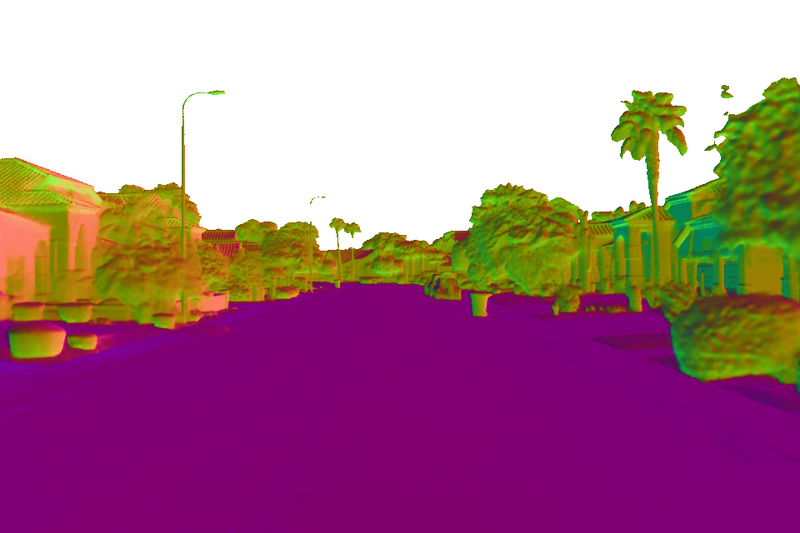}} &\frame{\includegraphics[width=0.6\linewidth]{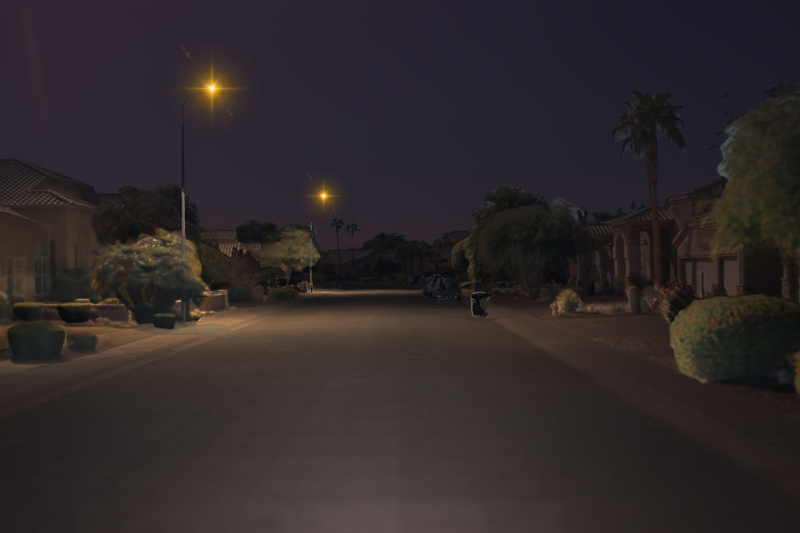}}

    \end{tabular}%
    }
    \vspace{-3mm}
    \caption{\textbf{Decomposition and relighting results of Tanks and Temples~\cite{Knapitsch2017} and Waymo Open Dataset~\cite{Sun_2020_CVPR}.}}
    \vspace{-3mm}
    \label{fig:truck}
\end{figure}

\subsection{Relighting Quality}
Relighting under various lighting conditions is evaluated in Fig.~\ref{tab:relight_night},~\ref{tab:relight_compare}. 
NeRF-OSR~\cite{rudnev2022nerfosr} cannot simulate shadows under novel light conditions. 
Instruct-NeRF2NeRF~\cite{instructnerf2023} leverages generative model~\cite{brooks2023instructpix2pix} to update the training views with text prompt and edits the neural field gradually. While it makes the overall color darker for night simulation, it fails to remove existing shadows and add new light sources.

In contrast, \method synthesizes sharp shadows and varying surface shading following the sun's direction. Further, the original scene shadows are largely absent. This allows synthesizing images at night (Fig.~\ref{tab:relight_night}) by inserting car headlights and streetlights, without distracting effects from the original shadows. Moreover, the relighting results obtained from \method are highly controllable, as demonstrated in Fig.~\ref{fig:waymo_relight}. Different light directions and intensities were used to adjust the relighting outcomes. Light sources were also added and turned on and off. \method~not only handles driving sequences but also performs well on multi-view datasets such as Tanks and Temples~\cite{Knapitsch2017}. In Fig.~\ref{fig:truck}, our method estimates accurate albedo and normal and simulates realistic nighttime images by inserting streetlights into the scene, showing that our method can generalize to diverse scenes and camera trajectories.

\begin{table*}[t]
    \centering
    \setlength\tabcolsep{0.1em} %
    \resizebox{\textwidth}{!}{
    \begin{tabular}{lccc}

        \raisebox{5mm}[0pt][0pt]{\rotatebox[origin=c]{90}{{\footnotesize{$-\mathcal{L}_{\text{visibility}}$}}}} & \frame{\includegraphics[width=0.32\textwidth]{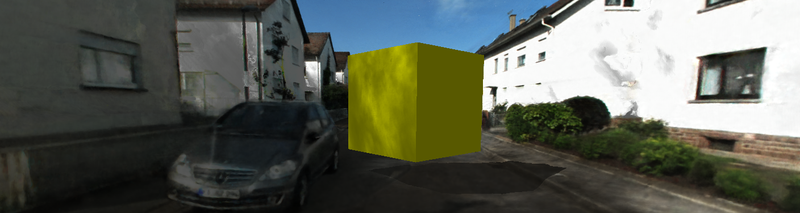}} & \frame{\includegraphics[width=0.32\textwidth]{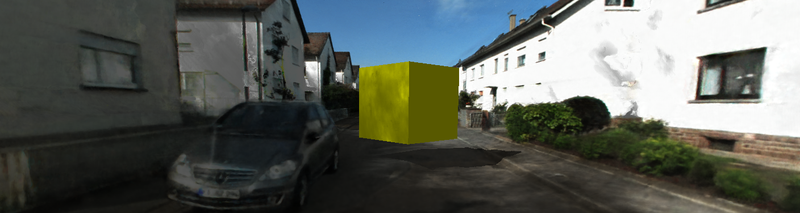}} & \frame{\includegraphics[width=0.32\textwidth]{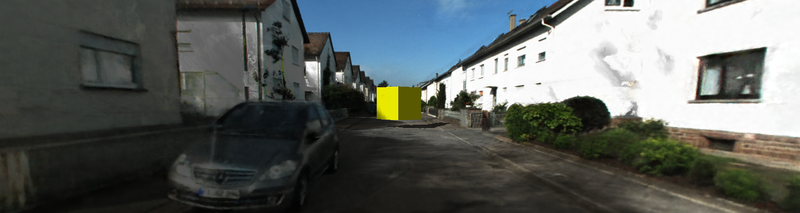}} \\

        \raisebox{7mm}[0pt][0pt]{\rotatebox[origin=c]{90}{{\footnotesize{Ours}}}}& \frame{\includegraphics[width=0.32\textwidth]{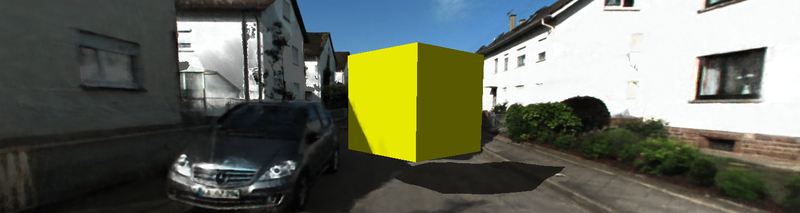}} & \frame{\includegraphics[width=0.32\textwidth]{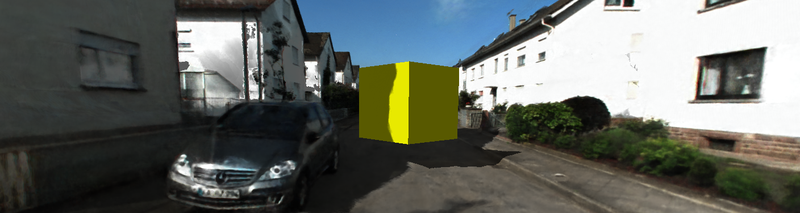}} & \frame{\includegraphics[width=0.32\textwidth]{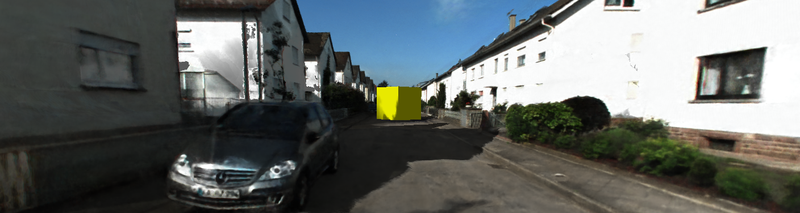}} \\
       
    \end{tabular}
    }
    \vspace{-5mm}
    \captionof{figure}{\textbf{Dynamic Object Insertion with Shadow Volume.} We insert a simple object (yellow cube) into the scene and move it along the road for evaluating object insertion. Without visibility loss, the geometry of the unseen region is noisy and casts wrong shadows. In contrast, our full model recovers geometry and produces accurate estimates of shadow according to the inserted object position.
    }
    \label{fig:insert_compare}
    \vspace{-12pt}
\end{table*}

\begin{table}[t]
    \centering
    \setlength\tabcolsep{0.1em}
    \resizebox{1.0\linewidth}{!}{%
    \begin{tabular}{lccc|ccc}
    \hline
    & \multicolumn{3}{c}{Novel View Synthesis} & \multicolumn{3}{c}{NVS + Novel light} \\
    
         & PSNR $\uparrow$ & SSIM$\uparrow$ & LPIPS$\downarrow$ & PSNR $\uparrow$ & SSIM$\uparrow$ & LPIPS$\downarrow$ \\
    \hline 
        NeRF-OSR~\cite{rudnev2022nerfosr} & 18.66 & 0.527 & 0.388 & 12.49 & 0.543 & 0.459 \\
        Instruct-N2N~\cite{instructnerf2023} & 20.55 & 0.688 & 0.169 & 13.93 & \textbf{0.707} & 0.320 \\
        UrbanIR (Ours) & \textbf{22.95} & \textbf{0.796} & \textbf{0.135} & \textbf{17.43} & 0.683 & \textbf{0.218}\\
    \hline
    \end{tabular}
    }
    \caption{\textbf{Quantitative evaluation.} We evaluate novel view synthesis (NVS) on KITTI-360~\cite{Liao2021ARXIV} and evaluate NVS + Novel light on the real-world outdoor data.}
    \vspace{-5pt}
    \label{tab:quantitative}
\end{table}

\begin{table}[t]
    \centering
    \setlength\tabcolsep{0.1em}
    \resizebox{1.0\linewidth}{!}{%
    \begin{tabular}{cccc}
GT & Ours Recon & Ours Relighting & NeRF-OSR Relighting \\
    \frame{\includegraphics[width=0.4\linewidth]{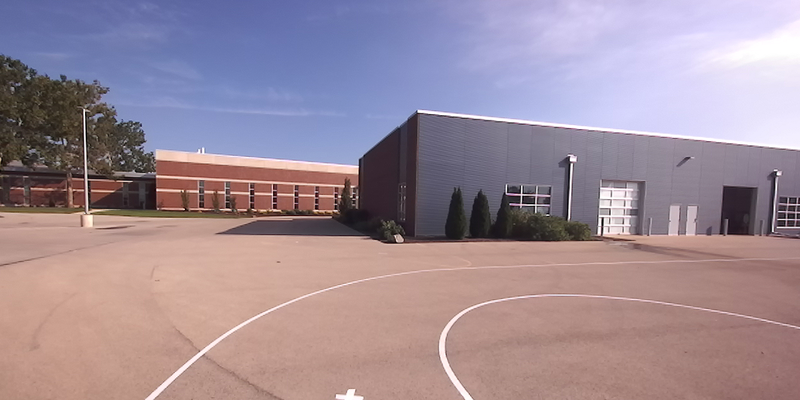}} & \frame{\includegraphics[width=0.4\linewidth]{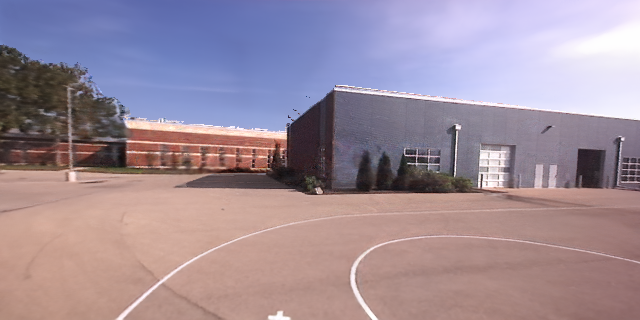}} & \frame{\includegraphics[width=0.4\linewidth]{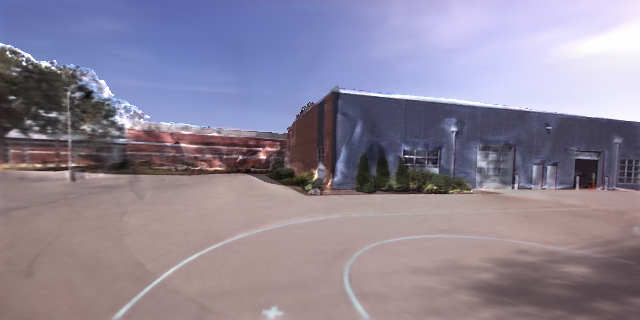}} & \frame{\includegraphics[width=0.4\linewidth]{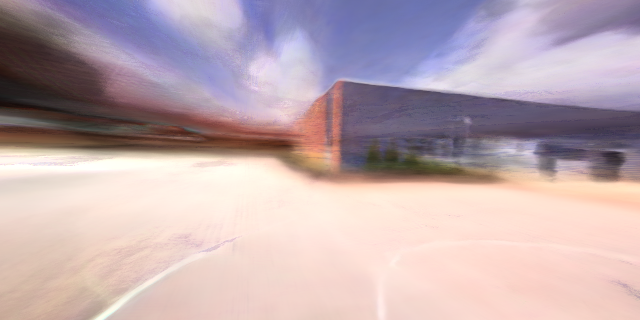}} \\

    \footnotesize\textcolor{blue}{A} Ground Truth  (9am) & \footnotesize\textcolor{blue}{A} Model + \textcolor{blue}{A} Light& \footnotesize\textcolor{red}{B} Model + \textcolor{blue}{A} Light & \footnotesize\textcolor{red}{B} Model + \textcolor{blue}{A} Light\\
    
    \frame{\includegraphics[width=0.4\linewidth]{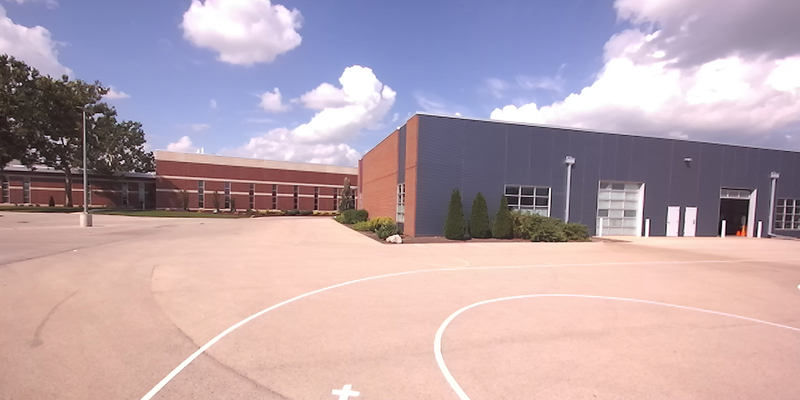}} & \frame{\includegraphics[width=0.4\linewidth]{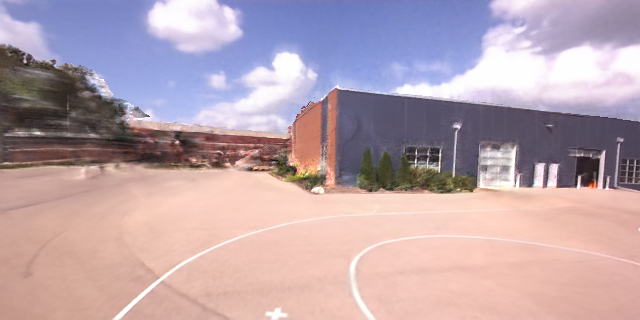}} & \frame{\includegraphics[width=0.4\linewidth]{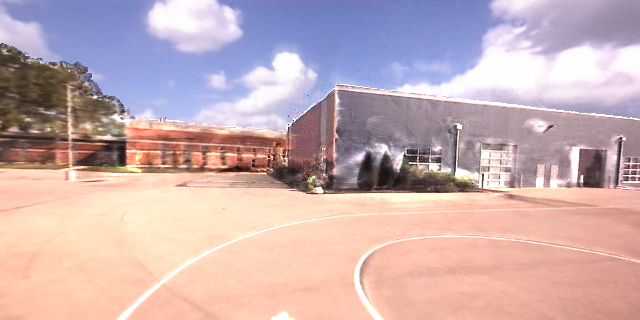}} & \frame{\includegraphics[width=0.4\linewidth]{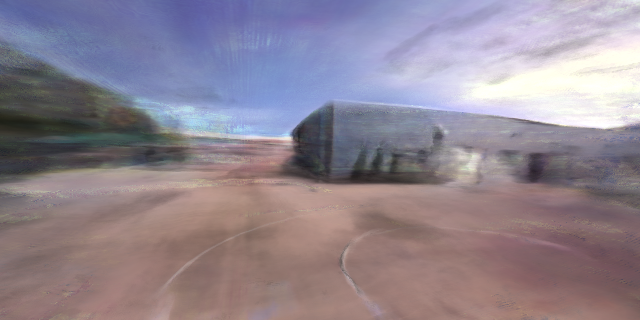}}\\

    \footnotesize\textcolor{red}{B} Ground Truth (3pm) & \footnotesize\textcolor{red}{B} Model + \textcolor{red}{B} Light& \footnotesize\textcolor{blue}{A} Model + \textcolor{red}{B} Light & \footnotesize\textcolor{blue}{A} Model + \textcolor{red}{B} Light
      
    \end{tabular}%
    }
    \vspace{-3mm}
    \captionof{figure}{\textbf{Novel view and novel light synthesis.}}
    \vspace{-10pt}
    \label{fig:highbay}
\end{table}

\subsection{Quantitative Evaluation}
The quantitative evaluation results can be found in Tab.~\ref{tab:quantitative}. We tested the novel view synthesis on KITTI-360~\cite{Liao2021ARXIV} using 10 images as the novel views for all 7 sequences. \method outperforms baselines such as NeRF-OSR~\cite{rudnev2022nerfosr} and Instruct-NeRF2NeRF~\cite{instructnerf2023} in all metrics, indicating that our model not only decomposes intrinsic well but also produces high-quality images.  To evaluate the relighting in novel views, we captured videos of the outdoor scenes in the morning and afternoon. After individually optimizing models at both sequences, we performed relighting by exchanging lighting parameters and camera poses, and the image metrics were calculated with the ground truth capture. Our method outperformed all baselines and demonstrated the effectiveness of our intrinsic decomposition and lighting parameterization. The qualitative results can be seen in Fig.~\ref{fig:highbay}. \method was successful in removing existing shadows, changing the shading on the building, and modifying the sky texture during different times of the day. Please note that we selected and compared with the most competitive baseline methods that are open-sourced, and other methods such as FEGR~\cite{wang2023fegr} and LightSim~\cite{pun2023lightsim} do not have codebase available publicly, making it impossible to make a fair comparison with them.

\subsection{Object Insertion}

\begin{figure}[t]
    \centering
    \setlength\tabcolsep{0.1em}
    \resizebox{1.0\linewidth}{!}{%
    \begin{tabular}{@{}ccc@{}}
    
     Input & $-\mathcal{L}_{\text{visibility}}$ & Ours \\[0.2em]

    \frame{\includegraphics[width=0.3\textwidth]{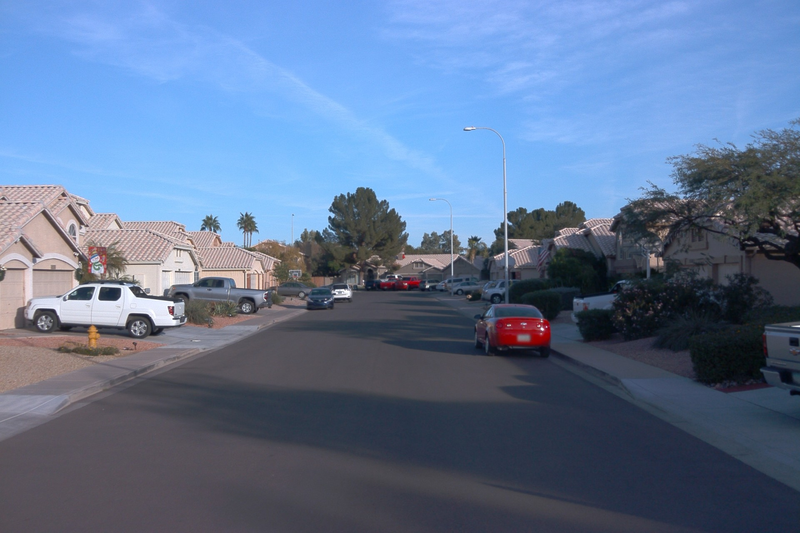}} &  \frame{\includegraphics[width=0.3\textwidth]{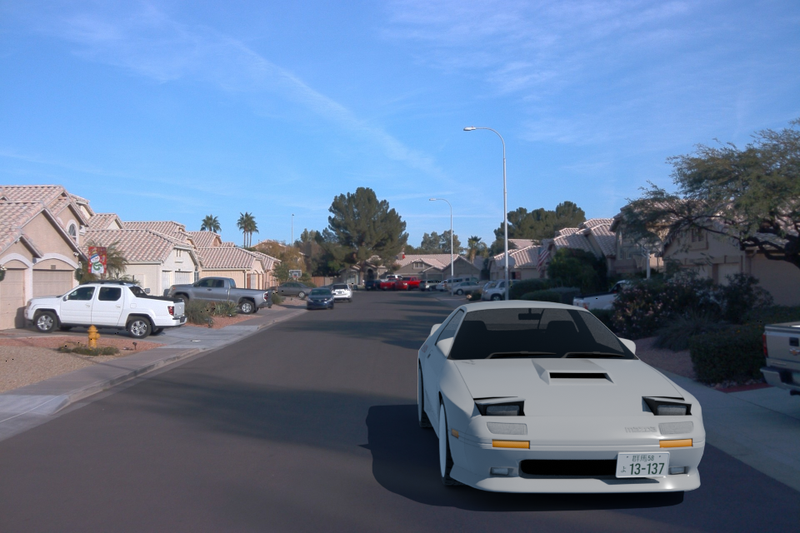}}  &  \frame{\includegraphics[width=0.3\textwidth]{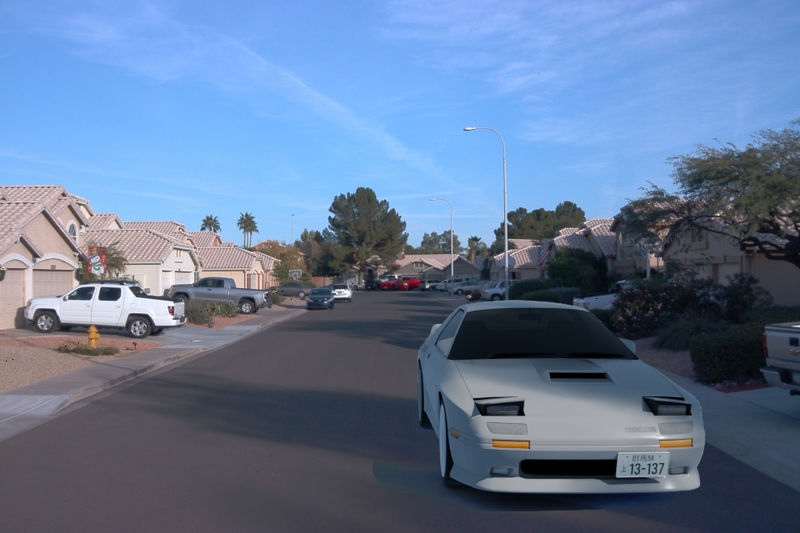}} \\

    \frame{\includegraphics[width=0.3\textwidth]{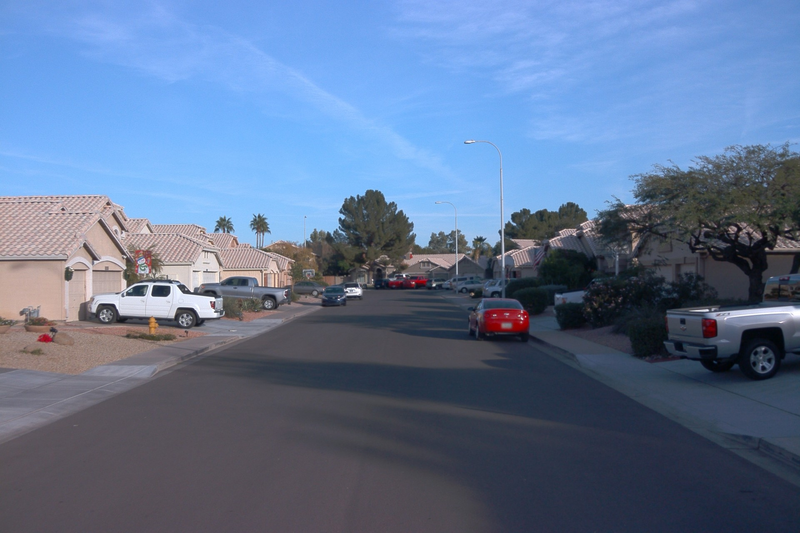}} &  \frame{\includegraphics[width=0.3\textwidth]{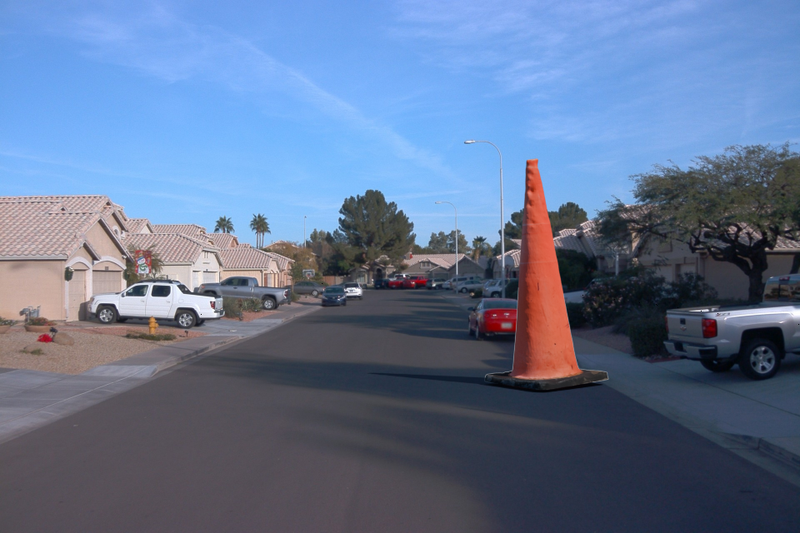}}  &  \frame{\includegraphics[width=0.3\textwidth]{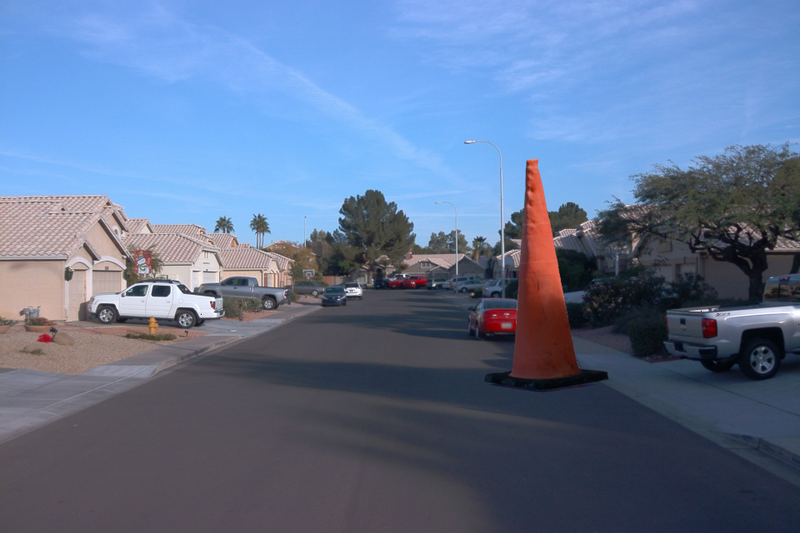}} \\

    \frame{\includegraphics[width=0.3\textwidth]{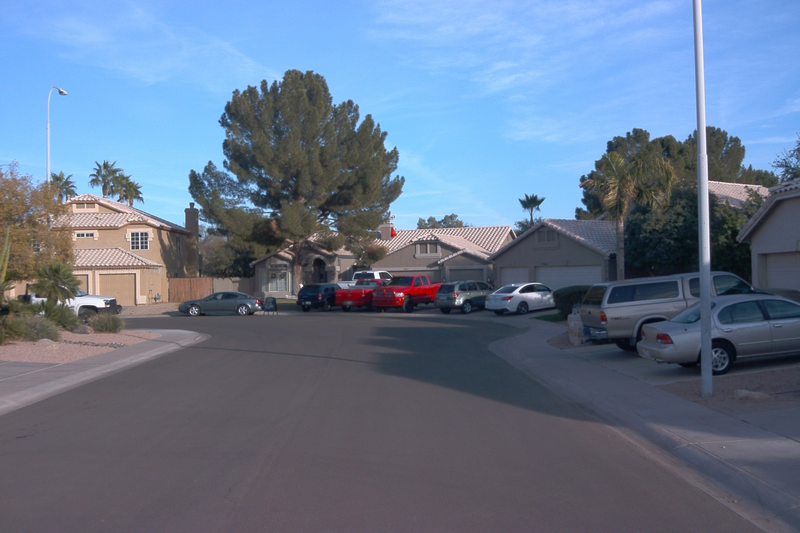}} &  \frame{\includegraphics[width=0.3\textwidth]{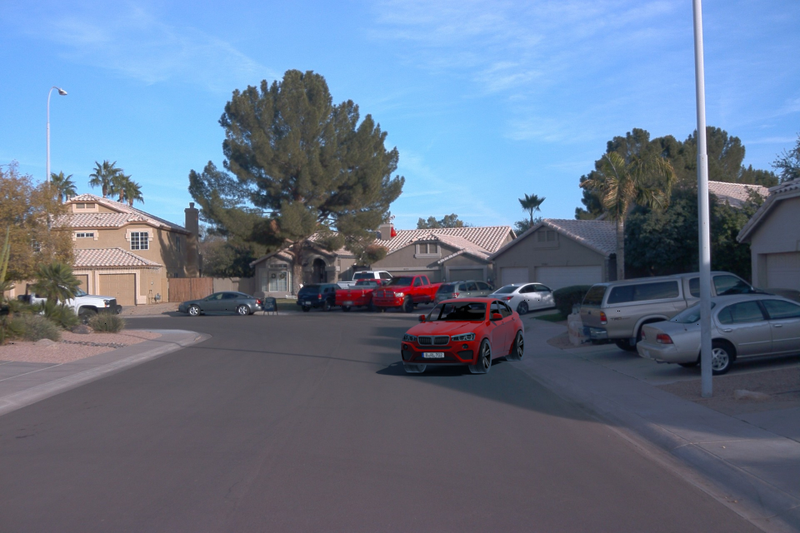}}  &  \frame{\includegraphics[width=0.3\textwidth]{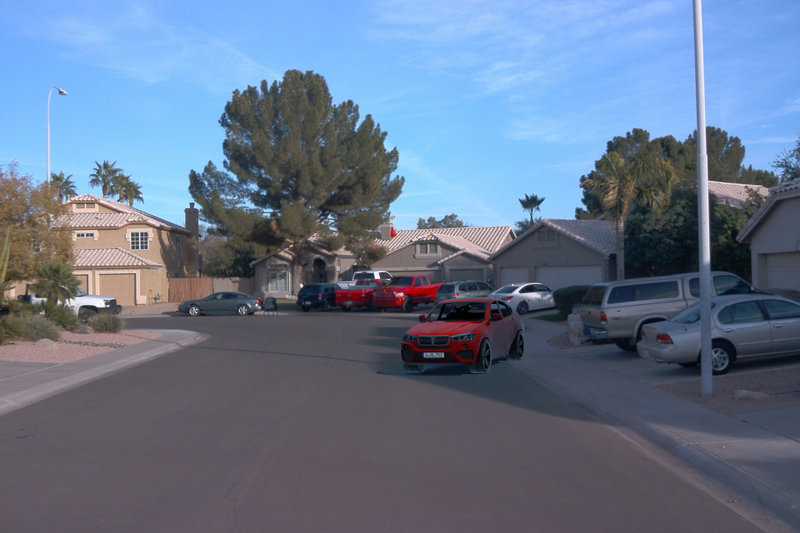}} 
    
    \end{tabular}%
    }
    \vspace{-3mm}
    \caption{\textbf{Object Insertion Qualitative Results.}Without visibility modeling (middle column), the scenes do not cast shadows on the inserted objects, and the original object shadow looks unrealistic in the existing shadow. Our full method (right column) simulates the better interaction between the reconstructed scenes and inserted objects with the help of visibility modeling.
    }
    \vspace{-5mm}
    \label{fig:insert_blender}
\end{figure}

Following~\cite{wang2023fegr, wang2022neural}, we build the object insertion pipeline with Blender~\cite{blender}, and the results are shown in Fig.~\ref{fig:insert_compare} and ~\ref{fig:insert_blender}. By tracing the rays from the object surface toward light sources (i.e. the sun), \method~estimates the visibility with volume rendering (Eq.~\ref{eq:visibility}). As a result, our full model can cast scene shadows on the inserted objects and also weaken the object shadow on the ground if it overlaps with the existing scene shadow. The visibility modeling (Sec.~\ref{eq:loss_vis}) recovers the geometry that is not captured well in the input views (e.g. building top), enabling~\method to simulate shadows better and to enhance the insertion realism significantly.

\section{Limitation and Discussion}
\label{sec:conclusion}
In this work, we investigated the task of inverse rendering of unbounded outdoor scenes under single illumination. This task is ill-posed and extremely challenging due to the sparsity of observations across space and time. To overcome this challenge and successfully decompose various scene intrinsic properties, we utilized prior knowledge such as pretrained networks and regularization to reduce the uncertainty space and improve the performance of downstream applications like relighting and object insertion. However, there are limitations. Our optimization process can be affected by the noisy predictions from prior models and requires careful tuning of our losses. Sometimes, shadows cannot be removed entirely in the albedo field, and they may still appear in the final images. Additionally, the visibility optimization refines only the geometry along the light direction, which means that large changes in the sun's direction can lead to poor shadows when the geometry estimates are not accurate.

\clearpage
\setcounter{page}{1}
\maketitlesupplementary

\begin{figure*}[t]
    \centering\setlength{\tabcolsep}{3pt}
    \resizebox{1.0\textwidth}{!}{%
    \begin{tabular}{@{}c@{}}

     \includegraphics[width=1\textwidth]{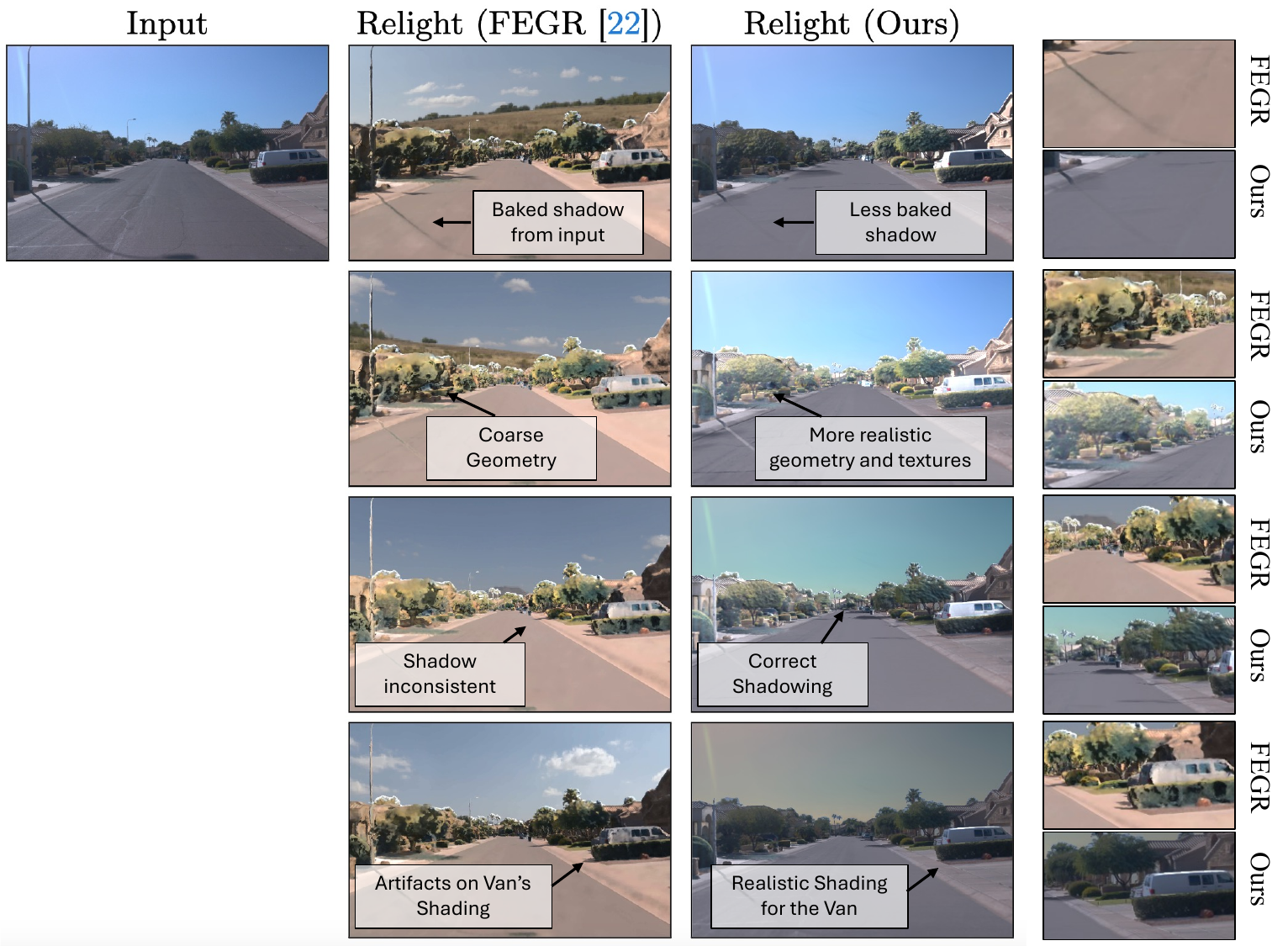} \\
    
    \end{tabular}%
    }
    \vspace{-3mm}
    \caption{\textbf{Relighting Comparison on Waymo Open Dataset~\cite{Sun_2020_CVPR}. The second and third columns compare the relighting quality. The authors provide the FEGR results and we match the lighting condition according to the shadow direction.} 
    }
    \vspace{-3mm}
    \label{fig:waymo_relight_seq3}
\end{figure*}

\section{More Qualitative Results}
We compare the relighting quality with FEGR~\cite{wang2023fegr} in Fig.~\ref{fig:waymo_relight_seq3}. FEGR~\cite{wang2023fegr} first extracts mesh and estimates the shading from the lighting configuration, and the imperfect mesh geometry produces artifacts and loses appearance details. On the other hand, our method alleviates the original shadow and produces relighting images while preserving appearance details. We show additional night simulation results on various Kitti360~\cite{Liao2021ARXIV} sequences in Fig.~\ref{tab:relight_night_supp}, demonstrating the generalization capability of \method. The Instruct-Pix2Pix~\cite{brooks2023instructpix2pix} leverages the large language model~\cite{brown2020language} and stable diffusion~\cite{rombach2022high} for abundant image editing tasks. 
However, such a data-driven method cannot move the daylight shading and shadow in the input images. 
On the contrary, \method~decomposes shadow-free albedo and performs physically-based rendering with new light sources (e.g., streetlights, headlights), significantly enhancing the visual quality of night simulation. 
The strong specular reflection is also simulated on the car region, boosting the realism of metal material. 
Please note that the simulation is flexible, and the user can adjust physical parameters (e.g., light color, light strength) to create various effects. Please refer to our supplementary videos to better visualize view consistency and controllable simulation.

\begin{table*}[t]
    \centering
    \setlength\tabcolsep{0.05em} 
    \renewcommand{\arraystretch}{0.2}%
    \resizebox{\textwidth}{!}{
    \begin{tabular}{lccc}
        \raisebox{7mm}[0pt][0pt]{\rotatebox[origin=c]{90}{\footnotesize{Input}}} & \includegraphics[width=0.32\textwidth]{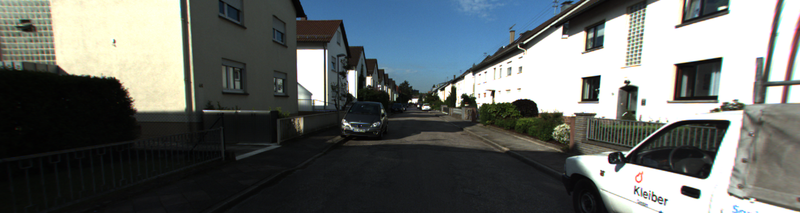} & \includegraphics[width=0.32\textwidth]{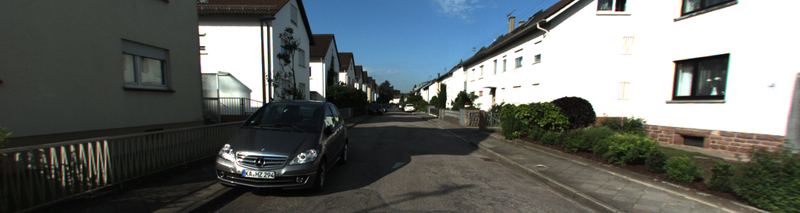} & \includegraphics[width=0.32\textwidth]{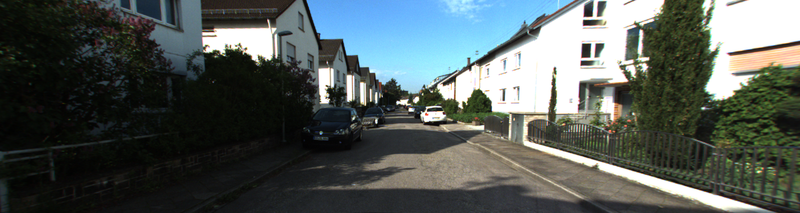} \\

        \raisebox{7mm}[0pt][0pt]{\rotatebox[origin=c]{90}{\footnotesize{I-p2p\cite{brooks2023instructpix2pix}}}} & \includegraphics[width=0.32\textwidth]{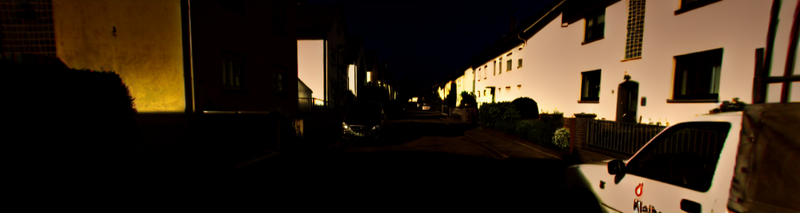} & \includegraphics[width=0.32\textwidth]{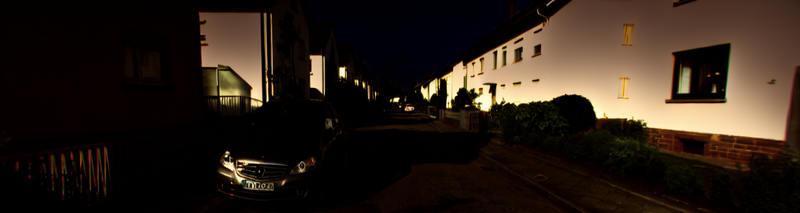} & \includegraphics[width=0.32\textwidth]{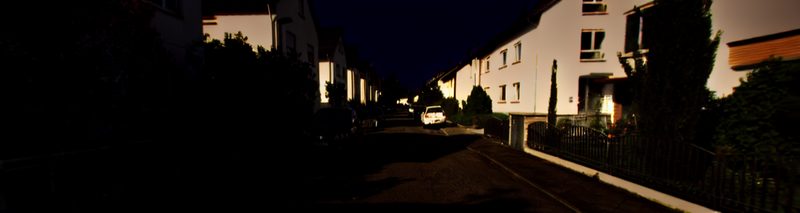} \\

        \raisebox{7mm}[0pt][0pt]{\rotatebox[origin=c]{90}{\footnotesize{Ours}}} & \includegraphics[width=0.32\textwidth]{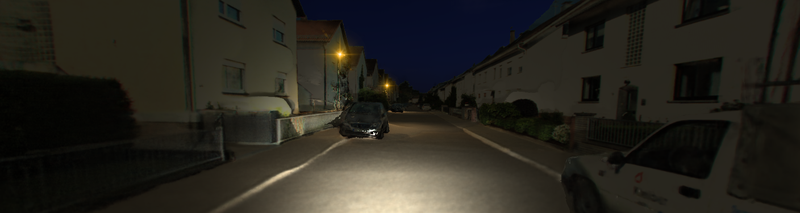} & \includegraphics[width=0.32\textwidth]{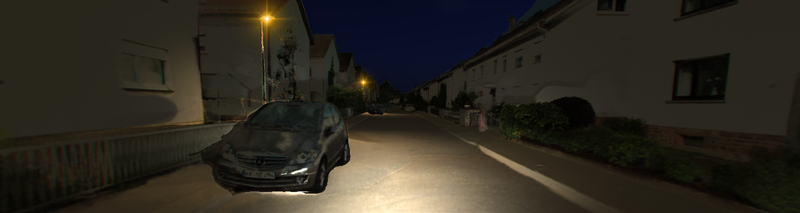} & \includegraphics[width=0.32\textwidth]{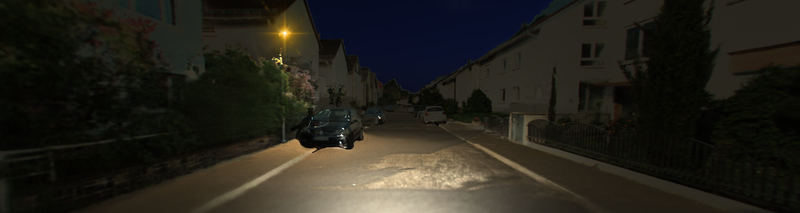} \\ \\ \\

        \raisebox{7mm}[0pt][0pt]{\rotatebox[origin=c]{90}{\footnotesize{Input}}} & \includegraphics[width=0.32\textwidth]{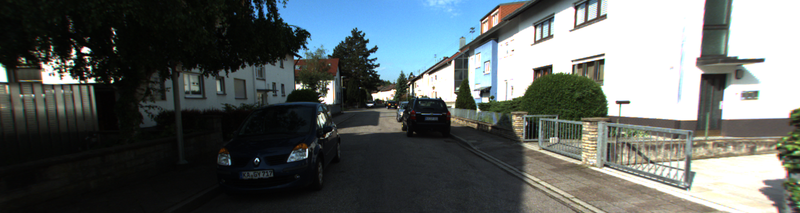} & \includegraphics[width=0.32\textwidth]{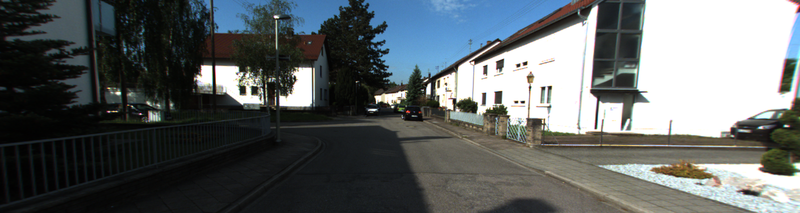} & \includegraphics[width=0.32\textwidth]{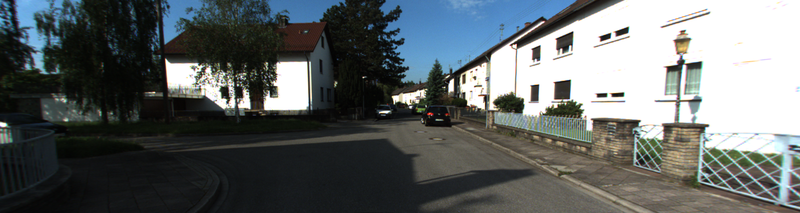} \\

        \raisebox{7mm}[0pt][0pt]{\rotatebox[origin=c]{90}{\footnotesize{I-p2p~\cite{brooks2023instructpix2pix}}}} & \includegraphics[width=0.32\textwidth]{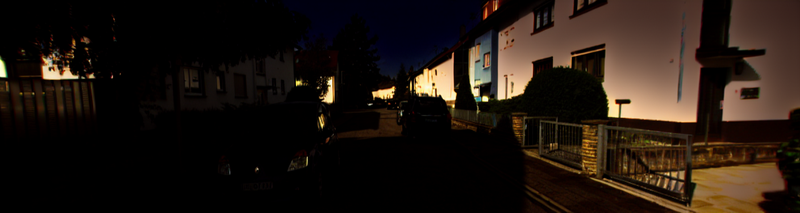} & \includegraphics[width=0.32\textwidth]{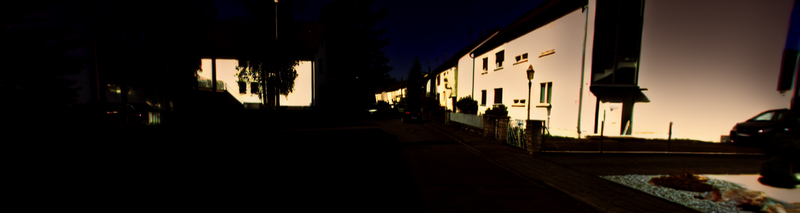} & \includegraphics[width=0.32\textwidth]{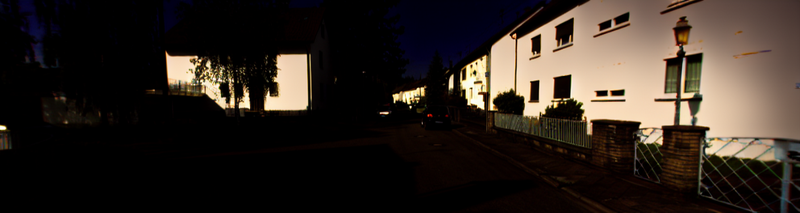} \\ 
        
        \raisebox{7mm}[0pt][0pt]{\rotatebox[origin=c]{90}{\footnotesize{Ours}}} & \includegraphics[width=0.32\textwidth]{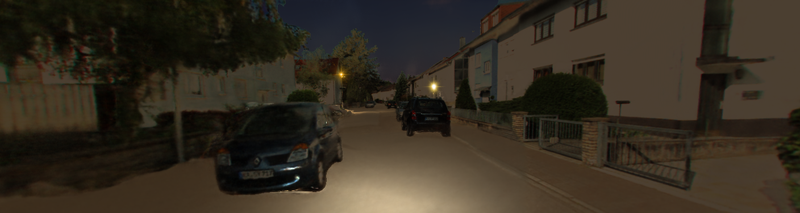} & \includegraphics[width=0.32\textwidth]{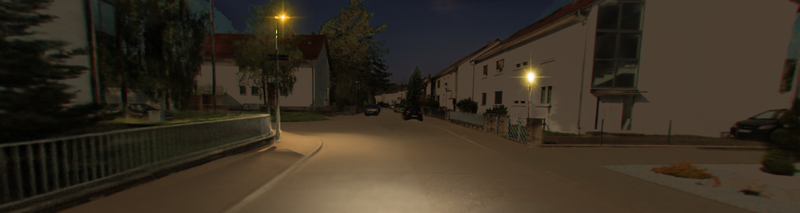} & \includegraphics[width=0.32\textwidth]{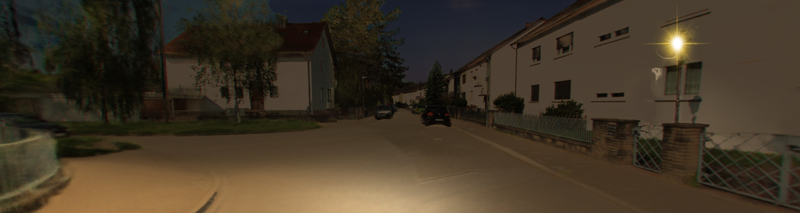} \\ \\ \\ 

        \raisebox{7mm}[0pt][0pt]{\rotatebox[origin=c]{90}{\footnotesize{Input}}} & \includegraphics[width=0.32\textwidth]{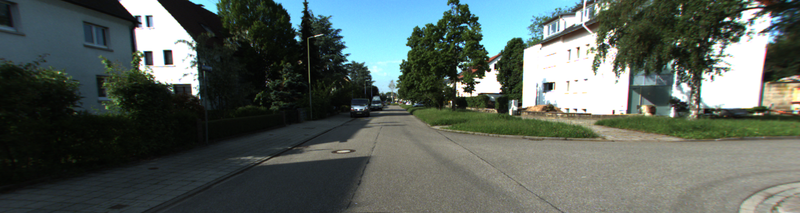} & \includegraphics[width=0.32\textwidth]{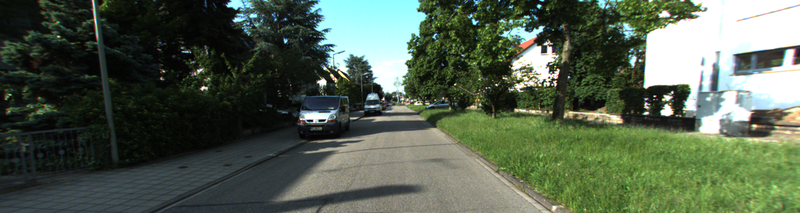} & \includegraphics[width=0.32\textwidth]{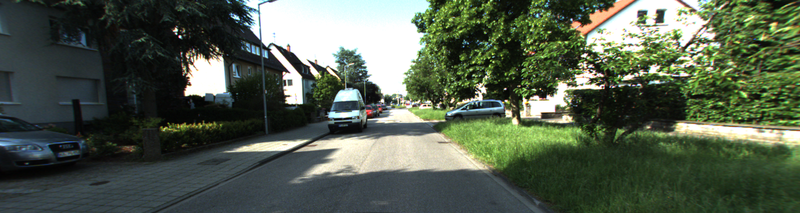} \\

        \raisebox{7mm}[0pt][0pt]{\rotatebox[origin=c]{90}{\footnotesize{I-p2p~\cite{brooks2023instructpix2pix}}}} & \includegraphics[width=0.32\textwidth]{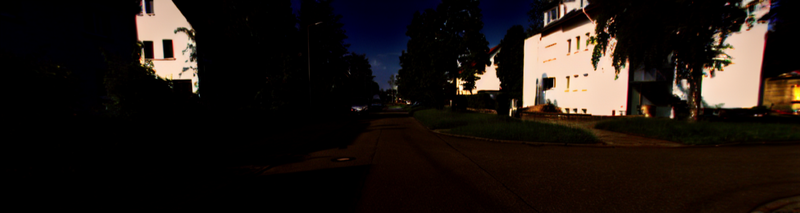} & \includegraphics[width=0.32\textwidth]{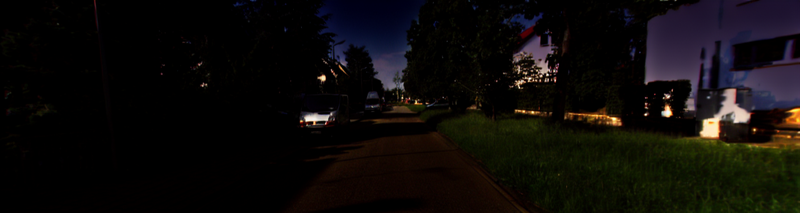} & \includegraphics[width=0.32\textwidth]{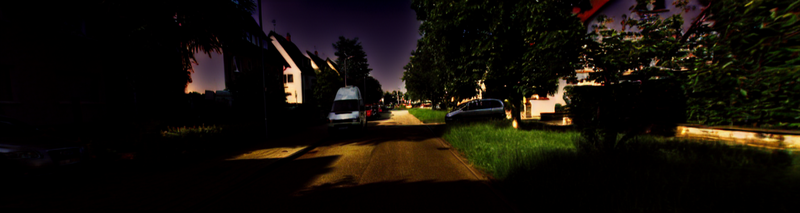} \\ 
        
        \raisebox{7mm}[0pt][0pt]{\rotatebox[origin=c]{90}{\footnotesize{Ours}}} & \includegraphics[width=0.32\textwidth]{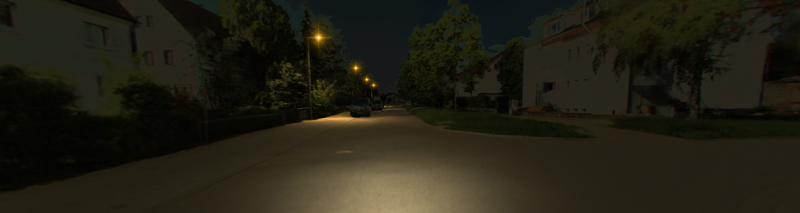} & \includegraphics[width=0.32\textwidth]{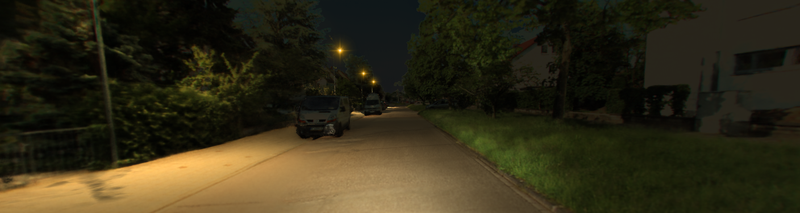} & \includegraphics[width=0.32\textwidth]{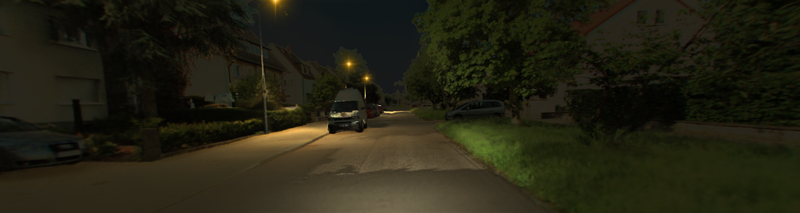}

    \end{tabular}
    }
    \captionof{figure}{{\bf Nighttime rendering.} The scene is transformed from daytime (1st row) to night-time (3rd row) by introducing new light sources: a headlight on a car and a street lamp. Top 3 and bottom 3 rows are from same driving sequence with different time stamp. Comparing with data-driven generative model and Instruct-Pix2Pix~\cite{brooks2023instructpix2pix}, the dark shadows with sharp boundaries are successfully removed with our decomposition, resulting more realistic rendering with new light sources (e.g. streetlights, headlight) during the nighttime simulation.}
    \label{tab:relight_night_supp}
\end{table*}

\section{Model Architecture}
Instant-NGP~\cite{muller2022instant} encodes the scene with a multi-scale hash table, and each entry contains learnable parameters. 
For point $\bx \in \bbR^3$, the model retrieves and interpolates the parameters with hash function: $F(\bx, \theta)$. \method~adopts the hash encoding from~\cite{muller2022instant} and maintain two separate hash tables for geometry and appearance, and predict the scene properties with:
\begin{align}
\begin{split}
    \sigma &= F_{g}(\bx, \theta_g)\\ 
    (\ba, \bn, s) &= F_a(\bx, \theta_a),
\end{split}
\end{align}
where $\sigma$ is density, $(\ba, \bn, s)$ are albedo, surface normal, and semantic. 
$\theta_g$, $\theta_a$ are learnable parameters for geometry and appearance. 
Please note that the density field $\sigma$ is not only involved in the volume rendering (Eq.~\ref{eq:volume_rendering}), but also involved in visibility estimation (Eq.~\ref{eq:visibility}) and normal loss calculation. 
The hash encoding is implemented with tiny-cuda-nn~\cite{tiny-cuda-nn}. 
We empirically find that maintaining separate learnable parameters for geometry and appearance leads to more stable convergence and higher rendering quality.

\begin{figure*}[ht]
    \centering
    \includegraphics[width=0.9\textwidth]{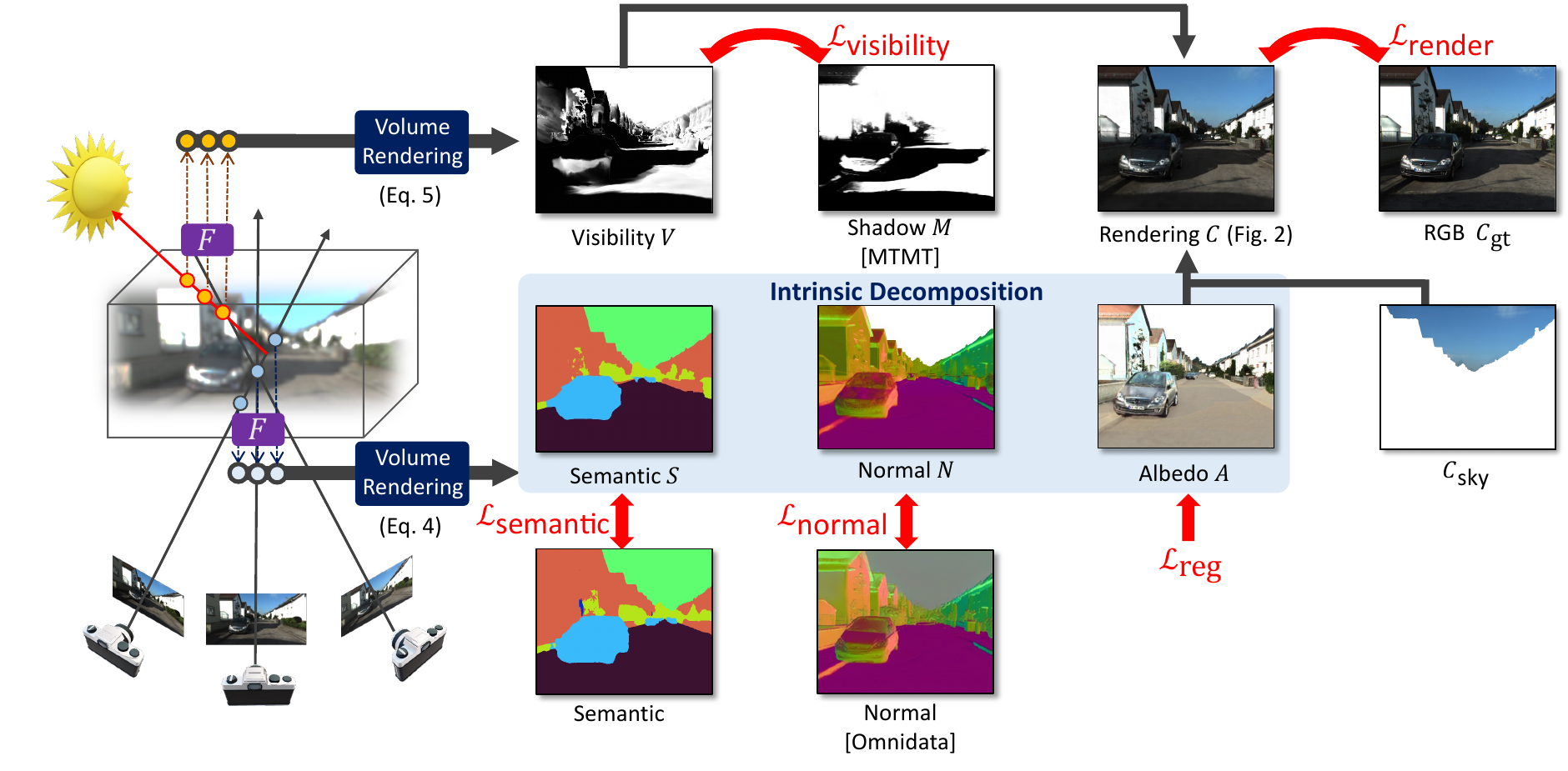}
    \caption{\textbf{Training Pipeline.} 
    \method retrieves scene intrinsics with volume rendering from camera rays, which is guided by semantic and normal priors. 
    Transmittance along tracing rays is supervised with shadow masks.
    }
    \label{fig:training}
\end{figure*}

\section{Training Details}
The training procedure is illustrated in Fig.~\ref{fig:training}. We leverage pretrained networks as 2D priors during training to address the ill-posed inverse problem. 
Specifically, the shadow mask is estimated with MTMT~\cite{chen20MTMT}. 
Omnidata normal estimation~\cite{eftekhar2021omnidata} helps refine scene geometry, which is critical in the shading quality and albedo decomposition. 
A semantic map is provided in Kitti360 dataset~\cite{Liao2021ARXIV} and can also be estimated with MMSegmentation~\cite{mmseg2020} if such information is not provided. 
The objective function of the optimization is:
\begin{equation*}
 \min_{\theta, \bL} \mathcal{L}_{\text{render}} + \lambda_1\mathcal{L}_{\text{visibility}} + \lambda_2\mathcal{L}_{\text{normal}} + \lambda_3\mathcal{L}_{\text{semantics}} + \lambda_4\mathcal{L}_{\text{reg}},
\end{equation*}
where $\lambda_{\text{1}}=0.001, \lambda_{\text{2}}=0.01, \lambda_{\text{3}}=0.04, \lambda_{\text{4}}=0.1$. We use Adam optimizer~\cite{kingma2014adam} with a learning rate of $0.002$ for a total of 100 epochs during the optimization.

\section{Application Details}
We provide the implementation of relighting and object insertion as follows:

\noindent\emph{Simulating night-time} proceeds by defining %
headlights and street lights, then illuminating with scene model considering specularity and lens flare. 
For sky regions $\bS(\br)\in \text{sky}$, we use
  $\bC(\br) =  \bL_\textrm{sky}(\br)$
and otherwise, we use
\begin{equation}
\label{eq:shading_night}
    \bA(\br) \left (\sum \bL_{\text{dif}}^{\text{i}} \bD_i \bV_i + \bL_{\text{amb}} \right) + \sum_i \bL^i_{\text{spec}}
\end{equation}
The spotlight we used is given by the center $\bo_L^i \in \bbR^3$ and direction $\bd_L^i \in
\bbR^3$ of the light.  
This spotlight produces a diffuse radiance at $\br$ given by
\begin{equation}
    \bL^i_{\text{dif}}(\br) =  \frac{1}{\|\bo^i_L - \bx(\br)\|^2}\left (l \cdot \bd^i_L\right)^k, l = \frac{\bo_L^i - \bx(\br)}{\|\bo^i_L - \bx(\br)\|},
\end{equation}
 Spotlight's diffuse color intensity is brightest on the central ray $\br(t) = \bo_L -t\bd_L$, decays with distance from ray
$\br(t)$ and angle. We modulate it with constant $k$.   

The realistic night-time simulation requires reproducing the strong specular effects on cars.  
We find car regions using a semantic field $\bS$ in Eq.~\ref{eq:volume_rendering}, then simulate specular reflection with the Blinn-Phong
model~\cite{blinn1977models}, where the $\gamma$ (specular strength) parameter is inherited from the semantic field.

At night, luminaires often display lens flares.
A pure simulation of lens flares is impractical, as it requires extensive ray tracing through the lens. 
We use the standard image-based approximation~\cite{akenine2019real} to simulate such light scattering effects. 
For directly visible luminaires, we composite a real-world lens flare image from a similar lighting source 
into the image, using location and depth. 
As Fig.~\ref{tab:relight_night},~\ref{fig:waymo_relight} in the main paper show, this simple method is effective. 

\noindent\emph{Object insertion} proceeds by a hybrid rendering strategy. 
We first cast rays from the camera and estimate ray-mesh intersections~\cite{trimesh} for the inserted object. If the ray hits the mesh and the distance is shorter than the volume rendering depth, the albedo $A(\br)$, normal $N(\br)$, and depth $D(\br)$ are replaced with the object
attributes. In the shadow pass, we calculate visibility from surface points to the light source
(Eq.~\ref{eq:visibility}), and also estimate the ray-mesh intersection for the tracing rays. 
If the rays hit the mesh (meaning occlusion by the object), the visibility is also updated %
: $V(\br) = 0$. 
With updated $A(\br), N(\br), V(\br)$, shading is applied to render images with virtual objects. 
Our method not only casts object shadows in the scene but also casts \emph{scene shadows} on the object, enhancing realism significantly. 
Similar approaches have been depicted in recent works~\cite{li2023climatenerf, qiao2023dynamic}. 
However, ours is the first to be visibility-aware, enabling us to render effects when an object enters into a shadow. 

\noindent\emph{Outdoor relighting} is done by simply adjusting lighting parameters (position or color of the sun;  sky color)
then re-rendering using Eq.~\ref{eq:rendering} in the main paper. 
 We also use semantics to interpret specular car surfaces and emulate their reflectance during the simulation.

\section{Baseline Details}
Description of the approach of baselines we compared to.

\paragraph{Instruct-Pix2Pix~\cite{brooks2023instructpix2pix}} edits images according to user instruction. The model leverages large language model GPT-3~\cite{brown2020language} and Stable Diffusion~\cite{rombach2022high} for generating image and instruction pairs and fine-tune diffusion model to perform editing. We use instructions ``change to night'', and ``It's now midnight'' for night image generation. 

\paragraph{Instruct-NeRF2NeRF~\cite{instructnerf2023}} aims to edit NeRF scenes with text instructions. 
It uses a generative image editing model~\cite{brooks2023instructpix2pix} to iteratively edit input images while optimizing the underlying scene model, resulting in an optimized 3D scene that respects the instruction. 
We compare Instruct NeRF2NeRF in night simulation, where we provide the instruction, ``{\it Make it look like it was taken at night}.''

\paragraph{NeRF-OSR~\cite{rudnev2022nerfosr}}~ is a recent work for outdoor scene reconstruction and relighting. 
We use the open-source project provided by the author to run this baseline.
This method represents lighting as spherical harmonics parameters. 
It is worth noting that NeRF-OSR was designed for inverse rendering in \emph{multi-illumination conditions}. 
For a fair comparison, we rotate the spherical vectors to simulate different light conditions. 

\paragraph{RelightNet~\cite{yu20relightNet}}~ is a single-image based relighting framework. 
We use the open-source project provided by the authors to produce intrinsic decomposition results, including shading and albedo for comparison.

\paragraph{ShadowFormer~\cite{guo2023shadowformer}} performs single-image shadow removal task. It leverages the transformer architecture and takes the original image and shadow masks as input. In Fig.~\ref{tab:deshadow} in the main paper, we first estimate the shadow mask with MTMT~\cite{chen20MTMT}, and use the open-source project and pre-trained weights provided by the authors to estimate the base color of an image.


\begin{thebibliography}{86}
\providecommand{\natexlab}[1]{#1}
\providecommand{\url}[1]{\texttt{#1}}
\expandafter\ifx\csname urlstyle\endcsname\relax
  \providecommand{\doi}[1]{doi: #1}\else
  \providecommand{\doi}{doi: \begingroup \urlstyle{rm}\Url}\fi

\bibitem[Akenine-Moller et~al.(2019)Akenine-Moller, Haines, and Hoffman]{akenine2019real}
Tomas Akenine-Moller, Eric Haines, and Naty Hoffman.
\newblock \emph{Real-time rendering}.
\newblock AK Peters/crc Press, 2019.

\bibitem[Barron and Malik(2014)]{barron2014shape}
Jonathan~T Barron and Jitendra Malik.
\newblock Shape, illumination, and reflectance from shading.
\newblock \emph{TPAMI}, 2014.

\bibitem[Barrow and Tenenbaum(1978)]{Barrow1978}
H.G. Barrow and Joan~M. Tenenbaum.
\newblock Recovering intrinsic scene characteristics from images.
\newblock \emph{Computer Vision Systems}, 1978.

\bibitem[Bell et~al.(2014)Bell, Bala, and Snavely]{Bell2014}
Sean Bell, Kavita Bala, and Noah Snavely.
\newblock Intrinsic images in the wild.
\newblock 2014.

\bibitem[Bhattad and Forsyth(2023)]{bhattad2023stylitgan}
Anand Bhattad and D.~A. Forsyth.
\newblock Stylitgan: Prompting stylegan to produce new illumination conditions, 2023.

\bibitem[Bhattad et~al.(2023)Bhattad, McKee, Hoiem, and Forsyth]{bhattad2023stylegan}
Anand Bhattad, Daniel McKee, Derek Hoiem, and DA Forsyth.
\newblock Stylegan knows normal, depth, albedo, and more.
\newblock \emph{arXiv preprint arXiv:2306.00987}, 2023.

\bibitem[Blinn(1977)]{blinn1977models}
James~F Blinn.
\newblock Models of light reflection for computer synthesized pictures.
\newblock In \emph{Proceedings of the 4th annual conference on Computer graphics and interactive techniques}, pages 192--198, 1977.

\bibitem[Boss et~al.(2021{\natexlab{a}})Boss, Braun, Jampani, Barron, Liu, and Lensch]{boss2021nerd}
Mark Boss, Raphael Braun, Varun Jampani, Jonathan~T. Barron, Ce Liu, and Hendrik~P.A. Lensch.
\newblock Nerd: Neural reflectance decomposition from image collections.
\newblock In \emph{ICCV}, 2021{\natexlab{a}}.

\bibitem[Boss et~al.(2021{\natexlab{b}})Boss, Jampani, Braun, Liu, Barron, and Lensch]{bossNeuralPILNeuralPreIntegrated2021}
Mark Boss, Varun Jampani, Raphael Braun, Ce Liu, Jonathan~T. Barron, and Hendrik~P.A. Lensch.
\newblock Neural-{{PIL}}: {{Neural Pre-Integrated Lighting}} for {{Reflectance Decomposition}}.
\newblock In \emph{Advances in {{Neural Information Processing Systems}} ({{NeurIPS}})}, 2021{\natexlab{b}}.

\bibitem[Brooks et~al.(2023)Brooks, Holynski, and Efros]{brooks2023instructpix2pix}
Tim Brooks, Aleksander Holynski, and Alexei~A Efros.
\newblock Instructpix2pix: Learning to follow image editing instructions.
\newblock In \emph{CVPR}, 2023.

\bibitem[Brown et~al.(2020)Brown, Mann, Ryder, Subbiah, Kaplan, Dhariwal, Neelakantan, Shyam, Sastry, Askell, et~al.]{brown2020language}
Tom Brown, Benjamin Mann, Nick Ryder, Melanie Subbiah, Jared~D Kaplan, Prafulla Dhariwal, Arvind Neelakantan, Pranav Shyam, Girish Sastry, Amanda Askell, et~al.
\newblock Language models are few-shot learners.
\newblock \emph{NeurIPS}, 2020.

\bibitem[Chen et~al.(2020)Chen, Zhu, Wan, Wang, Feng, and Heng]{chen20MTMT}
Zhihao Chen, Lei Zhu, Liang Wan, Song Wang, Wei Feng, and Pheng-Ann Heng.
\newblock A multi-task mean teacher for semi-supervised shadow detection.
\newblock In \emph{CVPR}, 2020.

\bibitem[Community(2018)]{blender}
Blender~Online Community.
\newblock \emph{Blender - a 3D modelling and rendering package}.
\newblock Blender Foundation, Stichting Blender Foundation, Amsterdam, 2018.

\bibitem[Contributors(2020)]{mmseg2020}
MMSegmentation Contributors.
\newblock {MMSegmentation}: Openmmlab semantic segmentation toolbox and benchmark.
\newblock \url{https://github.com/open-mmlab/mmsegmentation}, 2020.

\bibitem[{Dawson-Haggerty et al.}()]{trimesh}
{Dawson-Haggerty et al.}
\newblock trimesh.

\bibitem[Dong et~al.(2014)Dong, Chen, Peers, Zhang, and Tong]{dong2014appearance}
Yue Dong, Guojun Chen, Pieter Peers, Jiawan Zhang, and Xin Tong.
\newblock Appearance-from-motion: Recovering spatially varying surface reflectance under unknown lighting.
\newblock \emph{ACM Transactions on Graphics (TOG)}, 33\penalty0 (6):\penalty0 1--12, 2014.

\bibitem[Eftekhar et~al.(2021)Eftekhar, Sax, Malik, and Zamir]{eftekhar2021omnidata}
Ainaz Eftekhar, Alexander Sax, Jitendra Malik, and Amir Zamir.
\newblock Omnidata: A scalable pipeline for making multi-task mid-level vision datasets from 3d scans.
\newblock In \emph{ICCV}, 2021.

\bibitem[Forsyth and Rock(2021)]{forsyth2021intrinsic}
David Forsyth and Jason~J Rock.
\newblock Intrinsic image decomposition using paradigms.
\newblock \emph{IEEE transactions on pattern analysis and machine intelligence}, 44\penalty0 (11):\penalty0 7624--7637, 2021.

\bibitem[Grosse et~al.(2009)Grosse, Johnson, Adelson, and Freeman]{grosse2009ground}
Roger Grosse, Micah~K Johnson, Edward~H Adelson, and William~T Freeman.
\newblock Ground truth dataset and baseline evaluations for intrinsic image algorithms.
\newblock In \emph{ICCV}, 2009.

\bibitem[Guo et~al.(2022)Guo, Wang, Yang, Huang, Wang, Pfister, and Wen]{guo2022shadowdiffusion}
Lanqing Guo, Chong Wang, Wenhan Yang, Siyu Huang, Yufei Wang, Hanspeter Pfister, and Bihan Wen.
\newblock Shadowdiffusion: When degradation prior meets diffusion model for shadow removal.
\newblock \emph{arXiv preprint arXiv:2212.04711}, 2022.

\bibitem[Guo et~al.(2023)Guo, Huang, Liu, Cheng, and Wen]{guo2023shadowformer}
Lanqing Guo, Siyu Huang, Ding Liu, Hao Cheng, and Bihan Wen.
\newblock Shadowformer: Global context helps image shadow removal.
\newblock \emph{AAAI}, 2023.

\bibitem[Haque et~al.(2023)Haque, Tancik, Efros, Holynski, and Kanazawa]{instructnerf2023}
Ayaan Haque, Matthew Tancik, Alexei Efros, Aleksander Holynski, and Angjoo Kanazawa.
\newblock Instruct-nerf2nerf: Editing 3d scenes with instructions.
\newblock In \emph{Proceedings of the IEEE/CVF International Conference on Computer Vision}, 2023.

\bibitem[Hasselgren et~al.(2022)Hasselgren, Hofmann, and Munkberg]{hasselgren2022nvdiffrecmc}
Jon Hasselgren, Nikolai Hofmann, and Jacob Munkberg.
\newblock {Shape, Light, and Material Decomposition from Images using Monte Carlo Rendering and Denoising}.
\newblock \emph{arXiv:2206.03380}, 2022.

\bibitem[Hauagge et~al.(2013)Hauagge, Wehrwein, Bala, and Snavely]{hauagge2013photometric}
Daniel Hauagge, Scott Wehrwein, Kavita Bala, and Noah Snavely.
\newblock Photometric ambient occlusion.
\newblock In \emph{CVPR}, 2013.

\bibitem[Horn(1974)]{horn1974determining}
Berthold~KP Horn.
\newblock Determining lightness from an image.
\newblock \emph{Computer graphics and image processing}, 1974.

\bibitem[Horn(1975)]{horn1975obtaining}
Berthold~KP Horn.
\newblock Obtaining shape from shading information.
\newblock \emph{The psychology of computer vision}, 1975.

\bibitem[Jin et~al.(2023)Jin, Liu, Xu, Zhang, Han, Bi, Zhou, Xu, and Su]{jin2023tensoir}
Haian Jin, Isabella Liu, Peijia Xu, Xiaoshuai Zhang, Songfang Han, Sai Bi, Xiaowei Zhou, Zexiang Xu, and Hao Su.
\newblock Tensoir: Tensorial inverse rendering.
\newblock \emph{CVPR}, 2023.

\bibitem[Kajiya and Von~Herzen(1984)]{kajiya1984ray}
James~T Kajiya and Brian~P Von~Herzen.
\newblock Ray tracing volume densities.
\newblock \emph{ACM SIGGRAPH computer graphics}, 1984.

\bibitem[Kar et~al.(2022)Kar, Yeo, Atanov, and Zamir]{kar20223d}
O{\u{g}}uzhan~Fatih Kar, Teresa Yeo, Andrei Atanov, and Amir Zamir.
\newblock 3d common corruptions and data augmentation.
\newblock In \emph{CVPR}, 2022.

\bibitem[Kingma and Ba(2014)]{kingma2014adam}
Diederik~P. Kingma and Jimmy Ba.
\newblock Adam: A method for stochastic optimization.
\newblock \emph{ICLR}, 2014.

\bibitem[Knapitsch et~al.(2017)Knapitsch, Park, Zhou, and Koltun]{Knapitsch2017}
Arno Knapitsch, Jaesik Park, Qian-Yi Zhou, and Vladlen Koltun.
\newblock Tanks and temples: Benchmarking large-scale scene reconstruction.
\newblock \emph{ACM TOG}, 2017.

\bibitem[Laffont et~al.(2012)Laffont, Bousseau, and Drettakis]{laffont2012rich}
Pierre-Yves Laffont, Adrien Bousseau, and George Drettakis.
\newblock Rich intrinsic image decomposition of outdoor scenes from multiple views.
\newblock \emph{IEEE transactions on visualization and computer graphics}, 2012.

\bibitem[Laine et~al.(2005)Laine, Aila, Assarsson, Lehtinen, and Akenine-M{\"o}ller]{laine2005soft}
Samuli Laine, Timo Aila, Ulf Assarsson, Jaakko Lehtinen, and Tomas Akenine-M{\"o}ller.
\newblock Soft shadow volumes for ray tracing.
\newblock In \emph{ACM SIGGRAPH 2005 Papers}, pages 1156--1165. 2005.

\bibitem[Lalonde and Matthews(2014)]{lalonde2014lighting}
Jean-Fran{\c{c}}ois Lalonde and Iain Matthews.
\newblock Lighting estimation in outdoor image collections.
\newblock In \emph{International Conference on 3D Vision (3DV)}. IEEE, 2014.

\bibitem[Land and McCann(1971)]{land1971lightness}
Edwin~H Land and John~J McCann.
\newblock Lightness and retinex theory.
\newblock \emph{Josa}, 1971.

\bibitem[Lensch et~al.(2003)Lensch, Kautz, Goesele, Heidrich, and Seidel]{lensch2003image}
Hendrik~PA Lensch, Jan Kautz, Michael Goesele, Wolfgang Heidrich, and Hans-Peter Seidel.
\newblock Image-based reconstruction of spatial appearance and geometric detail.
\newblock \emph{TOG}, 2003.

\bibitem[Li et~al.(2023)Li, Lin, Forsyth, Huang, and Wang]{li2023climatenerf}
Yuan Li, Zhi-Hao Lin, David Forsyth, Jia-Bin Huang, and Shenlong Wang.
\newblock Climatenerf: Extreme weather synthesis in neural radiance field.
\newblock In \emph{Proceedings of the IEEE/CVF International Conference on Computer Vision}, pages 3227--3238, 2023.

\bibitem[Li et~al.(2018{\natexlab{a}})Li, Sunkavalli, and Chandraker]{li2018materials}
Zhengqin Li, Kalyan Sunkavalli, and Manmohan Chandraker.
\newblock Materials for masses: {SVBRDF} acquisition with a single mobile phone image.
\newblock In \emph{ECCV}, pages 72--87, 2018{\natexlab{a}}.

\bibitem[Li et~al.(2018{\natexlab{b}})Li, Xu, Ramamoorthi, Sunkavalli, and Chandraker]{li2018learning}
Zhengqin Li, Zexiang Xu, Ravi Ramamoorthi, Kalyan Sunkavalli, and Manmohan Chandraker.
\newblock Learning to reconstruct shape and spatially-varying reflectance from a single image.
\newblock \emph{ACM Transactions on Graphics (TOG)}, 37\penalty0 (6):\penalty0 1--11, 2018{\natexlab{b}}.

\bibitem[Li et~al.(2020)Li, Shafiei, Ramamoorthi, Sunkavalli, and Chandraker]{li2020inverse}
Zhengqin Li, Mohammad Shafiei, Ravi Ramamoorthi, Kalyan Sunkavalli, and Manmohan Chandraker.
\newblock Inverse rendering for complex indoor scenes: Shape, spatially-varying lighting and svbrdf from a single image.
\newblock In \emph{CVPR}, 2020.

\bibitem[Li et~al.(2022)Li, Shi, Bi, Zhu, Sunkavalli, Ha{\v{s}}an, Xu, Ramamoorthi, and Chandraker]{li2022physically}
Zhengqin Li, Jia Shi, Sai Bi, Rui Zhu, Kalyan Sunkavalli, Milo{\v{s}} Ha{\v{s}}an, Zexiang Xu, Ravi Ramamoorthi, and Manmohan Chandraker.
\newblock Physically-based editing of indoor scene lighting from a single image.
\newblock \emph{ECCV}, 2022.

\bibitem[Liao et~al.(2021)Liao, Xie, and Geiger]{Liao2021ARXIV}
Yiyi Liao, Jun Xie, and Andreas Geiger.
\newblock {KITTI}-360: A novel dataset and benchmarks for urban scene understanding in 2d and 3d.
\newblock \emph{in arXiv}, 2021.

\bibitem[Lichy et~al.(2021)Lichy, Wu, Sengupta, and Jacobs]{lichy2021shape}
Daniel Lichy, Jiaye Wu, Soumyadip Sengupta, and David~W Jacobs.
\newblock Shape and material capture at home.
\newblock In \emph{Proceedings of the IEEE/CVF Conference on Computer Vision and Pattern Recognition}, pages 6123--6133, 2021.

\bibitem[Lombardi and Nishino(2015)]{lombardi2015reflectance}
Stephen Lombardi and Ko Nishino.
\newblock Reflectance and illumination recovery in the wild.
\newblock \emph{IEEE transactions on pattern analysis and machine intelligence}, 38\penalty0 (1):\penalty0 129--141, 2015.

\bibitem[Lombardi and Nishino(2016)]{lombardi2016radiometric}
Stephen Lombardi and Ko Nishino.
\newblock Radiometric scene decomposition: Scene reflectance, illumination, and geometry from rgb-d images.
\newblock In \emph{2016 Fourth International Conference on 3D Vision (3DV)}, pages 305--313. IEEE, 2016.

\bibitem[Lombardi et~al.(2019)Lombardi, Simon, Saragih, Schwartz, Lehrmann, and Sheikh]{lombardi2019neural}
Stephen Lombardi, Tomas Simon, Jason Saragih, Gabriel Schwartz, Andreas Lehrmann, and Yaser Sheikh.
\newblock Neural volumes: Learning dynamic renderable volumes from images.
\newblock \emph{ACM Transactions on Graphics (TOG)}, 38\penalty0 (4):\penalty0 65, 2019.

\bibitem[Ma et~al.(2018)Ma, Chu, Zhou, Urtasun, and Torralba]{ma2018single}
Wei-Chiu Ma, Hang Chu, Bolei Zhou, Raquel Urtasun, and Antonio Torralba.
\newblock Single image intrinsic decomposition without a single intrinsic image.
\newblock In \emph{ECCV}, 2018.

\bibitem[Marschner(1998)]{marschner1998inverse}
Stephen~Robert Marschner.
\newblock \emph{Inverse rendering for computer graphics}.
\newblock Cornell University, 1998.

\bibitem[Mildenhall et~al.(2020)Mildenhall, Srinivasan, Tancik, Barron, Ramamoorthi, and Ng]{mildenhall2020nerf}
Ben Mildenhall, Pratul~P. Srinivasan, Matthew Tancik, Jonathan~T. Barron, Ravi Ramamoorthi, and Ren Ng.
\newblock Nerf: Representing scenes as neural radiance fields for view synthesis.
\newblock In \emph{ECCV}, 2020.

\bibitem[M\"uller(2021)]{tiny-cuda-nn}
Thomas M\"uller.
\newblock {tiny-cuda-nn}, 2021.

\bibitem[M{\"u}ller et~al.(2022)M{\"u}ller, Evans, Schied, and Keller]{muller2022instant}
Thomas M{\"u}ller, Alex Evans, Christoph Schied, and Alexander Keller.
\newblock Instant neural graphics primitives with a multiresolution hash encoding.
\newblock \emph{ACM TOG}, 2022.

\bibitem[Munkberg et~al.(2021)Munkberg, Hasselgren, Shen, Gao, Chen, Evans, M{\"u}ller, and Fidler]{munkbergExtractingTriangular3D2021}
Jacob Munkberg, Jon Hasselgren, Tianchang Shen, Jun Gao, Wenzheng Chen, Alex Evans, Thomas M{\"u}ller, and Sanja Fidler.
\newblock Extracting {{Triangular 3D Models}}, {{Materials}}, and {{Lighting From Images}}.
\newblock \emph{arXiv:2111.12503 [cs]}, 2021.

\bibitem[Munkberg et~al.(2022)Munkberg, Hasselgren, Shen, Gao, Chen, Evans, M{\"u}ller, and Fidler]{munkberg2022extracting}
Jacob Munkberg, Jon Hasselgren, Tianchang Shen, Jun Gao, Wenzheng Chen, Alex Evans, Thomas M{\"u}ller, and Sanja Fidler.
\newblock Extracting triangular 3d models, materials, and lighting from images.
\newblock In \emph{CVPR}, 2022.

\bibitem[Pun et~al.(2023{\natexlab{a}})Pun, Sun, Wang, Chen, Yang, Manivasagam, Ma, and Urtasun]{pun2023lightsim}
Ava Pun, Gary Sun, Jingkang Wang, Yun Chen, Ze Yang, Sivabalan Manivasagam, Wei-Chiu Ma, and Raquel Urtasun.
\newblock Lightsim: Neural lighting simulation for urban scenes.
\newblock \emph{NeurIPS}, 2023{\natexlab{a}}.

\bibitem[Pun et~al.(2023{\natexlab{b}})Pun, Sun, Wang, Chen, Yang, Manivasagam, Ma, and Urtasun]{pun2023neural}
Ava Pun, Gary Sun, Jingkang Wang, Yun Chen, Ze Yang, Sivabalan Manivasagam, Wei-Chiu Ma, and Raquel Urtasun.
\newblock Neural lighting simulation for urban scenes.
\newblock In \emph{Thirty-seventh Conference on Neural Information Processing Systems}, 2023{\natexlab{b}}.

\bibitem[Qiao et~al.(2023)Qiao, Gao, Xu, Feng, Huang, and Lin]{qiao2023dynamic}
Yi-Ling Qiao, Alexander Gao, Yiran Xu, Yue Feng, Jia-Bin Huang, and Ming~C Lin.
\newblock Dynamic mesh-aware radiance fields.
\newblock In \emph{Proceedings of the IEEE/CVF International Conference on Computer Vision}, pages 385--396, 2023.

\bibitem[Quei-An(2022)]{queianchen_ngp}
Chen Quei-An.
\newblock ngp\_pl: a pytorch-lightning implementation of instant-ngp, 2022.

\bibitem[Rombach et~al.(2022)Rombach, Blattmann, Lorenz, Esser, and Ommer]{rombach2022high}
Robin Rombach, Andreas Blattmann, Dominik Lorenz, Patrick Esser, and Bj{\"o}rn Ommer.
\newblock High-resolution image synthesis with latent diffusion models.
\newblock In \emph{CVPR}, 2022.

\bibitem[Rudnev et~al.(2022{\natexlab{a}})Rudnev, Elgharib, Smith, Liu, Golyanik, and Theobalt]{rudnev2021neural}
Viktor Rudnev, Mohamed Elgharib, William Smith, Lingjie Liu, Vladislav Golyanik, and Christian Theobalt.
\newblock Neural radiance fields for outdoor scene relighting.
\newblock \emph{ECCV}, 2022{\natexlab{a}}.

\bibitem[Rudnev et~al.(2022{\natexlab{b}})Rudnev, Elgharib, Smith, Liu, Golyanik, and Theobalt]{rudnev2022nerfosr}
Viktor Rudnev, Mohamed Elgharib, William Smith, Lingjie Liu, Vladislav Golyanik, and Christian Theobalt.
\newblock Nerf for outdoor scene relighting.
\newblock In \emph{European Conference on Computer Vision (ECCV)}, 2022{\natexlab{b}}.

\bibitem[Sato et~al.(2003)Sato, Sato, and Ikeuchi]{sato2003illumination}
Imari Sato, Yoichi Sato, and Katsushi Ikeuchi.
\newblock Illumination from shadows.
\newblock \emph{IEEE Transactions on Pattern Analysis and Machine Intelligence}, 2003.

\bibitem[Sato et~al.(1997)Sato, Wheeler, and Ikeuchi]{sato1997object}
Yoichi Sato, Mark~D Wheeler, and Katsushi Ikeuchi.
\newblock Object shape and reflectance modeling from observation.
\newblock In \emph{SIGGRAPH}, 1997.

\bibitem[Sengupta et~al.(2019)Sengupta, Gu, Kim, Liu, Jacobs, and Kautz]{sengupta2019neural}
Soumyadip Sengupta, Jinwei Gu, Kihwan Kim, Guilin Liu, David~W Jacobs, and Jan Kautz.
\newblock Neural inverse rendering of an indoor scene from a single image.
\newblock In \emph{Proceedings of the IEEE/CVF International Conference on Computer Vision}, pages 8598--8607, 2019.

\bibitem[Srinivasan et~al.(2021)Srinivasan, Deng, Zhang, Tancik, Mildenhall, and Barron]{srinivasan2021nerv}
Pratul~P Srinivasan, Boyang Deng, Xiuming Zhang, Matthew Tancik, Ben Mildenhall, and Jonathan~T Barron.
\newblock Nerv: Neural reflectance and visibility fields for relighting and view synthesis.
\newblock In \emph{CVPR}, 2021.

\bibitem[Story(2015)]{story2015hybrid}
Jon Story.
\newblock Hybrid ray traced shadows.
\newblock In \emph{Game Developer Conference}, 2015.

\bibitem[Sun et~al.(2020)Sun, Kretzschmar, Dotiwalla, Chouard, Patnaik, Tsui, Guo, Zhou, Chai, Caine, Vasudevan, Han, Ngiam, Zhao, Timofeev, Ettinger, Krivokon, Gao, Joshi, Zhang, Shlens, Chen, and Anguelov]{Sun_2020_CVPR}
Pei Sun, Henrik Kretzschmar, Xerxes Dotiwalla, Aurelien Chouard, Vijaysai Patnaik, Paul Tsui, James Guo, Yin Zhou, Yuning Chai, Benjamin Caine, Vijay Vasudevan, Wei Han, Jiquan Ngiam, Hang Zhao, Aleksei Timofeev, Scott Ettinger, Maxim Krivokon, Amy Gao, Aditya Joshi, Yu Zhang, Jonathon Shlens, Zhifeng Chen, and Dragomir Anguelov.
\newblock Scalability in perception for autonomous driving: Waymo open dataset.
\newblock In \emph{CVPR}, 2020.

\bibitem[Verbin et~al.(2022)Verbin, Hedman, Mildenhall, Zickler, Barron, and Srinivasan]{verbin2022refnerf}
Dor Verbin, Peter Hedman, Ben Mildenhall, Todd Zickler, Jonathan~T. Barron, and Pratul~P. Srinivasan.
\newblock {Ref-NeRF}: Structured view-dependent appearance for neural radiance fields.
\newblock \emph{CVPR}, 2022.

\bibitem[Wan et~al.(2022)Wan, Yin, Wu, Wu, Liu, and Wang]{wan2022style}
Jin Wan, Hui Yin, Zhenyao Wu, Xinyi Wu, Yanting Liu, and Song Wang.
\newblock Style-guided shadow removal.
\newblock In \emph{Computer Vision--ECCV 2022: 17th European Conference, Tel Aviv, Israel, October 23--27, 2022, Proceedings, Part XIX}, pages 361--378. Springer, 2022.

\bibitem[Wang et~al.(2021)Wang, Liu, Tucker, Wu, Curless, Seitz, and Snavely]{wang2021repopulating}
Yifan Wang, Andrew Liu, Richard Tucker, Jiajun Wu, Brian~L Curless, Steven~M Seitz, and Noah Snavely.
\newblock Repopulating street scenes.
\newblock In \emph{Proceedings of the IEEE/CVF Conference on Computer Vision and Pattern Recognition}, pages 5110--5119, 2021.

\bibitem[Wang et~al.(2022)Wang, Chen, Acuna, Kautz, and Fidler]{wang2022neural}
Zian Wang, Wenzheng Chen, David Acuna, Jan Kautz, and Sanja Fidler.
\newblock Neural light field estimation for street scenes with differentiable virtual object insertion.
\newblock In \emph{ECCV}, 2022.

\bibitem[Wang et~al.(2023{\natexlab{a}})Wang, Shen, Gao, Huang, Munkberg, Hasselgren, Gojcic, Chen, and Fidler]{wang2023fegr}
Zian Wang, Tianchang Shen, Jun Gao, Shengyu Huang, Jacob Munkberg, Jon Hasselgren, Zan Gojcic, Wenzheng Chen, and Sanja Fidler.
\newblock Neural fields meet explicit geometric representations for inverse rendering of urban scenes.
\newblock In \emph{CVPR}, 2023{\natexlab{a}}.

\bibitem[Wang et~al.(2023{\natexlab{b}})Wang, Shen, Gao, Huang, Munkberg, Hasselgren, Gojcic, Chen, and Fidler]{wang2023neural}
Zian Wang, Tianchang Shen, Jun Gao, Shengyu Huang, Jacob Munkberg, Jon Hasselgren, Zan Gojcic, Wenzheng Chen, and Sanja Fidler.
\newblock Neural fields meet explicit geometric representation for inverse rendering of urban scenes.
\newblock \emph{arXiv}, 2023{\natexlab{b}}.

\bibitem[Wu et~al.(2007)Wu, Tang, Brown, and Shum]{wu2007natural}
Tai-Pang Wu, Chi-Keung Tang, Michael~S Brown, and Heung-Yeung Shum.
\newblock Natural shadow matting.
\newblock \emph{ACM Transactions on Graphics (TOG)}, 26\penalty0 (2):\penalty0 8--es, 2007.

\bibitem[Yang et~al.(2022)Yang, Chen, Chen, Chen, and Wong]{yang2022s}
Wenqi Yang, Guanying Chen, Chaofeng Chen, Zhenfang Chen, and Kwan-Yee~K Wong.
\newblock S3-nerf: Neural reflectance field from shading and shadow under a single viewpoint.
\newblock \emph{arXiv preprint arXiv:2210.08936}, 2022.

\bibitem[Yu and Smith(2019)]{yu2019inverserendernet}
Ye Yu and William~AP Smith.
\newblock Inverserendernet: Learning single image inverse rendering.
\newblock In \emph{Proceedings of the IEEE/CVF Conference on Computer Vision and Pattern Recognition}, pages 3155--3164, 2019.

\bibitem[Yu et~al.(1999)Yu, Debevec, Malik, and Hawkins]{yu1999inverse}
Yizhou Yu, Paul Debevec, Jitendra Malik, and Tim Hawkins.
\newblock Inverse global illumination: Recovering reflectance models of real scenes from photographs.
\newblock In \emph{Proceedings of the 26th annual conference on Computer graphics and interactive techniques}, 1999.

\bibitem[Yu et~al.(2020)Yu, Meka, Elgharib, Seidel, Theobalt, and Smith]{yu20relightNet}
Ye Yu, Abhimitra Meka, Mohamed Elgharib, Hans-Peter Seidel, Christian Theobalt, and William A.~P. Smith.
\newblock Self-supervised outdoor scene relighting.
\newblock In \emph{ECCV}, 2020.

\bibitem[Zhang et~al.(2019{\natexlab{a}})Zhang, Sunkavalli, Hold-Geoffroy, Hadap, Eisenman, and Lalonde]{zhang2019all}
Jinsong Zhang, Kalyan Sunkavalli, Yannick Hold-Geoffroy, Sunil Hadap, Jonathan Eisenman, and Jean-Fran{\c{c}}ois Lalonde.
\newblock All-weather deep outdoor lighting estimation.
\newblock In \emph{CVPR}, 2019{\natexlab{a}}.

\bibitem[Zhang et~al.(2021{\natexlab{a}})Zhang, Luan, Wang, Bala, and Snavely]{zhang2021physg}
Kai Zhang, Fujun Luan, Qianqian Wang, Kavita Bala, and Noah Snavely.
\newblock Physg: Inverse rendering with spherical gaussians for physics-based material editing and relighting.
\newblock In \emph{CVPR}, 2021{\natexlab{a}}.

\bibitem[Zhang et~al.(2022{\natexlab{a}})Zhang, Luan, Li, and Snavely]{zhang2022iron}
Kai Zhang, Fujun Luan, Zhengqi Li, and Noah Snavely.
\newblock Iron: Inverse rendering by optimizing neural sdfs and materials from photometric images.
\newblock In \emph{CVPR}, 2022{\natexlab{a}}.

\bibitem[Zhang et~al.(1999)Zhang, Tsai, Cryer, and Shah]{Zhang-sfs-99}
Ruo Zhang, Ping-Sing Tsai, James~Edwin Cryer, and Mubarak Shah.
\newblock Shape from shading: A survey.
\newblock \emph{IEEE TPAMI}, 1999.

\bibitem[Zhang et~al.(2019{\natexlab{b}})Zhang, Liang, and Wang]{zhang2019shadowgan}
Shuyang Zhang, Runze Liang, and Miao Wang.
\newblock Shadowgan: Shadow synthesis for virtual objects with conditional adversarial networks.
\newblock \emph{Computational Visual Media}, 2019{\natexlab{b}}.

\bibitem[Zhang et~al.(2021{\natexlab{b}})Zhang, Srinivasan, Deng, Debevec, Freeman, and Barron]{zhang2021nerfactor}
Xiuming Zhang, Pratul~P Srinivasan, Boyang Deng, Paul Debevec, William~T Freeman, and Jonathan~T Barron.
\newblock Nerfactor: Neural factorization of shape and reflectance under an unknown illumination.
\newblock \emph{ACM TOG}, 2021{\natexlab{b}}.

\bibitem[Zhang et~al.(2020)Zhang, Chen, Ling, Gao, Zhang, Torralba, and Fidler]{zhang2020image}
Yuxuan Zhang, Wenzheng Chen, Huan Ling, Jun Gao, Yinan Zhang, Antonio Torralba, and Sanja Fidler.
\newblock Image gans meet differentiable rendering for inverse graphics and interpretable 3d neural rendering.
\newblock \emph{arXiv}, 2020.

\bibitem[Zhang et~al.(2022{\natexlab{b}})Zhang, Sun, He, Fu, Jia, and Zhou]{zhang2022invrender}
Yuanqing Zhang, Jiaming Sun, Xingyi He, Huan Fu, Rongfei Jia, and Xiaowei Zhou.
\newblock Modeling indirect illumination for inverse rendering.
\newblock In \emph{CVPR}, 2022{\natexlab{b}}.

\bibitem[Zhou et~al.(2015)Zhou, Krahenbuhl, and Efros]{zhou2015learning}
Tinghui Zhou, Philipp Krahenbuhl, and Alexei~A Efros.
\newblock Learning data-driven reflectance priors for intrinsic image decomposition.
\newblock In \emph{ICCV}, 2015.

\end{thebibliography}
\end{document}